\documentclass{article} % For LaTeX2e
\usepackage[utf8]{inputenc}
\usepackage{iclr2025_conference,times}
\usepackage{amsmath,amsfonts,bm}

\def\eqref#1{equation~\ref{#1}}
\def\1{\bm{1}}

\def\vs{{\bm{s}}}

\DeclareMathAlphabet{\mathsfit}{\encodingdefault}{\sfdefault}{m}{sl}
\SetMathAlphabet{\mathsfit}{bold}{\encodingdefault}{\sfdefault}{bx}{n}

\usepackage{graphicx}
\usepackage{booktabs} 
\usepackage{amsmath}  
\usepackage{pifont}
\usepackage[
    colorlinks=true,      
    citecolor=citecolor,  
]{hyperref}
\usepackage{amssymb}
\usepackage{bbding}
\usepackage{url}
\usepackage{graphicx}
\usepackage[table]{xcolor}
\usepackage{array}
\usepackage{multirow}
\usepackage{subcaption}
\usepackage{xspace}
\usepackage{makecell}
\usepackage{siunitx}
\usepackage{hhline}
\usepackage{longtable}
\usepackage{tabularx}
\usepackage{booktabs}
\usepackage{amsfonts}
\usepackage{microtype}
\usepackage{boldline}
\usepackage[export]{adjustbox}
\usepackage{fancyvrb}
\usepackage{fvextra}
\usepackage{wrapfig}
\usepackage{pifont}
\usepackage[most,breakable]{tcolorbox}
\usepackage[utf8]{inputenc}

\definecolor{citecolor}{HTML}{0071BC}
\definecolor{lightblue}{rgb}{0.88, 0.95, 1.0}  % Existing light blue
\definecolor{lightgreen}{rgb}{0.56, 0.93, 0.56} % Light green
\definecolor{lightyellow}{rgb}{1.0, 1.0, 0.88}  % Light yellow
\definecolor{lightpink}{rgb}{1.0, 0.71, 0.76}   % Light pink
\definecolor{lightorange}{rgb}{1.0, 0.83, 0.62} % Light orange
\definecolor{lightcyan}{rgb}{0.88, 1.0, 1.0}   % Light cyan
\definecolor{lightpurple}{rgb}{0.9, 0.8, 1.0}   % Light purple / lavender
\definecolor{lightgray}{rgb}{0.83, 0.83, 0.83}  % Light gray
\definecolor{lightgrey}{rgb}{0.83, 0.83, 0.83}  % Alias for light gray
\definecolor{lavender}{rgb}{0.9, 0.8, 1.0}       % 经典薰衣草色
\definecolor{lilac}{rgb}{0.78, 0.64, 0.8}        % 优雅丁香紫
\definecolor{periwinkle}{rgb}{0.8, 0.8, 1.0}     % 清新矢车菊蓝紫
\definecolor{mauve}{rgb}{0.87, 0.63, 0.87}       % 柔和淡紫红
\definecolor{orchid}{rgb}{0.85, 0.44, 0.84}      % 兰花紫
\definecolor{amethyst}{rgb}{0.6, 0.4, 0.8}       % 紫水晶色
\definecolor{wisteria}{rgb}{0.79, 0.63, 0.86}    % 紫藤色
\definecolor{dustylavender}{rgb}{0.76, 0.7, 0.86} % 灰薰衣草色
\definecolor{frenchlavender}{rgb}{0.8, 0.6, 0.7} % 法式薰衣草
\definecolor{heliotrope}{rgb}{0.87, 0.73, 1.0}   % 天芥菜紫
\definecolor{plum}{rgb}{0.87, 0.63, 0.87}        % 李子紫
\definecolor{LakeBlue}{RGB}{0,61,153}
\definecolor{veronica-red}{RGB}{196,30,58}

\newcommand{\cmark}{\textcolor{green}{\ding{51}}} % Green check mark
\newcommand{\xmark}{\textcolor{red}{\ding{55}}} % Red cross
\definecolor{softred}{rgb}{0.7,0.2,0.2}
\definecolor{softgreen}{rgb}{0.2,0.6,0.3}

\newcommand\blfootnote[1]{%
\begingroup
\renewcommand\thefootnote{}\footnote{#1}%
\addtocounter{footnote}{-1}%
\endgroup
}
\makeatletter

\DeclareRobustCommand{\modelname}{%
  \@ifnextchar\bgroup{\modelname@check}{\modelname@without}%
}
\newcommand{\modelname@check}[1]{%
  \@ifnextchar\bgroup{\modelname@two{#1}}{\modelname@one{#1}}%
}
\newcommand{\modelname@one}[1]{ScaleCUA-#1B\xspace}
\newcommand{\modelname@two}[2]{ScaleCUA-#1-#2B\xspace}
\newcommand{\modelname@without}{ScaleCUA\xspace}
\def\ourdataset{ScaleCUA-Data\xspace}
\makeatother

\title{\modelname: Scaling Open-Source Computer Use Agents with Cross-Platform Data}

\author{
    \scalebox{0.9}{
    \textbf{\;\qquad Zhaoyang Liu*, 
    Jingjing Xie*, 
    Zichen Ding*, 
    Zehao Li*, 
    Bowen Yang*, 
    Zhenyu Wu*,}}
\\
\scalebox{0.9}{\textbf{
    \; Xuehui Wang, 
    Qiushi Sun,
    Shi Liu,
    Weiyun Wang,
    Shenglong Ye,
    Qingyun Li,
    Xuan Dong, 
    Yue Yu, 
}}
\\
\scalebox{0.9}{\textbf{
    \; Chenyu Lu, 
    YunXiang Mo, 
    Yao Yan, 
    Zeyue Tian,
    Xiao Zhang,
    Yuan Huang,
    Yiqian Liu,
    Weijie Su,
}}
\\
\scalebox{0.9}{\textbf{
    Gen Luo,
    Xiangyu Yue,
    Biqing Qi,
    Bowen Zhou,
    Kai Chen,
    Yu Qiao,
    Qifeng Chen\textsuperscript{\Envelope},
    Wenhai Wang\textsuperscript{\Envelope}
}}
\vspace{8px}
\\
\scalebox{1}{\, \qquad \qquad \qquad \qquad \qquad \qquad \qquad Shanghai AI Laboratory}
\vspace{-22px}
}

\def\ie{\emph{i.e.}}
\def\eg{\emph{e.g.}}
\def\etc{\emph{etc.}}

\def\vs{{\em vs.~}}

\newcommand{\mytoprule}{
    \toprule
    \noalign{\vspace{-0.8mm}}
}

\newcommand{\mymidrule}{
    \noalign{\vspace{-0.8mm}}
    \midrule
    \noalign{\vspace{-1mm}}
}

\newcommand{\mybottomrule}{
    \noalign{\vspace{-0.6mm}}
    \bottomrule
}

\iclrfinalcopy % Uncomment for camera-ready version, but NOT for submission.
\begin{document}

\maketitle

\blfootnote{
* equal contribution; \Envelope\ corresponding authors.
}
\begin{figure}[thb!]
    \vspace{-16px}
    \centering
    \includegraphics[width=0.86\linewidth]{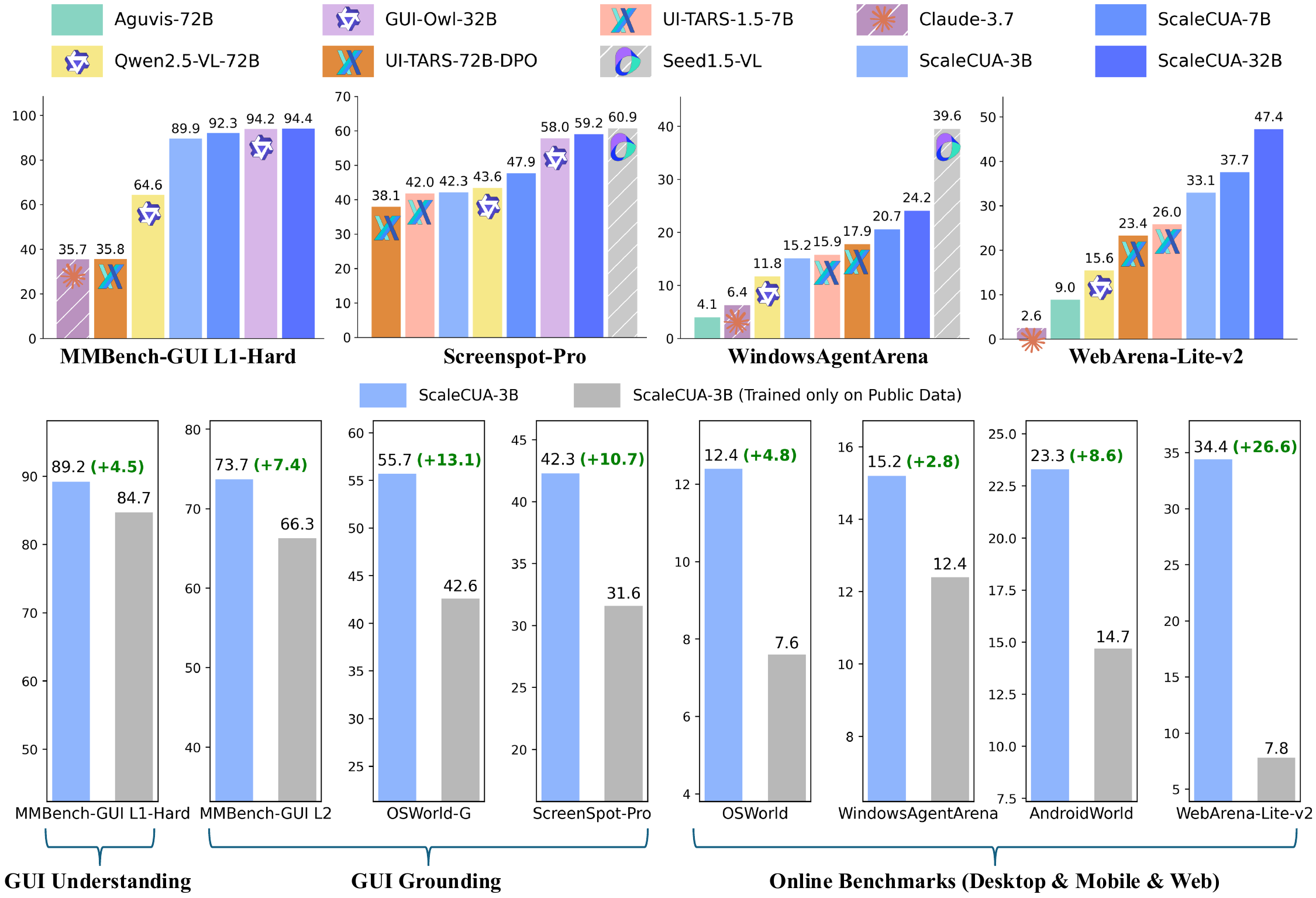}
    \small
    % \caption{Performance comparison on GUI-centric benchmarks. The top row reports results on MMBench-GUI L1-Hard~\citeyearpar{liu2024mmbench}, ScreenSpot-Pro~\citeyearpar{li2025screenspotpro}, WindowsAgentArena~\citeyearpar{bonatti2024windows}, and WebArena-Lite-v2. The bottom row provides a focused comparison between \modelname and baseline model, showing consistent improvements across different task domains.
    \caption{\textbf{Performance comparison}. The top row showcases performance overview on GUI-centric benchmarks. The bottom row demonstrates the consistent improvements from our collected data.
}
    \label{fig:teaser}
\end{figure}
\begin{abstract}
% Vision-Language Models (VLMs) have enabled computer use agents (CUAs) that operate GUIs autonomously, showing great potential, yet progress is limited by the lack of large-scale, open-source computer use data and foundation models.
% %
% In this work, we introduce \modelname, a step toward scaling open-source CUAs. It offers a large-scale dataset spanning 6 operating systems and 3 task domains, built via a closed-loop pipeline uniting automated agents with human experts. Trained on this scaled-up data, \modelname can operate seamlessly across platforms. 
% %
% Specifically, it delivers strong gains over baselines (+26.6 on WebArena-Lite-v2, +10.7 on ScreenSpot-Pro) and sets new state-of-the-art results (94.4\% on MMBench-GUI L1-Hard, 60.6\% on OSWorld-G, 47.4\% on WebArena-Lite-v2). These findings underscore the power of data-driven scaling for general-purpose computer use agents. We release data, models, and code to advance future research: \url{https://github.com/OpenGVLab/ScaleCUA}.

Vision-Language Models (VLMs) have enabled computer use agents (CUAs) that operate GUIs autonomously with great potential. 
However, developing robust CUAs requires extensive in-domain knowledge about software interfaces and operations. 
Unlike image–text pairs that are widely available on the Internet, computer-use data, particularly operation trajectories, are rare, costly to collect. 
Consequently, progress in this field remains constrained by both data scale and the limited transferability of existing VLMs. 
In this work, we introduce \modelname, a step toward scaling open-source CUAs. It offers a large-scale dataset spanning 6 operating systems and 3 task domains, via a closed-loop pipeline uniting automated agents with human experts. Trained on this scaled-up data, \modelname can operate seamlessly across platforms. 
Specifically, it delivers strong gains over baselines (+26.6 on WebArena-Lite-v2, +10.7 on ScreenSpot-Pro) and sets new state-of-the-art results (94.4\% on MMBench-GUI L1-Hard, 60.6\% on OSWorld-G, 47.4\% on WebArena-Lite-v2). These findings underscore the power of data-driven scaling for general-purpose cross-platform CUAs. We release data, models, and code to advance future research: \url{https://github.com/OpenGVLab/ScaleCUA}.
\end{abstract}

\section{Introduction}
Humans are able to interact with digital environments through graphical user interfaces (GUI) to acquire information and accomplish tasks efficiently. The recent advances in Vision-Language Models (VLMs), which exhibit powerful capabilities in visual perception and task planning, have made it increasingly feasible to automate such interactions. 
Consequently, recent research has increasingly focused on computer use agents, also referred to as GUI agents, aiming to autonomously operate desktop, mobile, and web platforms by relying exclusively on visual observations.

% Some computer use agents (CUAs)~\citep{qin2025uitars,anthropic2025claude37,hong2025glm,cua2025} demonstrate strong performance, while they are typically built on closed-source models or inaccessible proprietary datasets.
% % 
% Moreover, it is insufficient to train cross-platform and general-purpose CUAs built upon the existing open-source datasets~\citep{cheng2024seeclick,gou2024navigating,zhang2025guimid,xie2025scalingcomputerusegroundinguser,rawles2023androidinthewild,lu2024gui_odyssey,wu2024osatlas,sun2024osgenesis}.
% % 
% As a result, developing such computer use agent and investigating its associated scaling properties remains a challenging problem.
% % 
% To address these limitations, we make fundamental efforts on two aspects: a) \textbf{a large-scale, cross-platform GUI-centric training corpora}; and b) \textbf{a family of versatile foundation models for robust and general-purpose computer use agents}.

Some computer use agents (CUAs)~\citep{qin2025uitars,anthropic2025claude37,hong2025glm,cua2025} demonstrate strong performance, while they are typically built on closed-source models or inaccessible proprietary datasets.
More fundamentally, effective computer use requires rich in-domain knowledge of software and operational procedures, which remains a substantial gap for current foundation models. Unlike image–text pairs that are abundantly available on the Internet, computer-use data, particularly fine-grained action trajectories, are scarce, costly to collect, expensively to annotate and not naturally archived online.
Furthermore, as software, web pages and operating systems evolve rapidly, existing trajectories face the risk of obsolescence, further limiting their utility. 
% Consequently, the development of cross-platform and general-purpose CUAs is constrained by both the scarcity of high-quality data and the limited adaptability of existing foundation VLMs.
% 
These challenges result in a significant bottleneck for scaling CUAs in both data scale and model generalizability. 
To tackle these limitations,we make significant efforts on two aspects: (a) \textbf{constructing a large-scale, cross-platform GUI-centric training corpus}, and (b) \textbf{developing a family of scalable, versatile foundation models for general-purpose computer use agents}.

We first present a \textit{Cross-Platform Interactive Data Pipeline} composed of two synergistic loops. The \textit{Agent-Environment Interaction Loop} enables automated agents to interact with diverse GUI environments, while the \textit{Agent-Human Hybrid Data Acquisition Loop} integrates expert-annotated trajectories to ensure coverage and quality. The pipeline spans six major platforms, including Windows, macOS, Linux, Android, iOS, and Web, which facilitates the collection of rich screen-state observations, metadata (\eg, A11y Trees, XLM, DOM structures, \etc), and raw trajectories. In this pipeline, we design a unified action space, allowing for more consistent and efficient interaction with diverse real-world environments. 
Leveraging this infrastructure, we curate and annotate a comprehensive training dataset for open computer use dataset, comprising three major task families: (a) \textit{GUI Understanding} with 471K examples covering regional captioning, OCR, and layout comprehension, \etc; (b) \textit{GUI Grounding} with 17.1M training samples supporting more accurate UI element localization; and (c) \textit{Task Completion} with over 15K weak-semantic trajectories and 4K high-level goal-directed trajectories. 
We use advanced VLMs such as GPT-4o and Claude-3.7-Sonnet to annotate collected data and also explore data augmentations to further enhance diversity. 
To this end, we establish a fully open and diverse training dataset.

Building upon this corpus, we train a series of base agent models termed as \textbf{\modelname} with Qwen2.5-VL~\citep{bai2025qwen25vl}.
Our \modelname support three distinct inference paradigms to offer enhanced flexibility and compatibility with various agent frameworks: (a) a \textit{Grounding Mode}, which focuses on precisely locating UI elements based on textual descriptions, allowing for efficient integration with more powerful reasoning planners in modular agent setups, (b) a \textit{Direct Action Mode}, which enables highly efficient task completion by directly generating low-level executable actions without consuming additional tokens for intermediate reasoning and (c) a \textit{Reasoned Action Mode}, which enhances task planning accuracy by first generating a thought process based on current observations and historical context before generating the following action. The unified action space designed in our data construction enables our agents to interface with heterogeneous environments through a standardized control schema.
We conduct extensive empirical studies to investigate how different data sources, diverse training tasks, agent designs, \etc, influence agent performance.
Our findings highlight the benefits of data augmentation, weak semantic trajectories, and general reasoning data for enhancing planning capabilities. 
Moreover, we evaluate various agent paradigms across multiple GUI agent benchmarks and make systematic comparisons between agentic workflows and native models. 
As previous studies~\citep{xu2024aguvis,qin2025uitars,cua2025,anthropic2025claude37} also probe into the important research questions with limited open-sourced training data or under closed conditions with proprietary data, our investigations aim to provide foundational and unified insights for advancing vision-language-based computer automation.

Our contributions are summarized as follows:

1) We curate a cross-platform computer use dataset, collected via an interactive data pipeline that integrates automated agents with human experts. It covers six major platforms (Windows, macOS, Linux, Android, iOS, and Web) and three GUI-centric task domains (\ie, understanding, grounding and task completion), which provide a robust foundation for studying and training universal CUAs.

2) We develop \modelname, a family of robust base agent models that unify perception, reasoning, and action into a single model. It supports flexible inference paradigms including grounding, direct action, and reasoned action, along with a unified action space for seamless cross-platform interaction. 

3)We conduct a comprehensive evaluation spanning understanding, grounding, and end-to-end task completion across several cross-platform online benchmarks.
The experiments not only demonstrate that our agents can achieve state-of-the-art or competitive performance, but also provide insightful directions for developing more powerful computer use agents.

\section{Related Work}
\label{related_work}
\textbf{Vision-Language Models (VLM).}
Vision-Language Models (VLMs) have experienced substantial development in recent years, with progress spanning both proprietary API services~\citep{team2023gemini, team2024gemini, anthropic2024claude35, xai2025grok3, OpenAI2023GPT4TR, hurst2024gpt} and open-source models~\citep{wang2024qwen2vl, bai2025qwen25vl, chen2024internvl, chen2024expanding, zhu2025internvl3, coreteam2025mimovltechnicalreport, team2025gemma, metaai2025llama4}, expanding their capabilities to a wide range of tasks.
Some general VLMs~\citep{team2025kimi, guo2025seed1, bai2025qwen25vl, zhu2025internvl3, coreteam2025mimovltechnicalreport,wang2025internvl35} integrate extensive GUI knowledge during pre-training or SFT stages, and thus acquire explicitly computer use capabilities.  
However, while some works~\citep{hurst2024gpt,anthropic2024claude35,guo2025seed1,zhu2025internvl3,bai2025qwen25vl} demonstrate strong generalization and planning capabilities in vision-language tasks, they still struggle in solving easy computer use tasks. 
Furthermore, the GUI corpora employed to train these models remain largely proprietary. 

\textbf{GUI Agents.} Recent advances in general-purpose vision-language models (VLMs), such as GPT-4o, have enabled the development of modular GUI agents that decompose decision-making and perception into two distinct roles: planning and grounding. These agents typically operate within the \emph{planner–grounder paradigm}~\citep{cheng2024seeclick,hong2024cogagent, lu2024omniparserpurevisionbased,yu2025omniparser,wu2025gui,gou2024navigating,zhang2025guimid,wu2024osatlas,zhou2025guig1}, where a VLM-based planner predicts the next high-level operation and its associated object description, and a specialized GUI grounding model localizes this object on the interface. \cite{yang2024aria} further incorporates action histories into grounding, allowing the grounder to adapt to contextual GUI dynamics.
Inspired by general-purpose \textit{agentic workflows} \citep{wu2023autogen,li2023camel,hong2023metagpt}, these works~\citep{wu2024copilot,liu2025pc,zhao2025cola,Agent-S2,chen2025map} extend this design to GUI environments by coordinating multiple agents across planning, reflection, and memory-augmented decision-making. 
Despite strong performance, agentic workflows typically exhibit high computational latency and significant token consumption, where their performance is fundamentally constrained by the capabilities of underlying VLMs.
In contrast to modular agents, \textit{native computer use agents}~\citep{xu2024aguvis,wu2024osatlas,sun2024osgenesis,qin2025uitars,luo2025gui,liu2025infigui,sun2025scienceboard} integrate planning and grounding into a unified model trained end-to-end. These agents directly predict low-level executable actions from raw visual inputs. For example, \textit{AGUVIS}~\citep{xu2024aguvis} and \textit{UI-TARS}~\citep{qin2025uitars} are trained on extensive task trajectories and demonstrate impressive reasoning and adaptability. 
Generally, native agents offer tighter alignment between perception and action, and modular agents and agentic workflows can also benefit from improvements of native agents. 
In this work, we aim to not only focus on training base models for cross-platform GUI agents, but also commit to open-sourcing all data, accelerating development across the entire domain.

\textbf{GUI Datasets.} 
Open-source datasets have significantly advanced GUI agent development by capturing diverse forms of interaction, visual perception, and instruction-following behaviors across platforms. For \textit{mobile} environments, \textit{RICO}~\citep{deka2017rico} provides 70k+ Android UI screens with gesture traces, while \textit{AITW}~\citep{rawles2023androidinthewild} offers $\sim$715k human demonstrations aligned with 30k natural language commands. Moreover, AitZ~\citep{zhang2024aitz} constructs 18,643 screen-action pairs together with chain-of-action-thought annotations.
In the \textit{web} domain, \textit{MiniWoB}~\citep{Tianlin2017miniwob} simulates diverse web-based tasks with interaction sequences and rewards.
\textit{WebShop}~\citep{shunyu2022webshop} provides language-driven e-commerce tasks with human trajectories over a real product catalog, and \textit{Mind2Web}~\citep{deng2023mind2web} scales this idea to 137 real websites, offering 2,350 long-horizon open-ended tasks.
As for desktop platforms, \cite{xie2024osworld} constructs a 4 million‑example dataset synthesized via interface decomposition, boosting grounding accuracy. 
\cite{he2025efficientagenttrainingcomputer} collect 312 human-annotated trajectories augmented via trajectory boost.
Furthermore, some works explore scalable data generation. For example, \textit{OS-Genesis}~\citep{sun2024osgenesis} synthesizes mobile/web trajectories via exploration, while \textit{AGUVIS}~\citep{xu2024aguvis} curates a large-scale dataset with multimodal grounding and reasoning annotations derived from heterogeneous open-source corpora. 
In addition, due to the high acquisition costs of GUI interaction trajectories, several efforts focus on collecting GUI tutorial datasets to mitigate the scarcity of data.
\textit{META-GUI}\citep{sun2022meta} proposes dialogue-based annotations within tutorials. \textit{TongUI}\citep{zhang2025tongui} collects $\sim$143k task trajectories from instructional sources (e.g., YouTube, WikiHow), linking instructions with screenshots and actions. \textit{GUI-World}~\citep{chen2024guiworld} provides over 12k screen-recorded GUI videos to facilitate temporal understanding of GUI dynamics.
Despite these advances, existing GUI corpora remain limited in both coverage and diversity, particularly for desktop environments. The data scarcity poses significant challenges, especially in UI element grounding and multi-step task planning.

\section{Cross-Platform Interactive Data Pipeline}
\label{sec:data_pipeline}
A critical challenge in developing robust computer use agents is insufficient training data with high quality and diversity in the computer use domain. 
Collecting computer use trajectories is particularly costly and inefficient, primarily due to the dynamic nature of environments and their frequent dependency on task-specific resource files, which severely limit the feasibility of fully automated data collection pipelines. In this section, we elaborate on the pipeline of data acquisition and annotation for training computer use agents.

\subsection{Overview}

\begin{figure}[th]
    \centering
    \includegraphics[width=0.9\linewidth]{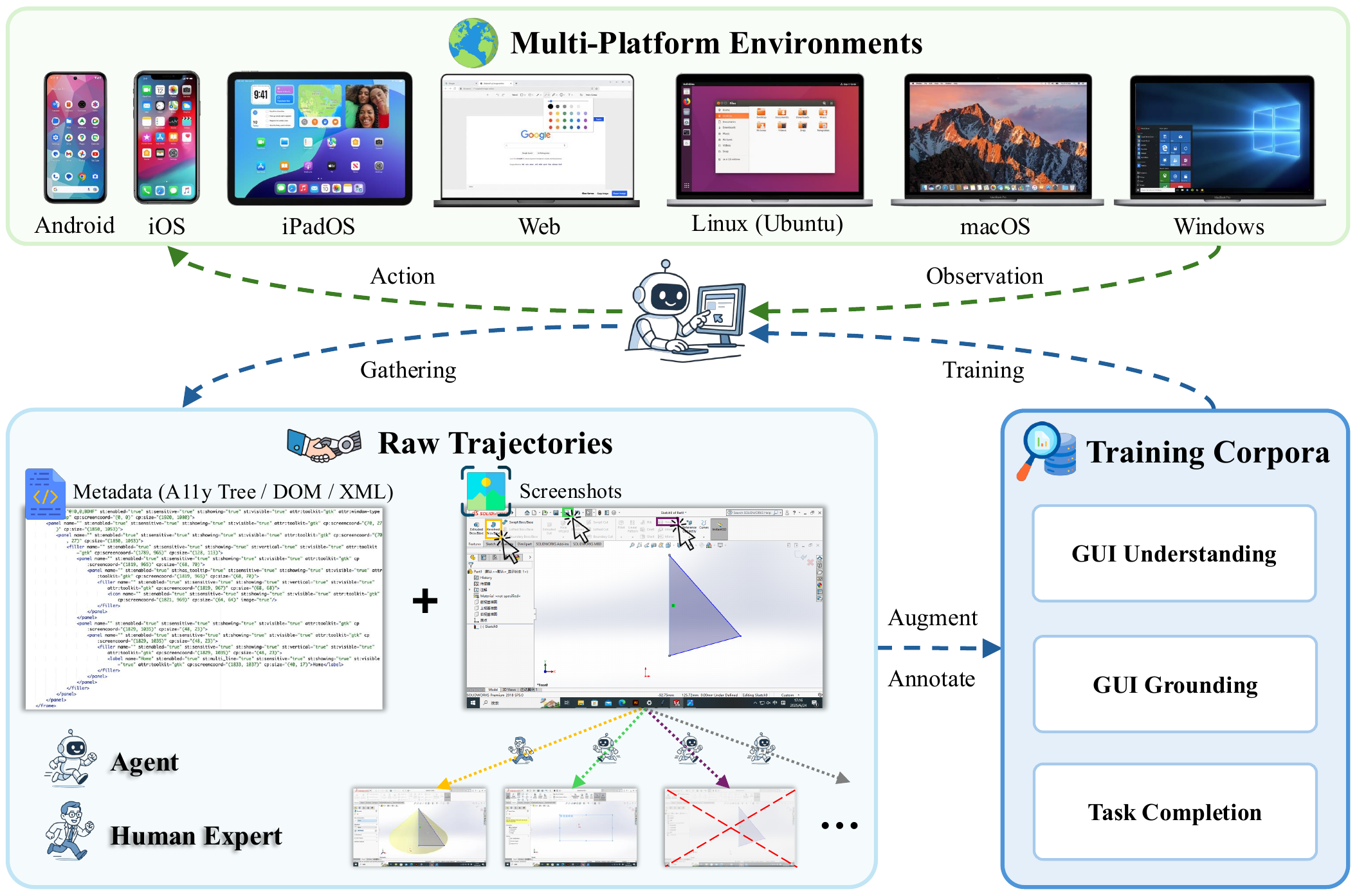}
    \small
    \caption{\textbf{Cross-Platform Interactive Data Pipeline.} Our Pipeline consists of two synergistic loops: (1) the \textbf{Agent-Environment Interaction Loop}, where agents interact with multi-platform GUI environments (including desktop, mobile, and web) via observation and action; and (2) the \textbf{Agent-Human Hybrid Data Acquisition Loop}, where both autonomous agents and human experts contribute to collecting raw trajectories, including screenshots and structural metadata. The resulting trajectories are then annotated and transformed into training corpora for tasks such as GUI understanding, GUI grounding, and sequential action modeling.
}
    \label{fig:data_pipeline}
\vspace{-1em}
\end{figure}

Existing datasets in the computer use domain predominantly fall into two categories: manual trajectory collection and automated search-based exploration. Manual collection~\citep{zhang2024aitz,rawles2023androidinthewild,deng2023mind2web,lu2024gui_odyssey}, although capable of producing high-quality trajectories, is prohibitively expensive, labor-intensive, and inherently difficult to scale. Conversely, the automated exploration method~\citep{sun2024osgenesis}, while scalable, generates trajectories with considerable noise. Therefore, neither strategy alone can effectively provide the quality and diversity required for training versatile GUI agents.

Motivated by these limitations, we propose a \textit{Cross-Platform Interactive Data Pipeline} that strategically integrates agents and humans into data acquisition. As illustrated in Fig.~\ref{fig:data_pipeline}, the pipeline operates in two synergistic loops: the Agent-Environment Interaction Loop and the Agent-Human Hybrid Data Acquisition Loop. The former involves agents interacting with multi-platform GUI environments, while the latter integrates trajectories collected by both autonomous agents and human experts. These trajectories, comprising screenshots and structural metadata, are then annotated and transformed into training corpora for tasks like GUI understanding, grounding, and sequential action modeling. This dual-loop framework ensures comprehensive coverage of GUI states while maintaining high data quality, laying a robust foundation for training generalizable computer use agents.

\textbf{Agent-Environment Interaction Loop}. Initially, we establish a unified interface across heterogeneous platforms, including Windows, Ubuntu, macOS, Web browsers, Android, and iOS, to standardize observation acquisition and action execution.
This unified abstraction facilitates closed-loop data collection and provides a flexible interface compatible with various agent architectures, significantly enhancing the scalability and efficiency of online evaluations.
To construct our dual-mode data collection framework, we investigate platform-specific methods for extracting GUI metadata. For desktop environments (Windows, Ubuntu, macOS), interface layouts and UI element coordinates are generally accessible via Accessibility Trees (A11y Trees). Web platforms provide element information through Document Object Model (DOM) structures rendered by browsers. For Android applications, UI element locations are available through parsed XML layout files. 
However, we observe that A11y trees in the desktop platforms often contain missing attributes, and iOS/iPadOS systems restrict access to layout information. For these cases, we employ OmniParser~\citep{yu2025omniparser} to estimate bounding boxes of UI elements, balancing noise tolerance with efficiency gains.

\textbf{Agent-Human Hybrid Data Acquisition Loop}. After unifying the interface for interacting with diverse environments, the next natural question is how to deploy automated agents to collect interaction trajectories while ensuring maximal coverage of GUI states. We experimented with two distinct exploration strategies. The first employed VLM-driven agents, \ie, GPT‑4o, Claude, Gemini, \etc, to navigate the interface. However, due to inherent biases in these models, the resulting trajectories lacked sufficient diversity. In contrast, the second strategy utilized a rule‑driven random‑walk agent. Our rule-driven agents perform exploration using a depth-first search (DFS) strategy, where an action is randomly selected from the available action space at each step. To improve efficiency, heuristic pruning is applied to reduce redundant or uninformative branches during search. The detailed design of this strategy is provided in the~\ref{sec:the_details_of_data_curation}. After deduplication, this simple method achieves substantially broader GUI coverage. Although the resulting trajectories often lack clear high‑level goals at the macro scale, many of these sub‑sequences still provide valuable supervised signals for agent model learning.
Despite the efficiency gains from automation, extensive quality assessments have revealed considerable noise in both system-derived metadata and vision-based bounding box predictions. Therefore, to mitigate noise and ensure high-quality annotations, we complement automated collection with expert-curated trajectories. 
To streamline expert, we build a unified cross-platform data acquisition system where the frontend interaction interface is decoupled from the backend environment, enabling annotators to efficiently switch between platforms. Experts are asked to complete domain-specific goals spanning productivity, entertainment, and system operations.
Combining both agent-driven and expert-driven styles yields a substantial corpus of raw GUI trajectories, which comprises extensive screenshots along with corresponding metadata. Subsequently, we apply advanced VLMs (e.g., GPT-4o~\citep{hurst2024gpt} and Claude-3.5~\citep{anthropic2024claude35} for data annotation, categorizing the corpus into three primary GUI-centric tasks: Understanding, Grounding, and Task Completion. This comprehensive dataset ultimately provides a robust foundation for training generalizable computer use agents.

\textbf{Unified Action Space}. As shown in Table~\ref{tab:action_space}, we establish a cross-platform action space covering three major environments: Desktop, Browser, and Mobile. This unified interface enables consistent behavior modeling and facilitates training across heterogeneous platforms. It includes a core subset of universal operations (\texttt{click}, \texttt{write}, \etc) and platform-specific actions (\texttt{long\_press}, \texttt{open\_app}, \etc\ for mobile) are essential for mobile environments.
A unified action space allows agents to generalize core behaviors while retaining the flexibility to exploit environment-specific actions. This design also simplifies data annotation and downstream policy learning, as each action is expressed in a platform-aware but semantically consistent manner. More details can refer to Sec.~\ref{sec:action_space}.

\subsection{Task-Oriented Data Curation}
Combining both agent-driven and expert-driven styles yields a substantial corpus of raw GUI trajectories, which comprises extensive screenshots along with corresponding metadata. Subsequently, we apply advanced VLMs (e.g., GPT-4o~\citep{hurst2024gpt} and Claude-3.7~\citep{anthropic2025claude37}) for data annotation, categorizing the corpus into three primary tasks: GUI Understanding, GUI Grounding, and Annotated Trajectories. 
Based on the annotated GUI-centric corpora, we aim to build an Open-sourced Computer Use Dataset (\textit{\ourdataset}).
This comprehensive dataset ultimately provides a robust foundation for training generalizable computer use agents.

\begin{table}[t]
\centering
\caption{Datasets comparisons on computer-use datasets in terms of platform coverage, task types 
(Understanding, Grounding and Trajectories), and collection methods.}
\small
\label{tab:data_comparision}
\resizebox{\linewidth}{!}{
\begin{tabular}{lccc ccc ccc c}
    \toprule
\multirow{2}{*}{\textbf{Data source}}  & \multicolumn{3}{c}{\textbf{Platform}} & \textbf{Understanding} & \textbf{Grounding} & \multicolumn{2}{c}{\textbf{Trajectories}} & \multirow{2}{*}{\makecell{\textbf{Collection} \\ \textbf{Method}}} \\ 
\cmidrule(lr){2-4} \cmidrule(lr){5-5} \cmidrule(lr){6-6} \cmidrule(lr){7-8}
 & Desktop & Mobile & Web & \small \#Samples & \small \#Elements & \small \#Samples & \small Avg. Steps & \\
\midrule
SeeClick~\citeyearpar{cheng2024seeclick}  & \xmark & \cmark & \cmark & -- & 763K  & -- & -- & Auto \\
GUIEnv~\citeyearpar{chen2024guicourse}    & \xmark & \xmark & \cmark & -- & 10.7M & --  & -- & Auto \\
WebUI~\citeyearpar{wu2023webui}           & \xmark & \xmark & \cmark & -- & 400K & --  & -- & Auto \\
Widget Captioning~\citeyearpar{li2020widget}  & \xmark & \cmark & \xmark & -- & 163K & --  & -- & Human \\
RicoSCA~\citeyearpar{li2020mapping}       & \xmark & \cmark & \xmark & -- & 178K & --  & -- & Auto \\
RICO~\citeyearpar{deka2017rico}           & \xmark & \cmark & \xmark & -- & 72K  & --  & -- & Hybrid \\
OmniACT~\citeyearpar{kapoor2024omniact}   & \cmark & \xmark & \cmark & -- & 9.8K & --  & -- & Auto \\
MM-Mind2Web~\citeyearpar{zheng2024seeact} & \xmark & \xmark & \cmark & -- & -- & 1K & 7.7 & Human \\
GUIAct~\citeyearpar{chen2024guicourse}    & \xmark & \xmark & \cmark & -- & 67K  & 5.7K & 6.7 & Auto \\
AitZ~\citeyearpar{zhang2024aitz}          & \xmark & \cmark & \xmark & -- & -- & 2.5K & 6.0 & Human \\
AndroidControl~\citeyearpar{li2024effectsdatascalecomputer} & \xmark & \cmark & \xmark & -- & -- & 13.6K & 5.5 & Human \\
GUI Odyssey~\citeyearpar{lu2024gui_odyssey} & \xmark & \cmark & \xmark & -- & -- & 7.7K & 15.3 & Human \\
AMEX~\citeyearpar{chai2024amex}           & \xmark & \cmark & \xmark & -- & --  & 3.0K & 11.9 & Human \\
AitW~\citeyearpar{rawles2023androidinthewild} & \xmark & \cmark & \xmark & -- & -- & 2.3K & 8.1 & Human \\ 
OS-Atlas~\citeyearpar{wu2024osatlas}      & \xmark & \cmark & \cmark & -- & 13.6M & -- & -- & Auto \\
OS-Genesis~\citeyearpar{sun2024osgenesis} & \xmark & \cmark & \cmark & -- & -- & 2.5K & 6.4 & Auto \\
JEDI~\citeyearpar{xie2025scalingcomputerusegroundinguser} & \cmark & \xmark & \xmark & 877K & 3.1M & -- & -- & Auto \\
AgentNet~\citeyearpar{wang2025opencuaopenfoundationscomputeruse} & \cmark & \xmark & \xmark & -- & -- & 22K & -- & Auto \\
\midrule
Ours & \cmark & \cmark & \cmark & 471K & 17.1M & 19.0K & 9.0 & Hybrid \\
    \bottomrule
\end{tabular}
}
\vspace{-1.5em}
\end{table}

\textbf{GUI Understanding.}
We design a set of tasks to facilitate fine-grained perception and reasoning over GUIs. At the element level, the model learns to describe visual appearance, extract textual content (Referring OCR), understand spatial layout, infer functional semantics, and anticipate user intent. At the screenshot level, we include Interface Captioning to summarize global layout and content, and Screen Transition Captioning to describe state changes over time. These tasks serve as a foundation for training agents with strong interface comprehension.

\textbf{GUI Grounding.}
Grounding tasks enable agents to map natural language instructions to corresponding interface regions. We formulate three types of grounding supervision: (1) point grounding identifies precise click locations, (2) bounding box grounding localizes regions for region-based operations, and (3) action grounding connects spatial targets with low-level commands. All grounding tasks are derived from previously annotated understanding data and further expanded via LLM-driven augmentation to enhance coverage and generalization.

\textbf{Weak-Semantic Trajectories.}
Rule-based agents perform depth-first exploration to cover diverse GUI states. Although these trajectories lack explicit task intent, we segment them into sub-trajectories based on screen similarity, yielding weakly semantic interaction flows. These sequences capture common GUI patterns with low-cost supervision for learning general interface navigation.

\textbf{Human-Curated Trajectories.}
To complement automated data, we employ expert operators to collect realistic, goal-directed demonstrations. Tasks are selected from daily, entertainment, and productivity domains. To this end, we develop a unified cross-platform system that isolates the environment for each platform. These curated trajectories provide high-quality supervision for reasoning and planning, and serve as reference signals to align model behavior with human intent.

\textbf{Annotation and Augmentation.}
Each screenshot is paired with metadata that includes UI layout structures. We deduplicate frequent states and sample diverse interactive elements, which are annotated using VLMs to generate appearance, position, and function descriptions. These annotations support QA and grounding tasks. To increase data variety, we apply augmentation techniques including element cropping, synthetic resolution scaling, and background substitution. Trajectory-level annotations are enriched through reasoning-based prompts that generate instructions, thought chains, and alternative actions. All prompt templates are detailed in the~\ref{sec:prompt_template}.

\subsection{Data Statistics}

\begin{wrapfigure}{r}{0.45\textwidth}
    \vspace{-20pt} 
    \centering
    \includegraphics[width=0.85\linewidth]{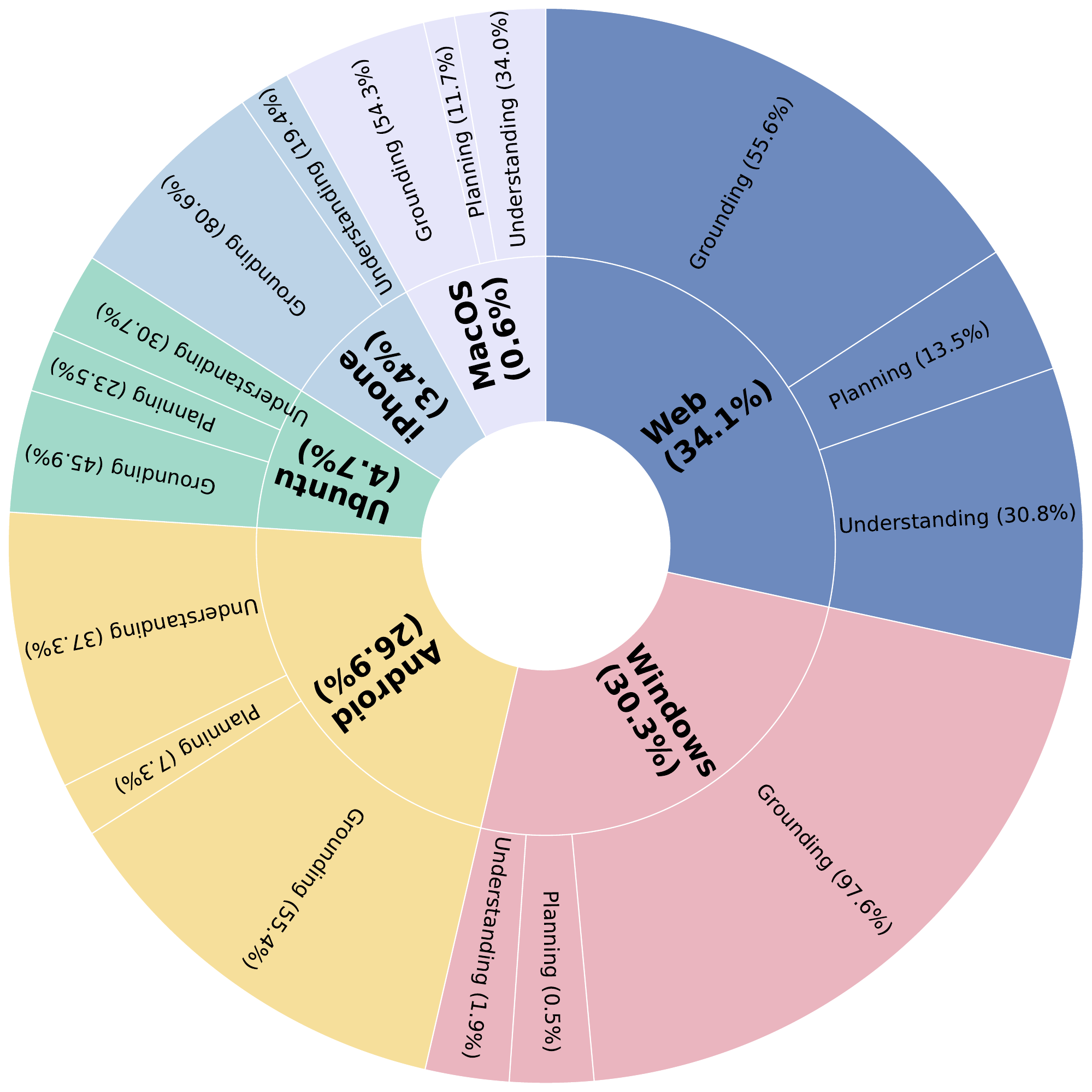}
    \small
    \caption{\textbf{Data distribution of our dataset}.}
    \label{fig:data_distribution}
    \vspace{-10pt}
\end{wrapfigure}
Our collected corpus spans three major task domains: \textit{Understanding}, \textit{Grounding}, and \textit{Task Planning}. As summarized in Table~\ref{tab:data_comparision}, the dataset contains $471$K understanding examples across diverse tasks such as VQA, OCR, layout understanding, and screen transition reasoning, \etc 
For grounding, we curate over $17.1$M annotations with multiple formats (bounding box, point, and action), enabling fine-grained alignment between textual instructions and GUI elements. In addition, we provide $19$K trajectories for task planning, covering both human-curated and enhanced trajectories, with an average of $9$ steps per trajectory.  

Fig.~\ref{fig:data_distribution} illustrates the distribution across platforms, where desktop tasks dominate due to broader application coverage and richer interface diversity, while mobile and web environments are also substantially represented. Compared with existing resources, our dataset (Table~\ref{tab:data_comparision}) is the most comprehensive, covering three task families across desktop (Windows, Linux, macOS), mobile, and web. Prior datasets ~\citep{deka2017rico,chen2024guicourse,wu2024osatlas} typically focus on a single task domain, whereas our dataset integrates heterogeneous supervision signals into a unified format.  

\textbf{Discussions}. This scale and diversity enable a balanced evaluation of computer use agents under both perception and action paradigms. By leveraging a dual-loop collection pipeline, we ensure coverage of low-level element recognition, mid-level grounding, and high-level task planning. Compared with current works~\citep{sun2024osgenesis,wu2024osatlas,zhang2024aitz,rawles2023androidinthewild}, we explore more diverse data collection strategies (human experts, VLM-driven agents, and rule-driven agents) and cover a broader range of platforms (desktop, mobile, and web). Specifically, as for the random-walk agent, we designed a more efficient algorithm through extensive experimentation and iterative improvements, significantly enhancing both data collection efficiency and GUI coverage. With this pipeline, we have collected over 2M raw screenshots across multiple platforms. 
Furthermore, we make substantial efforts on training robust computer use agents by constructing diverse proxy tasks spanning understanding, grounding and planning. 

\section{Scaling up Computer Use Agents}
\label{sec:method}

\subsection{Task Definition}
CUAs have emerged for programmatic automation. Early approaches, such as \textit{Selenium}\footnote{https://github.com/SeleniumHQ/selenium}, which leverages DOM parsing for web automation, and \textit{AutoHotkey}\footnote{https://github.com/AutoHotkey/AutoHotkey}, which scripts deterministic desktop interactions, enabled precise but constrained interface control. Large-scale VLMs have significantly broadened this paradigm, allowing agents~\citep{cheng2024seeclick,wang2024mobile,zhang2025appagent} to achieve pixel-level perception and interaction on graphical user interfaces.

The interaction between the agent and environment at time step can be simply defined as follows:
\begin{equation}
a_t = \pi_{\theta}(task, o_t, h_{<t}), \quad o_{t+1} = \mathcal{E}(a_t),
\end{equation}
where $\pi$ denotes the agent model parameterized by $\theta$, and $\mathcal{E}$ represents the environment, such as virtual machines or Docker containers. The $task$ denotes the task instruction. The observation $o$ encompasses elements such as raw screen pixels, accessibility trees, or DOM data. The history \(h_{<t} = \{(a_0, o_0), \dots, (a_{t-1}, o_{t-1})\}\) provides context for agent's decision-making process. Some works~\citep{sun2024osgenesis,xu2024aguvis} choose to generate natural language descriptions for $(a_{i}, o_{i})$ as history, as it can save a large amount of inference cost budget. Each action specifies an operation with corresponding arguments, as detailed in Table~\ref{tab:action_space}, which is then executed in the environment.

In this work, we persist in modeling computer use tasks based on screenshots. This paradigm aligns with human behavior and effectively avoids interference from noisy accessibility Tree and DOM data.

\subsection{Base Agent Models}
\label{sec:agentmode}
\label{sec:basemodels}
\begin{figure}[tb!]
    \centering
    \includegraphics[width=0.9\linewidth]{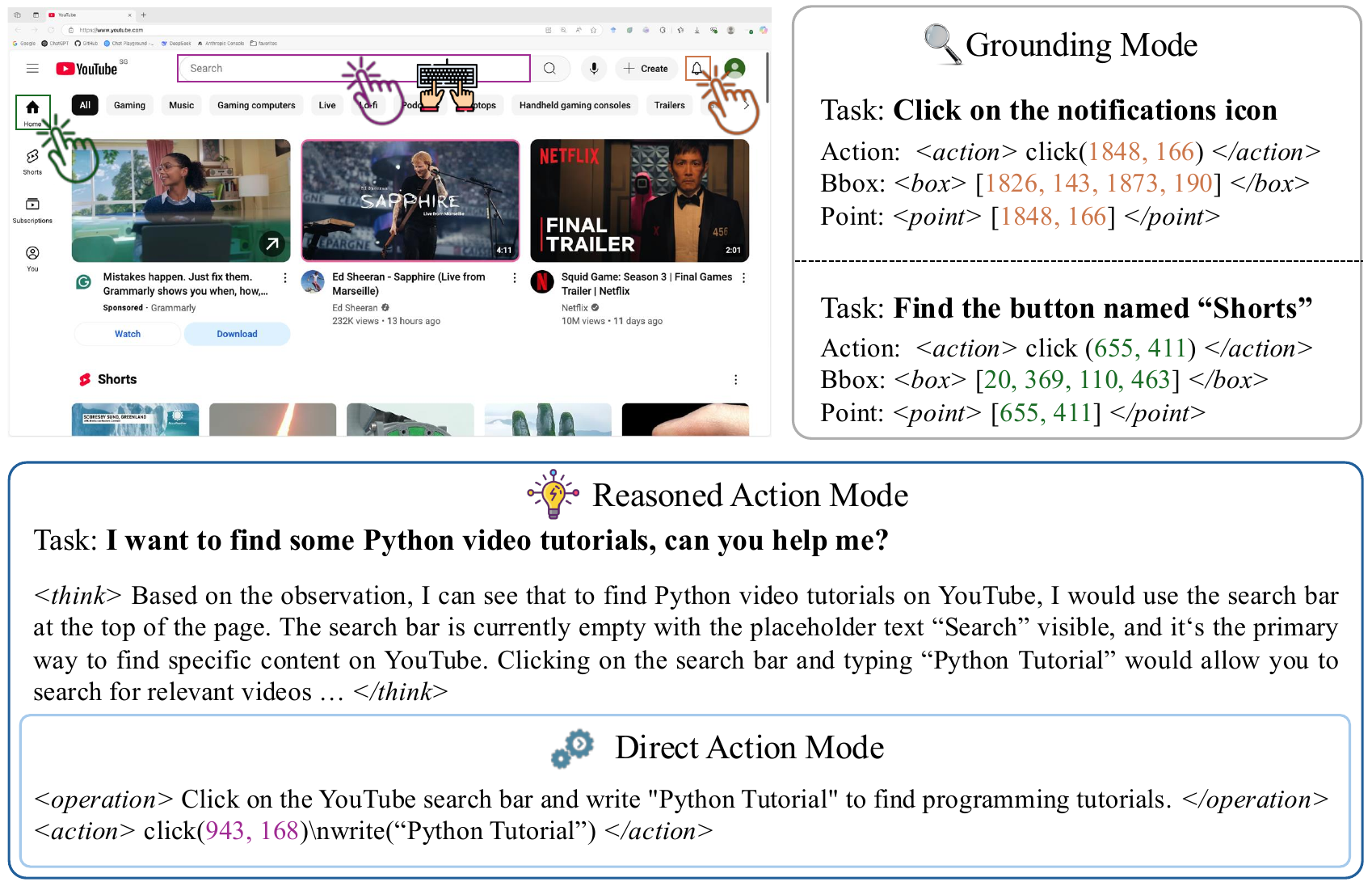}
    \small
    \caption{\textbf{Three Inference Paradigms of Our Computer Use Agents}: (1) \textbf{Grounding Mode}, which focuses on identifying target UI elements with their spatial coordinates and bounding boxes; (2) \textbf{Direct Action Mode}, where the agent solely generates executable actions based on current observations and instructions; and (3) \textbf{Reasoned Action Mode}, where the agent first generates a chain-of-thought rationale before producing structured actions. These modes enable varying levels of functionality for computer use agents to complete tasks.
}
    \label{fig:infer_mode}
\vspace{-2em}
\end{figure}
To train robust computer use agents, we leverage the state-of-the-art Vision-Language Model, Qwen-2.5-VL~\citep{bai2025qwen25vl}, as our base model. It is selected for its strong multimodal understanding capabilities and scalability across different GUI platforms. 
As illustrated in Fig.~\ref{fig:infer_mode}, the models operate in three modes:
1) \textit{Grounding Mode}, where the model focuses on UI grounding from the current observation and instruction. 
2) \textit{Direct Action Mode}, where the model directly generates low-level instructions with executable actions from the current observation and historical context.
3) \textit{Reasoned Action Mode}, where the model first produces a reasoning process before generating the action. The outputs generated by models are formatted in XML tags.

\textbf{Grounding Mode} represents the model only focuses on the UI element localization. 
It leverages visual observation to guide the model toward generating either spatial points, bounding boxes, or coordinate-referenced actions (e.g., \textit{click}, \textit{doubleClick},\textit{moveTo}, \etc), facilitating precise grounding. In this mode, the model only takes screenshots and instructions as inputs. 
When deployed, this paradigm enables the powerful general-purpose VLMs (e.g., GPT-4o) acting as planners and our model functioning as a grounder, to complete computer use tasks in a modular paradigm. 

\textbf{Direct Action Mode} requires the model to efficiently predict executable actions directly, without generating any intermediate reasoning steps or extraneous outputs. Given a task instruction, the current screen observation, and the interaction history, the model outputs both a low-level instruction and an executable action. The low-level instruction provides a concise natural language description that explains the intended operation (\eg, “Open the settings menu”), while the action within \texttt{<action>} tags corresponds to a concrete low-level control command (e.g., \texttt{click(x=213, y=234)}). The generated low-level instruction is appended to the dialogue history and serves as human-interpretable descriptions that serve as historical context for subsequent steps, allowing the model to maintain context over multi-step interactions. Notably, the low-level instruction and actions are wrapped in \texttt{<operation>...</operation>} and \texttt{<action>...</action>}.
This mode enables rapid task execution by eliminating the need for explicit intermediate thought or reasoning steps before action. It supports a streamlined perception-action loop where the model continuously observes the environment and emits grounded actions. By integrating minimal but informative linguistic feedback into the interaction history, the agent can sustain coherent behavior while reducing inference latency. Empirically, this design is able to complete complex GUI tasks efficiently, particularly in real-time tasks (\eg, online games) where fast and reactive behavior is critical. 

\textbf{Reasoned Action Mode} refers to a chain-of-thought process before generating actions. In this mode, the input to the model remains consistent with the Direct Action Mode.
The reasoning trace explicitly outlines the agent’s interpretation of the task objective and the rationale behind selecting a particular action, often referencing visual cues from the screen and correlating them with the user instruction. For example, the model might reason, ``The task is to adjust brightness. I see a settings icon on the top right, which likely leads to display settings,'' followed by a proposed action such as \texttt{click(x=1040, y=75)}. This intermediate thought process wrapped in \texttt{<think>...</think>} tags enhances action reliability, especially in ambiguous or multi-step tasks where context comprehension and logical grounding are critical.
Although this mode introduces higher computational cost and latency due to the extended generation process, it significantly improves the quality and interpretability of the agent’s behavior. The explicit reasoning provides human-understandable justifications for each step, enabling fine-grained analysis of decision patterns and failure cases, which can be used to iteratively refine the model’s reasoning heuristics. 
By incorporating structured thought before action, it empowers agents to perform complex tasks with transparency, making it particularly suitable for long-horizon interaction scenarios where precision and explainability are essential.

\subsection{Training Recipes}
We adopt different configurations for each ScaleCUA model according to model size and available hardware resources. For the ScaleCUA-3B model, we use a mini-batch size of 4 with gradient accumulation of 1, trained on 128 A100 GPUs. The ScaleCUA-7B model follows the same batch configuration (mini-batch size 2, gradient accumulation 2) and GPU setup. For the ScaleCUA-32B model, due to its significantly larger parameter count, we reduce the mini-batch size to 2 while applying gradient accumulation of 2 to maintain an effective batch size of 4. Training is conducted on 128 H200 GPUs to accommodate the higher memory footprint. We use a learning rate of $1.0\times10^{-5}$ and the maximum token length of 40960 for all models.

Across all models, we experiment with varying proportions of general-purpose multimodal data in the training corpus. Specifically, the 3B, 7B, and 32B models are trained with general data ratios of 25\%, 50\%, and 75\%, respectively. In the experiments, we observe that when general-purpose data accounts for 25\% of the GUI-specific training corpus, the 3B model achieves strong performance on GUI agent benchmarks after training. Considering the larger models are able to memorize more data samples, we raise this ratio to 50\% for the 7B model and further to 75\% for the 32B model.

\section{Experiments}
\label{sec:exp}

\subsection{Evaluation setup}
\begin{table*}[!t]
  \centering
  \caption{Results on MMBench-GUI L1 (GUI Content Understanding)~\citep{wang2025mmbenchgui}.}
  \small 
  \resizebox{0.85\textwidth}{!}{%
  \begin{tabularx}{\textwidth}{l*{6}{>{\centering\arraybackslash}X}>{\centering\arraybackslash}X}
    \toprule
    \textbf{Model} & \textbf{Windows} & \textbf{MacOS} & \textbf{Linux} & \textbf{iOS} & \textbf{Android} & \textbf{Web} & \textbf{Overall} \\
    \midrule
     \rowcolor{gray!20} \multicolumn{8}{c}{\textcolor{gray}{Easy Level}} \\
    GPT-4o~\citeyearpar{hurst2024gpt} & 62.5 & 67.9 & 62.4 & 58.5 & 56.4 & 58.5 & 60.2  \\
    Claude-3.5~\citeyearpar{anthropic2024claude35} &  41.3  &   50.0 & 41.6 & 42.0 & 39.0 & 41.8 &  41.5 \\
    Claude-3.7~\citeyearpar{anthropic2025claude37} & 34.7 &  49.1  & 39.4 & 42.8 & 37.5 & 40.8 &  39.1 \\
    Qwen-Max-VL~\citeyearpar{Qwen-VL} & 69.1   & 72.5   & 69.9 & 70.8 & 63.1 & 69.5 & 68.2 \\
    Qwen2.5-VL-72B~\citeyearpar{bai2025qwen25vl} &  65.9  & 75.2   & 73.0 & 67.2 & 58.1 & 72.1 & 67.0 \\
    UI-TARS-72B-DPO~\citeyearpar{qin2025uitars} & 41.6   &  28.5  & 35.2 & 31.1 & 52.3 & 35.3 &  40.2 \\
    InternVL3-72B~\citeyearpar{zhu2025internvl3} &  74.7  & 78.7 & 79.2 & 83.6 & 80.1 & 81.2 & 79.2 \\  
    GUI-Owl-7B~\citeyearpar{ye2025mobile} & 83.0 & 84.5 & 85.6 & 82.6 & 83.3 & 88.1 & 84.5 \\
    GUI-Owl-32B~\citeyearpar{ye2025mobile} & \textbf{93.7} & 89.3 & 93.3 & \textbf{95.7} & \underline{90.5} & \textbf{94.1} & \textbf{92.8} \\
\hline
    \modelname{3}  & 86.4 & 83.5 & 79.9 & 85.4  & 80.3 & 87.4 & 83.6\\
    \modelname{7}  & 89.5 & 86.9 & 89.1 & 86.2 & 87.0 & 90.1 & 88.4\\
    \modelname{32} & \underline{93.4} & \textbf{91.7} & \textbf{94.3} & \underline{93.1} & \textbf{90.5} & \underline{92.3} & \underline{92.5} \\

     \rowcolor{gray!20} \multicolumn{8}{c}{\textcolor{gray}{Medium Level}} \\
    GPT-4o~\citeyearpar{hurst2024gpt} &  56.3  & 63.1   & 59.7 & 54.1 & 57.7 & 55.0 & 57.2  \\
    Claude-3.5~\citeyearpar{anthropic2024claude35} &  39.3  &  47.6  & 46.0 & 44.6 & 42.0 & 34.33 & 41.26  \\
    Claude-3.7~\citeyearpar{anthropic2025claude37} &  39.3  & 39.2   & 42.3 & 39.5 & 36.1 & 36.2  & 38.4  \\
    Qwen-Max-VL~\citeyearpar{Qwen-VL} &  63.4  & 73.9  & 66.9 & 68.0 & 63.7 & 64.6 & 65.4 \\
    Qwen2.5-VL-72B~\citeyearpar{bai2025qwen25vl} & 66.3 &  72.7  & 72.6 & 59.3 & 66.2 & 68.2 & 67.5 \\
    UI-TARS-72B-DPO~\citeyearpar{qin2025uitars} & 38.8 & 41.6 & 37.1 & 41.7 & 54.7 & 31.6 & 41.8  \\
    InternVL3-72B~\citeyearpar{zhu2025internvl3} & 71.5  & 78.6   & 79.9 & 78.4 & 81.4 & 78.7 & 77.9 \\ 
    GUI-Owl-7B~\citeyearpar{ye2025mobile} &88.9 & 88.1 & 91.2 & 84.4 & 85.3 & 83.6 & 86.9\\
    GUI-Owl-32B~\citeyearpar{ye2025mobile} & \underline{94.1} & 84.5 & \underline{95.9} & \underline{87.8} & \textbf{92.8} & \textbf{88.6} & \underline{91.7} \\
    \hline
    \modelname{3} & 91.8 & 78.5 & 88.7 & 74.8 & 88.6 & 79.5 & 85.6 \\
    \modelname{7}  & 93.6 & \textbf{91.7} & 93.4 & 84.3 & 89.6 & 85.8 & 90.1 \\
    \modelname{32}  & \textbf{95.1} & \underline{89.4} & \textbf{96.3} & \textbf{92.2} & \underline{92.6} & \underline{87.2} & \textbf{92.5}\\
    
     \rowcolor{gray!20} \multicolumn{8}{c}{\textcolor{gray}{Hard Level}} \\
    GPT-4o~\citeyearpar{hurst2024gpt} &  60.7  &  60.4  & 52.4 & 45.3 & 50.9 & 50.8 &  53.5 \\
    Claude-3.5~\citeyearpar{anthropic2024claude35} & 37.40   &  42.70  & 34.07 & 40.86 & 36.96 & 38.1 & 37.6  \\
    Claude-3.7~\citeyearpar{anthropic2025claude37} &  33.0  &  34.5  & 32.0 & 39.2 & 37.0 & 38.9 &  35.7 \\
    Qwen-Max-VL~\citeyearpar{Qwen-VL} &  66.6  &  67.6  & 65.8 & 60.2 & 58.8 & 65.3 & 63.7 \\
    Qwen2.5-VL-72B~\citeyearpar{bai2025qwen25vl} &  70.7  &  68.9  & 71.0 & 57.6 & 53.9 & 68.1 & 64.6 \\
    UI-TARS-72B-DPO~\citeyearpar{qin2025uitars} &  31.5  &  35.9  & 24.2 & 36.3 & 58.1 & 19.9 & 35.8  \\
    InternVL3-72B~\citeyearpar{zhu2025internvl3} & 75.1  & 77.4   & 76.2 & 70.4 & 75.7 & 78.1 & 75.7 \\
    GUI-Owl-7B~\citeyearpar{ye2025mobile} & 87.8 & \underline{96.4} & 94.3 & 87.8 & 88.9 & \textbf{94.1} & 90.9 \\
    GUI-Owl-32B~\citeyearpar{ye2025mobile} & \textbf{93.3} & 95.2 & \underline{95.9} & \underline{92.2} & \textbf{95.4} & 92.69 & \underline{94.2} \\  \hline
    \modelname{3}  & 92.3 & 89.4 & 93.8 & 85.3 & 88.3 & 88.6 & 89.9 \\
    \modelname{7}  & 91.9 & 91.9  & 94.9 & 89.6 & 92.9 & 91.4 & 92.3\\
    \modelname{32} & \underline{93.0} & \textbf{96.5} & \textbf{96.4} & \textbf{93.1} & \underline{94.5} & \underline{94.0} & \textbf{94.4} \\
    
    \bottomrule
  \end{tabularx}
  }
  \label{tab:mmbengui-l1}
\end{table*}
To comprehensively assess the capabilities of our computer use agents, the evaluations span understanding, grounding, and agent capability of task completion. Please note that \textit{our models are evaluated with pure visual observation on all benchmarks.}

To validate \textbf{GUI understanding performance}, we resort to MMBench-GUI L1~\citep{wang2025mmbenchgui}, which evaluates an agent’s fine-grained perception and reasoning about interface content across diverse platforms by posing multiple-choice questions. It can serve as a standardized test of basic GUI understanding without requiring explicit action execution.

To quantify the \textbf{GUI grounding performance}, we conduct structured evaluations on \textit{ScreenSpot-v2}~\citep{wu2024osatlas} \textit{ScreenSpot-Pro}~\citep{li2025screenspotpro}, and \textit{OSWorld-G}~\citep{xie2025scalingcomputerusegroundinguser}. These datasets are designed for benchmarking spatial localization for UI elements. ScreenSpot-v2 emphasizes cross-platform grounding accuracy for both textual and visual targets, while ScreenSpot-Pro categorizes applications by usage domain, enabling fine-grained analysis of grounding performance across development, creative, scientific, and productivity contexts. The screenshots in OSWorld-G are sampled by previous models on OSWorld, which facilitates a more precise analysis of whether the performance bottleneck of the current model in OSWorld stems from deficiencies in grounding capability or planning ability. 
By default, we only use 2K resolution in ScreenSpot-Pro and 1080p for other grounding benchmarks.

For \textbf{end-to-end agent evaluations}, we test our models on a suite of offline and online GUI interaction benchmarks. Offline evaluations are conducted on the \textit{AndroidControl} dataset, which provides fine-grained control benchmarks with both low and high task complexity. For online settings, we deploy agents in dynamic real-time environments including \textit{OSWorld}~\citep{xie2024osworld}, \textit{WindowAgentArena}~\citep{bonatti2024windows}, \textit{MacOSArena}~\citep{wang2025mmbenchgui}, \textit{AndroidWorld}~\citep{rawles2024androidworld}, and \textit{WebArena-Lite-v2}. We set 50-step budget for online benchmarks in ablation studies if not stated. 
These platforms cover desktop, mobile, and web environments, enabling a thorough assessment of the agents’ capabilities in realistic and platform-specific task completion.
It's worth noting that WebArena-Lite~\citep{liu2024visualagentbench} is not originally designed for pure visual evaluation. Additionally, several tasks are either impossible to complete or even require actions beyond the desktop environment. To address these limitations, we upgrade it into \textit{WebArena-Lite-v2}, providing a more suitable framework for evaluating visual-based web agents. Please refer to~\ref{appendix:web_online_evaluation} for details.

In addition, we also validate the \textbf{general capabilities} of our models on four widely used vision-language benchmarks: \textit{MMMU}~\citep{yue2024mmmu}, \textit{MathVista}~\citep{lu2023mathvista}, \textit{OCRBench}~\citep{liu2024ocrbench} and \textit{RealWorldQA}~\citep{xai-grok15v-2024}. These benchmarks collectively cover a wide spectrum of vision-language tasks, including comprehensive multimodal understanding, mathematical reasoning, OCR-based visual QA and real-world visual QA. 
These tasks are related to the computer use agent performance, as OCR proficiency forms the foundation for effective grounding, while multimodal reasoning provides the necessary prerequisites for sophisticated planning capabilities. 

In addition, to ensure scalability and platform adaptability for online evaluations, we use vLLM\footnote{https://github.com/vllm-project/vllm}~\citep{kwon2023efficient} to deploy QwenVL models. 
This multi-faceted evaluation setup allows us to systematically analyze each component of the agent framework—from perception and grounding to planning and execution—across diverse datasets and platforms.

\subsection{Comprehensive Evaluation}

\subsubsection{GUI Understanding}
We evaluate models on MMBench-GUI L1 (GUI Content Understanding), which measures fine-grained perception and reasoning across six platforms following MMBench-GUI protocols. As shown in Table~\ref{tab:mmbengui-l1}, our \modelname consistently delivers competitive or superior results. Even the lightweight \modelname{3} attains 83.6\% in the easy level, surpassing Qwen2.5-VL-72B by +16.6 points. \modelname{7} further improves to 88.4\%, while \modelname{32} reaches 92.5\%, matching GUI-Owl-32B despite using a different training corpus. On medium-level tasks, \modelname{32} yields 92.5\%, and on hard-level tasks it achieves 94.4\%, outperforming all proprietary and open-sourced models. These results highlight the effectiveness of scaling with cross-platform GUI-specific data, confirming that diverse training corpora substantially enhance visual comprehension across heterogeneous environments.
\begin{table}[t]
    \centering
    \small
    \caption{Results on ScreenSpot-v2~\citep{wu2024osatlas}.}
    \resizebox{0.85\textwidth}{!}{
    \begin{tabular}{lccccccc}
        \toprule
        \multirow{2}{*}{\textbf{Method}} & \multicolumn{2}{c}{\textbf{Mobile}} & \multicolumn{2}{c}{\textbf{Desktop}} & \multicolumn{2}{c}{\textbf{Web}} & \multirow{2}{*}{\textbf{Avg}} \\
        \cmidrule(lr){2-3} \cmidrule(lr){4-5} \cmidrule(lr){6-7} % Partial horizontal line
        & \textbf{Text} & \textbf{Icon/Widget} & \textbf{Text} & \textbf{Icon/Widget} & \textbf{Text} & \textbf{Icon/Widget} & \\
        \midrule
        Operator~\citeyearpar{cua2025} & 47.3 & 41.5 & 90.2 & 80.3 & 92.8 & 84.3 & 70.5 \\
        Claude-3.7-Sonnet~\citeyearpar{anthropic2025claude37} & -- & -- & -- & -- & -- & -- & 87.6 \\
        UI-TARS-1.5~\citeyearpar{qin2025uitars} & -- & -- & -- & -- & -- & -- & 94.2 \\
        Seed-1.5-VL~\citeyearpar{guo2025seed1} & -- & -- & -- & -- & -- & -- & \textbf{95.2} \\
        Kimi-VL-A3B-Thinking-2506~\citeyearpar{kimiteam2025kimivltechnicalreport} & -- & -- & -- & -- & -- & -- & 91.4 \\
        MiMo-VL-7B-RL~\citeyearpar{coreteam2025mimovltechnicalreport} & -- & -- & -- & -- & -- & -- & 90.5 \\
        InternVL3.5-241B-A28B~\citeyearpar{wang2025internvl35} & 97.9 &	91.5 & 97.4 & 82.9 & 94.0 &	\underline{89.2} & 92.9  \\
        SeeClick~\citeyearpar{cheng2024seeclick} & 78.4 & 50.7 & 70.1 & 29.3 & 55.2 & 32.5 & 55.1 \\
        OmniParser-v2~\citeyearpar{lu2024omniparserpurevisionbased} & 95.5 & 74.6 & 92.3 & 60.9 & 88.0 & 59.6 & 80.7 \\
        OS-Atlas-Base-7B~\citeyearpar{wu2024osatlas} & 95.2 & 75.8 & 90.7 & 63.6 & 90.6 & 77.3 & 84.1 \\
        UI-TARS-2B~\citeyearpar{qin2025uitars} & 95.2 & 79.1 & 90.7 & 68.6 & 87.2 & 78.3 & 84.7 \\
        UI-TARS-7B~\citeyearpar{qin2025uitars} & 96.9 & 89.1 & 95.4 & 85.0 & 93.6 & 85.2 & 91.6 \\
        UI-TARS-72B~\citeyearpar{qin2025uitars} & 94.8 & 86.3 & 91.2 & 87.9 & 91.5 & 87.7 & 90.3 \\
        Qwen2.5-VL-3B~\citeyearpar{bai2025qwen25vl} & 93.4 & 73.5 & 88.1 & 58.6 & 88.0 & 71.4 & 80.9 \\
        Qwen2.5-VL-7B~\citeyearpar{bai2025qwen25vl} & 97.6 & 87.2 & 90.2 & 74.2 & 93.2 & 81.3 & 88.8 \\
        Qwen2.5-VL-32B~\citeyearpar{bai2025qwen25vl} & 97.9 & 88.2 & \textbf{98.5} & 79.3 & 91.2 & 86.2 & 91.3 \\
        JEDI-3B~\citeyearpar{xie2025scalingcomputerusegroundinguser} & 96.6 & 81.5 & 96.9 & 78.6 & 88.5 & 83.7 & 88.6 \\
        JEDI-7B~\citeyearpar{xie2025scalingcomputerusegroundinguser} & 96.9 & 87.2 & 95.9 & 87.9 & \underline{94.4} & 84.2 & 91.7 \\
        GUI-Actor-7B~\citeyearpar{wu2025gui} & 97.6 & 88.2 & 96.9 & 85.7 & 93.2 & 86.7 & 92.1 \\
        GTA1-7B~\citeyearpar{yang2025gta1} & \textbf{99.0} & 88.6 & 94.9 & \underline{89.3} & 92.3 & 86.7 & 92.4 \\
        GTA1-32B~\citeyearpar{yang2025gta1} & \underline{98.6} & 89.1 & 96.4 & 86.4 & \textbf{95.7} & 88.7 & 93.2 \\
        GUI-Owl-7B~\citeyearpar{ye2025mobile} & 99.0 & \textbf{92.4} & 96.9 & 85.0 & 93.6 & 85.2 & 92.8 \\
        GUI-Owl-32B~\citeyearpar{ye2025mobile} & 98.6 & 90.0 & 97.9 & 87.8 & \underline{94.4} & 86.7 & 93.2 \\
        \midrule
        \modelname{3} & 94.1 & 86.3 & 94.9 & 79.3 & 89.7 & 85.7 & 89.2\\
        \modelname{7} & 97.3 & 90.5 & 95.4 & 87.9 & 94.0 & 88.7 & 92.7 \\
        \modelname{32} & \underline{98.6} & \underline{91.9} & \textbf{99.0} & \textbf{90.0} & \underline{94.4} & \textbf{91.6} & \underline{94.7} \\
        \bottomrule
    \end{tabular}%
    }
    \label{tab:screenspot_v2_comparison}
\vspace{-1em}
\end{table}
\begin{table}[htbp] % Use [htbp] for better placement control
   \centering
   \small
   \caption{Results on ScreenSpot-Pro~\citep{li2025screenspotpro}.} % Add your desired caption
   \label{tab:screenspot_pro_comparison} % Add your desired label
   \resizebox{\textwidth}{!}{% Make the table fit the text width
   \begin{tabular}{@{}l rrr rrr rrr rrr rrr rrr r @{}}
       \toprule
       \multirow{2}{*}{\textbf{Agent Model}} & \multicolumn{3}{c}{\textbf{Development}} & \multicolumn{3}{c}{\textbf{Creative}} & \multicolumn{3}{c}{\textbf{CAD}} & \multicolumn{3}{c}{\textbf{Scientific}} & \multicolumn{3}{c}{\textbf{Office}} & \multicolumn{3}{c}{\textbf{OS}} & \multirow{2}{*}{\textbf{Avg}} \\
       \cmidrule(lr){2-4} \cmidrule(lr){5-7} \cmidrule(lr){8-10} \cmidrule(lr){11-13} \cmidrule(lr){14-16} \cmidrule(lr){17-19} 
       & Text & Icon & Avg & Text & Icon & Avg & Text & Icon & Avg & Text & Icon & Avg & Text & Icon & Avg & Text & Icon & Avg & \\
       \midrule
       SeeClick~\citeyearpar{cheng2024seeclick} & 0.6 & 0.0 & 0.3 & 1.0 & 0.0 & 0.6 & 2.5 & 0.0 & 1.9 & 3.5 & 0.0 & 2.0 & 1.1 & 0.0 & 0.9 & 2.8 & 0.0 & 1.5 & 1.1 \\
       Qwen2-VL-7B~\citeyearpar{wang2024qwen2vl} & 2.6 & 0.0 & 1.3 & 1.5 & 0.0 & 0.9 & 0.5 & 0.0 & 0.4 & 6.3 & 0.0 & 3.5 & 3.4 & 1.9 & 3.0 & 0.9 & 0.0 & 0.5 &  1.6 \\
       ShowUI-2B~\citeyearpar{lin2024showui} & 16.9 & 1.4 & 9.4 & 9.1 & 0.0 & 5.3 & 2.5 & 0.0 & 1.9 & 13.2 & 7.3 & 10.6 & 15.3 & 7.5 & 13.5 & 10.3 & 2.2 & 6.6 & 7.7 \\
       CogAgent-18B~\citeyearpar{hong2024cogagent} & 14.9 & 0.7 & 8.0 & 9.6 & 0.0 & 5.6 & 7.1 & 3.1 & 6.1 & 22.2 & 1.8 & 13.4 & 13.0 & 0.0 & 10.0 & 5.6 & 0.0 & 3.1 & 7.7 \\
       Aria-UI~\citeyearpar{yang2024aria} & 16.2 & 0.0 & 8.4 & 23.7 & 2.1 & 14.7 & 7.6 & 1.6 & 6.1 & 27.1 & 6.4 & 18.1 & 20.3 & 1.9 & 16.1 & 4.7 & 0.0 & 2.6 & 11.3 \\
       Claude~\citeyearpar{AnthropicModelCA} & 22.0 & 3.9 & 12.6 & 25.9 & 3.4 & 16.8 & 14.5 & 3.7 & 11.9 & 33.9 & 15.8 & 25.8 & 30.1 & 16.3 & 26.9 & 11.0 & 4.5 & 8.1 & 17.1 \\
       Operator~\citeyearpar{cua2025} & 50.0 & 19.3 & 35.1 & 51.5 & 23.1 & 39.6 & 16.8 & 14.1 & 16.1 & 58.3 & 24.5 & 43.7 & 60.5 & 28.3 & 53.0 & 34.6 & 30.3 & 32.7 & 36.6 \\
       OS-Atlas-7B~\citeyearpar{wu2024osatlas}  & 33.1 & 1.4 & 17.7 & 28.8 & 2.8 & 17.9 & 12.2 & 4.7 & 10.3 & 37.5 & 7.3 & 24.4 & 33.9 & 5.7 & 27.4 & 27.1 & 4.5 & 16.8 & 18.9 \\
       UGround-V1-7B~\citeyearpar{gou2024navigating} & -- & -- & 35.5 & -- & -- & 27.8 & -- & -- & 13.5 & -- & -- & 38.8 & -- & -- & 48.8 & -- & -- & 26.1 & 31.1 \\
       UGround-V1-72B~\citeyearpar{gou2024navigating} & -- & -- & 31.1 & -- & -- & 35.8 & -- & -- & 13.8 & -- & -- & 50.0 & -- & -- & 51.3 & -- & -- & 25.5 & 34.5 \\
       UI-TARS-2B~\citeyearpar{qin2025uitars} & 47.4 & 4.1 & 26.4 & 42.9 & 6.3 & 27.6 & 17.8 & 4.7 & 14.6 & 56.9 & 17.3 & 39.8 & 50.3 & 17.0 & 42.6 & 21.5 & 5.6 & 14.3 & 27.7 \\
       UI-TARS-7B~\citeyearpar{qin2025uitars} & 58.4 & 12.4& 36.1 & 50.0 & 9.1 & 32.8 & 20.8 & 9.4 & 18.0 & 63.9 & 31.8 & 50.0 & 63.3 & 20.8 & 53.5 & 30.8 & 16.9 & 24.5 & 35.7 \\
       UI-TARS-1.5-7B~\citeyearpar{qin2025uitars} & 58.4 & 12.4 & 31.8 & 58.1 & 15.4 & 40.2 & 38.6 & 11.0 & 31.8 & 66.7 & 21.9 & 47.2 & 74.6 & 35.9 & 65.6 & 49.5 & 13.5 & 33.2 & 42.0 \\
       UI-TARS-72B~\citeyearpar{qin2025uitars} & 63.0 & 17.3 & 40.8 & 57.1 & 15.4 & 39.6 & 18.8 & 12.5 & 17.2 & 64.6 & 20.9 & 45.7 & 63.3 & 26.4 & 54.8 & 42.1 & 15.7 & 30.1 & 38.1 \\
       Qwen2.5-VL-3B~\citeyearpar{bai2025qwen25vl} & 38.3 & 3.4 & 21.4 & 40.9 & 4.9 & 25.8 & 22.3 & 6.3 & 18.4 & 44.4 & 10.0 & 29.5 & 48.0 & 17.0 & 40.9 & 33.6 & 4.5 & 20.4 & 25.9 \\
       Qwen2.5-VL-7B~\citeyearpar{bai2025qwen25vl} & 51.9 & 4.8 & 29.1 & 36.9 & 8.4 & 24.9 & 17.8 & 1.6 & 13.8 & 48.6 & 8.2 & 31.1 & 53.7 & 18.9 & 45.7 & 34.6 & 7.9 & 22.4 & 27.6 \\
       Qwen2.5-VL-32B~\citeyearpar{bai2025qwen25vl} & 74.0 & 21.4 & 48.5 & 61.1 & 13.3 & 41.1 & 38.1 & 15.6 & 32.6 & 78.5 & 29.1 & 57.1 & 76.3 & 37.7 & 67.4 & 55.1 & 27.0 & 42.3 & 47.6 \\
       JEDI-3B~\citeyearpar{xie2025scalingcomputerusegroundinguser} & 61.0 & 13.8 & 38.1 & 53.5 & 8.4 & 34.6 & 27.4 & 9.4 & 23.0 & 54.2 & 18.2 & 38.6 & 64.4 & 32.1 & 57.0 & 38.3 & 9.0 & 25.0 & 36.1 \\
       JEDI-7B~\citeyearpar{xie2025scalingcomputerusegroundinguser} & 42.9 & 11.0 & 27.4 & 50.0 & 11.9 & 34.0 & 38.0 & 14.1 & 32.2 & 72.9 & 25.5 & 52.4 & 75.1 & 47.2 & 68.7 & 33.6 & 16.9 & 26.0 & 39.5 \\
       UI-R1-3B~\citeyearpar{lu2025ui} & 22.7 & 4.1 & -- & 27.3 & 3.5 & -- & 11.2 & 6.3 & -- & 42.4 & 11.8 & -- & 32.2 & 11.3 & -- & 13.1 & 4.5 & -- & 17.8 \\
       InfiGUI-R1-3B~\citeyearpar{liu2025infigui} & 51.3 & 12.4 & 32.4 & 44.9 & 7.0 & 29.0 & 33.0 & 14.1 & 28.4 & 58.3 & 20.0 & 41.7 & 65.5 & 28.3 & 57.0 & 43.9 & 12.4 & 29.6 & 35.7 \\
       GUI-G1-3B~\citeyearpar{zhou2025guig1} & 50.7 & 10.3 & 31.1 & 36.6 & 11.9 & 26.6 & 39.6 & 9.4 & 32.2 & 61.8 & 30.0 & 48.0 & 67.2 & 32.1 & 59.1 & 23.5 & 10.6 & 16.1 & 37.1 \\
       GUI-Owl-7B~\citeyearpar{yuan2025enhancing} & 76.6 & 31.0 & 54.5 & 59.6 & 27.3 & \underline{46.1} & 64.5 & 21.9 & \underline{54.1} & 79.1 & 37.3 & 61.0 & 77.4 & 39.6 & 68.7 & 59.8 & 33.7 & 47.9 & 54.9 \\
       GUI-Owl-32B~\citeyearpar{yuan2025enhancing} & 84.4 & 39.3 & \textbf{62.5} & 65.2 & 18.2 & 45.5 & 62.4 & 28.1 & 54.0 & 82.6 & 39.1 & \textbf{63.8} & 81.4& 39.6 & \underline{71.8} & 70.1 & 36.0 & \textbf{54.6}  & \underline{58.0} \\
       \midrule
       \modelname{3}  & 57.8  & 18.6 & 38.8 & 42.9 & 16.8 & 32.0 & 54.3 & 28.1 & 47.9 & 64.6 & 35.5 & 52.0 & 66.7 & 37.7 & 53.9 & 31.8 & 16.9 & 25.0 &42.3 \\
       \modelname{7}  & 66.2 & 20.7 & 44.1  & 56.6  &20.3 & 41.3  & 54.8 & 21.9 & 46.7 & 77.1 & 24.5 & 54.3 & 74.0 & 45.3 & 67.4 & 49.5 & 18.0 & 35.2 & 47.9 \\
       \modelname{32}  & 75.3 & 35.2  & \underline{55.8} & 73.2 & 30.8 & \textbf{55.4} & 60.4 & 39.1 & \textbf{55.2} & 76.4 & 46.4 & \underline{63.4} & 81.4 & 49.1 & \textbf{73.9} & 63.6 & 41.6 & \underline{53.6} & \textbf{59.2} \\
       \bottomrule
   \end{tabular}%
   } % End resizebox
   \vspace{-1em}
\end{table}

\subsubsection{GUI Grounding}
We then evaluate models on GUI grounding, which measures the ability to localize and associate visual elements with textual or functional references across desktop, mobile, and web environments. As shown in Tables~\ref{tab:screenspot_v2_comparison} and \ref{tab:screenspot_pro_comparison}, our \modelname consistently achieves state-of-the-art performance across different benchmarks. On ScreenSpot-v2, \modelname{32} attains an average score of $94.7$\%, surpassing all open VLMs. Even the lightweight \modelname{3} delivers $89.2$\%, outperforming Qwen2.5-VL-7B by $+0.4$\% despite using far fewer parameters. On the more challenging ScreenSpot-Pro, \modelname{32} again dominates, achieving $59.2$\% overall and delivering strong accuracy across diverse domains such as Creative software, CAD, and office applications.

More results on OSWorld-G and MMBench-GUI L2 can refer to~\ref{sec:more_results}. Overall, these results demonstrate that scaling with GUI-specific data yields substantial benefits for grounding. The consistent improvements across GUI grounding benchmarks confirm the effectiveness of our dual-loop data pipeline in learning robust UI element localization.

\subsubsection{Task Completion}
We evaluate end-to-end task completion on Mobile (AndroidWorld~\citep{rawles2024androidworld}), Ubuntu (OSWorld~\citep{xie2024osworld}), Windows (WindowsAgentArena~\citep{bonatti2024windows}), macOS (MacOSArena~\citep{wang2025mmbenchgui}), and Web (WebArena-Lite-v2), considering both native agents and planner–grounder workflows. The results in Table~\ref{tab:agent_comparison} show three outcomes. First, our native \modelname{32} achieves the strongest Web performance: 44.2\% (15 steps budget) and 47.4\% (50 steps), outperforming the best native baseline (UI-TARS-72B-DPO) by +20.8 and +26.0 points, respectively, and substantially surpassing Qwen2.5-VL-72B. 
Second, on WindowsAgentArena (15 steps), \modelname{32} reaches 24.2\%, exceeding UI-TARS-72B-DPO (17.9\%) by +6.3 points, though the 50 steps regime remains challenging for all native models. 
Third, the workflow setting with GPT-4o as planner and \modelname{7} as the grounder yields 48.3\% on AndroidWorld and 35.1\% on WebArena-Lite-v2 (50 steps), outperforming other strong grounders such as UI-TARS-1.5-7B (28.6\%) and UGround-V1-7B (26.5\%). 

Beyond these highlights, several trends emerge. (i) Scaling from 3B→7B→32B produces monotonic gains on Windows and Web, indicating that our cross-platform data and unified action space translate into stronger computer use agents as capacity grows. (ii) The effect of the step budget is consistent: a majority of the agents, including \modelname, achieve substantial performance improvements under a 50-step limit. (iii) Cross-platform variability persists: AndroidWorld favors planner-grounder workflows, narrowing but not closing the gap with mobile-focused native models; The results on macOS benchmark remain uniformly low, pointing to OS-specific affordance gaps rather than pure grounding limitations. Collectively, these results show that \modelname offers robust task completion as a foundation model for computer use agents and serves as a competitive, transferable grounder within planner–grounder pipelines.
\begin{table}[t]

\newcommand{\origin}{*}

\centering
\setlength{\tabcolsep}{4pt}
\caption{Online evaluation across different platforms and benchmarks for agents. In AndroidWorld, each task has its own predefined maximum number of steps. $^\clubsuit$ denotes the steps is unkown for the model and $^\bigstar$ indicates the steps are more than 50 steps.}
\label{tab:agent_comparison}
\resizebox{0.85\textwidth}{!}{% Adjusted to full text width to better accommodate new columns
\begin{tabular}{@{}c l c cc cc cc cc @{}} % Defines 2 columns for Method, then 5 pairs of columns for scores
\toprule
% --- Main Header Row (Spans 3 rows for the Method column) ---
\multicolumn{2}{c}{\multirow{3}{*}{\textbf{Method}}} & \multicolumn{1}{c}{\textbf{Mobile}} & \multicolumn{2}{c}{\textbf{Ubuntu}} & \multicolumn{2}{c}{\textbf{Windows}} & \multicolumn{2}{c}{\textbf{MacOS}} & \multicolumn{2}{c}{\textbf{Web}} \\

% --- Benchmark Name Row ---
\multicolumn{2}{c}{} & \multicolumn{1}{c}{\footnotesize (AndroidWorld)} & \multicolumn{2}{c}{\footnotesize (OSWorld)} & \multicolumn{2}{c}{\footnotesize (WindowsAgentArena)} & \multicolumn{2}{c}{\footnotesize (MacOSArena)} & \multicolumn{2}{c}{\footnotesize (WebArena-Lite-v2)} \\

% --- Horizontal Rules to Group Settings ---
\cmidrule(lr){3-3} \cmidrule(lr){4-5} \cmidrule(lr){6-7} \cmidrule(lr){8-9} \cmidrule(lr){10-11}

\multicolumn{2}{c}{} & \multicolumn{1}{c}{ {Predefined Steps}} &  15 Steps & 50 Steps &  15 Steps &  50 Steps &  15 Steps &  50 Steps &  15 Steps & 50 Steps \\
\midrule
\multicolumn{11}{l}{\textit{Native Agent}} \\ 
\midrule
% Kimi-VL-A3B: Unknow steps
\multicolumn{2}{l}{Kimi-VL-A3B~\citeyearpar{kimiteam2025kimivltechnicalreport}}  & -- &  \multicolumn{2}{c}{8.2$^\clubsuit$} & \multicolumn{2}{c}{10.4$^\clubsuit$} & -- & -- & -- & -- \\
% Seed1.5-VL: 100steps
\multicolumn{2}{l}{Seed1.5-VL~\citeyearpar{guo2025seed1}} & \textbf{62.1} & \multicolumn{2}{c}{36.7$^\bigstar$} & \multicolumn{2}{c}{39.6$^\bigstar$} & -- & -- & -- & -- \\
% GLM-4.1V-Thinking: 100steps
\multicolumn{2}{l}{GLM-4.1V-Thinking~\citeyearpar{hong2025glm}}  & 41.7 & \multicolumn{2}{c}{14.9$^\bigstar$} & -- & -- & -- & -- & -- & -- \\
\multicolumn{2}{l}{GLM-4.5V-Thinking~\citeyearpar{hong2025glm}}  & 57.0 & \multicolumn{2}{c}{35.8$^\bigstar$} & -- & -- & -- & -- & -- & -- \\
\multicolumn{2}{l}{PC Agent-E~\citeyearpar{he2025efficientagenttrainingcomputer}}   & -- & \multicolumn{2}{c}{14.9$^\clubsuit$} & -- & -- & -- & -- & -- & -- \\
\multicolumn{2}{l}{COMPUTERRL~\citeyearpar{lai2025computerrl}}   & -- & \multicolumn{2}{c}{47.3$^\clubsuit$} & -- & -- & -- & -- & -- & -- \\
\multicolumn{2}{l}{GPT-4o~\citeyearpar{hurst2024gpt}} & 21.6  & 6.8 & 10.1 & 5.6 & 3.5 & 0.0 & 1.4 & 2.0 & 3.3 \\
\multicolumn{2}{l}{Claude-3.7~\citeyearpar{anthropic2025claude37}}  & 11.2 & 7.4 & 10.3 & 7.1 & 6.4 & 5.7 & \underline{7.1} & 2.0 & 2.6 \\
\multicolumn{2}{l}{Qwen2.5-VL-72B~\citeyearpar{bai2025qwen25vl}}  & 27.6 & 9.8 & 10.6 & 11.8 & 9.7 & 1.4 & 5.7 & 15.6 & 14.4 \\
\multicolumn{2}{l}{InternVL3.5-241B-A28B~\citeyearpar{wang2025internvl35}}  & 29.7 & 11.1 & 11.6 & 15.2 & 18.0 & 2.9 & 5.7 & 11.7 & 11.7 \\
\multicolumn{2}{l}{Aguvis-72B~\citeyearpar{xu2024aguvis} }  & 26.1 & 3.8 & 4.2 & 4.1 & 3.5 & 0.0 & 0.0 & 5.8 & 9.0 \\
\multicolumn{2}{l}{UI-TARS-7B-SFT~\citeyearpar{qin2025uitars}}  & 33.0 & 17.7 & -- & -- & -- & -- & -- & 11.0 & 13.6 \\
\multicolumn{2}{l}{UI-TARS-1.5-7B~\citeyearpar{qin2025uitars} } & 31.6 & 22.1 & 23.9 & 11.1 & 15.9 & \underline{7.1} & \underline{7.1} & 20.8 & 26.0 \\
\multicolumn{2}{l}{UI-TARS-72B-DPO~\citeyearpar{qin2025uitars}}   &\underline{46.6} & 24.2  & 25.2 & 11.1 & 17.9 & \textbf{8.6} & \textbf{8.6} & 23.4 & 21.4 \\
\multicolumn{2}{l}{OpenCUA-7B~\citeyearpar{wang2025opencuaopenfoundationscomputeruse}}   & -- & \underline{24.3}  & \underline{28.1} & -- & -- & -- & -- & -- & -- \\
\multicolumn{2}{l}{OpenCUA-32B~\citeyearpar{wang2025opencuaopenfoundationscomputeruse}}   & -- & \textbf{29.7}  & \textbf{34.1} & -- & -- & -- & -- & -- & -- \\
\midrule
\multicolumn{2}{l}{\modelname{3}} & 23.7 & 9.6 & 12.4 & 13.1 & 15.2 & 0.0 & 1.4 & 31.8 & 33.1 \\ 
\multicolumn{2}{l}{\modelname{7}} & 27.2 & 14.3 & 15.0 & \underline{18.0} & \underline{20.7} & 4.3 & 4.3 & \underline{37.7} & \underline{37.7} \\ 
\multicolumn{2}{l}{\modelname{32}} & 30.6 & 16.5 & 17.7 & \textbf{21.4} & \textbf{24.2} & \underline{7.1} & \underline{7.1} & \textbf{44.2} & \textbf{47.4} \\ 
\midrule
\midrule
\multicolumn{11}{l}{\quad\quad\textit{Agentic Workflow}} \\
\cmidrule(l){1-2}
\textit{Planner} &\quad\quad\quad\textit{Grounder} & \multicolumn{9}{c}{} \\
\midrule
\multirow{7}{*}{GPT-4o} 
% Aria-UI: Unknow steps
 &  Aria-UI~\citeyearpar{yang2024aria}   & \underline{44.8} & \multicolumn{2}{c}{15.2$^\clubsuit$}& -- & -- & -- & -- & -- & -- \\ 
% OS-Atlas-7B: Unknow steps
&  OS-Atlas-7B~\citeyearpar{wu2024osatlas}   & -- & \multicolumn{2}{c}{14.6$^\clubsuit$} & -- & -- & -- & -- & -- & -- \\ 
& UGround-V1-7B~\citeyearpar{gou2024navigating} & 32.8 & 13.1 & 16.1 & 13.1 & 20.7 & \underline{1.4} & 0.0 & \underline{23.2} & 26.5 \\
 &  UI-TARS-1.5-7B~\citeyearpar{qin2025uitars}   & 37.9 & 16.5 & 19.1 & 14.5 & 26.2 & \underline{1.4} & 0.0 & 22.6 & \underline{28.6} \\
 &  JEDI-3B~\citeyearpar{xie2025scalingcomputerusegroundinguser}   & -- & 22.4 & -- & 29.1 & -- & -- & -- & -- & -- \\ 
 &  JEDI-7B~\citeyearpar{xie2025scalingcomputerusegroundinguser}   & -- & \underline{22.7} & \underline{25.0} & \underline{30.2} & \underline{32.9} & -- & -- & -- & -- \\ 
 \cmidrule(lr){2-11}
 &  \modelname{7}   & \textbf{48.3} & \textbf{22.9} & \textbf{28.1} & \textbf{31.7} & \textbf{36.6} & \textbf{5.7} & \textbf{8.6} & \textbf{28.6} & \textbf{35.1} \\
\bottomrule
\end{tabular}
}
\vspace{-1em}
\end{table}

\subsection{Ablation on Our Dataset}
In this section, we aim to ablate our dataset. As shown in Fig.~\ref{fig:teaser}, training with our curated training corpus yields consistent improvements over the baseline trained on public data. In Table~\ref{tab:ablation_data}, we further highlight the effects of augmentation, weak semantic trajectories, coordinate formats, and resolution.

First, we evaluate the effect of data augmentation. Without augmentation, the model reaches $37.8$\% on ScreenSpot-Pro, whereas simple augmentations such as element cropping and screenshot concatenation increase performance to $41.3$\%. This confirms that augmentation enhances generalization and robustness by exposing the model to a wider range of visual conditions.
Second, we investigate weak semantic trajectories derived from rule-based random exploration. Despite lacking explicit high-level goals, these trajectories provide low-cost supervision of interface navigation. 
Third, we study the impact of coordinate representations in grounding. Models trained with raw coordinates outperform those with normalized coordinates ($42.3$\% \vs $37.9$\% on ScreenSpot-Pro). This indicates that precise absolute positions better capture cross-platform layout regularities, reducing the ambiguity introduced by normalization.
Finally, we ablate the training resolution. Higher resolutions yield trade-offs across benchmarks: while 2K improves grounding on ScreenSpot-Pro ($45.5$\% \vs $42.3$\%) and preserves OSWorld-G accuracy ($52.5$\% \vs $54.3$\%), it slightly reduces agent success rates on OSWorld and AndroidWorld. This suggests that fine-grained grounding benefits from high-resolution supervision, whereas agentic benchmarks may suffer from overfitting to pixel-level details.
The improvements across UI element grounding and online agent benchmarks demonstrate that the design of training data is central to building scalable and generalizable computer use agents.
\begin{table}[tbhp]
\centering
\small
\vspace{-1em}
\caption{Ablation studies on data. The maximum steps used in online benchmarks are set to 50.} 
\label{tab:ablation_data}
\begin{minipage}{0.42\linewidth}
\centering
\subcaption{The ablation on data augmentation. We only use GUI-related data in training.}
\label{tab:ablation_data_aug}
\resizebox{0.9\textwidth}{!}{% Adjusted to full text width to better 
\begin{tabular}{lccc}
    \toprule
    \textbf{Model} & \textbf{Training Data} & \textbf{Aug.} & \textbf{SS-Pro} \\
    \midrule
    \multirow{2}{*}{Qwen2.5VL-3B} & \multirow{2}{*}{ours-only} & \xmark & 37.8 \\
    &  & \cmark & \textbf{41.3}  \\
    \bottomrule
\end{tabular}}
\end{minipage}
% \hfill
\quad
\begin{minipage}{0.54\linewidth}
\centering
\subcaption{The ablation on weak semantic trajectories. The public datasets used are shown in Table~\ref{tab:public_data}.}
\label{tab:ablation_weak_semantic}
\resizebox{0.92\textwidth}{!}{% Adjusted to full text width to better 
\begin{tabular}{lcccc}
    \toprule
    \textbf{Model} & \textbf{Training Data} & \textbf{+ WS} & \textbf{OSWorld} & \textbf{WAL-v2} \\
    \midrule
    \multirow{2}{*}{Qwen2.5VL-3B} & \multirow{2}{*}{public-only} & \xmark & 7.9 & 8.4  \\
    & & \cmark & \textbf{8.5} & \textbf{14.3} \\
    \bottomrule
\end{tabular}}
\end{minipage}

\vspace{0.8em}

\begin{minipage}{0.4\linewidth}
\centering
\subcaption{The ablation on coordinate types.}
\label{tab:ablation_coord_types}
\resizebox{0.9\textwidth}{!}{% Adjusted to full text width to better 
\begin{tabular}{lcc}
    \toprule
    \textbf{Model} & \textbf{Type} & \textbf{ScreenSpot-Pro} \\
    \midrule
    \multirow{2}{*}{Qwen2.5VL-3B} & Norm. & 37.9 \\
    & Raw & \textbf{42.3}\\
    \bottomrule
\end{tabular}}
\end{minipage}
\quad
\begin{minipage}{0.55\linewidth}
\centering
\subcaption{The ablation on the maximum resolution during training.}
\label{tab:ablation_res}
\resizebox{0.92\textwidth}{!}{% Adjusted to full text width to better 
\begin{tabular}{lccccc}
    \toprule
    \textbf{Model} & \textbf{Res.} & \textbf{SS-Pro} & \textbf{OSWOrld-G} &\textbf{OSWorld} & \textbf{AW} \\
    \midrule
    \multirow{2}{*}{Qwen2.5VL-3B} & 1080P & 42.3 & \textbf{54.3} & \textbf{12.4} & \textbf{23.3} \\
    & 2K & \textbf{45.5} & 52.5 & 11.2 & 13.4 \\
    \bottomrule
\end{tabular}}
\end{minipage}
\end{table}

\subsection{Diagnostic Analysis on Agents}
\begin{figure}[t]
    \centering
    \includegraphics[width=1\linewidth]{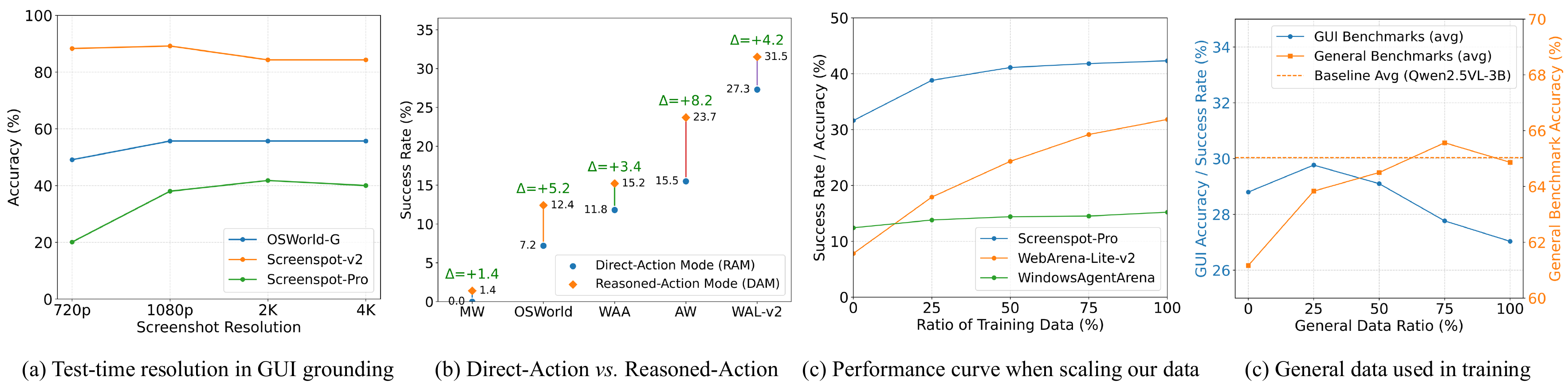}
    \small
    \caption{Evaluations across diverse conditions. (a) Accuracy of GUI grounding under different screenshot resolutions. (b) Success rates of Direct Action vs. Reasoned Action Modes, where reasoning consistently improves performance. (c) Training data scaling. (d) Effect of using general data, showing distinct trends between GUI and multimodal benchmarks.}
    \label{fig:empirical_study}
\end{figure}
To better understand the factors that influence the performance of computer use agents, we conduct additional analyses on resolution, inference mode, data scaling, and the inclusion of general multimodal data, as illustrated in Fig.~\ref{fig:empirical_study}. These studies provide deeper insights into the trade-offs between efficiency, accuracy, and generalization.

\textbf{Impact of Resolution.}
Fig.~\ref{fig:empirical_study} (a) evaluates grounding accuracy across different screenshot resolutions. We observe that accuracy consistently improves when moving from 720p to 2K, with diminishing returns at 4K. ScreenSpot-Pro benefits most significantly, demonstrating that fine-grained element localization requires high-resolution details. 
Images with resolutions less than or equal to 1080p play a dominant role in both the ScreenSpot-v2 and OSWorld-G. As a result, increasing the input resolution beyond 1080p during evaluation does not yield performance gains, and in some cases, may even lead to slight degradation.

\textbf{Direct Action Mode vs. Reasoned Action Mode.}
Fig.~\ref{fig:empirical_study} (b) compares the two inference modes for computer use agents. Across all benchmarks, Reasoned Action Mode (RAM) yields higher success rates than Direct Action Mode (DAM), with absolute gains ranging from +$1.4$\% to + $8.2$\%. This advantage is particularly pronounced in complex, multi-step environments such as WebArena-Lite-v2. The results indicate that explicit reasoning helps the model to maintain task coherence and reduce error propagation in long-horizon interactions. Nonetheless, the improvements come at the cost of higher inference latency, suggesting a trade-off between efficiency and robustness depending on deployment scenarios.

\textbf{Scaling Behavior with Training Data.}
Fig.~\ref{fig:empirical_study} (c) illustrates the success rate accuracy as a function of training data ratio. All benchmarks benefit from increased data, but the gains exhibit a logarithmic trend, with most improvements saturating beyond $75$\% of the data. Notably, WebArena-Lite-v2 shows a nearly linear improvement across the entire range, underscoring its higher data hunger and greater sensitivity to scale. By contrast, ScreenSpot-Pro reaches competitive accuracy with only half of the available data, highlighting that certain grounding benchmarks are less data-intensive. This observation suggests that data allocation should be tailored to the intrinsic complexity of benchmarks.

\textbf{Effect of General Multimodal Data.}
Fig.~\ref{fig:empirical_study} (d) analyzes the effect when employing general-purpose multimodal data in training. We find a clear divergence: GUI benchmarks suffer a gradual decline in performance as the ratio of general data increases, while general benchmarks improve steadily, peaking around 75\%. This trade-off indicates that mixing heterogeneous data sources introduces conflicting optimization signals. 
As the multimodal corpus expands, the model’s general capabilities improve, but GUI-specific knowledge is diluted.
The results suggest that data-balanced training strategies may be necessary to preserve GUI specialization without sacrificing general reasoning ability. Since the larger models are able to memorize more knowledge, it is reasonable to increase the ratio to 50\% for the 7B model and further to 75\% for the 32B model.

\textbf{Discussion.}
These figures highlight several key observations. High-resolution inputs and reasoning-based inference enhance grounding and task completion, but incur computational costs. Data scaling remains essential, though different benchmarks vary in sensitivity to scale. Finally, while incorporating general multimodal data improves reasoning ability, it can compromise GUI grounding accuracy, pointing to the need for careful data mixing strategies. These findings underscore the importance of scalable cross-platform training data to build robust general-purpose agents.

\section{Conclusion}
In this work, we propose \textbf{\modelname}, a new family of models and datasets tailored for CUAs. By curating large-scale cross-platform data and standardizing the action space across desktop, mobile, and web environments, we established a unified foundation for evaluating and training agents. Our dual-loop data pipeline—combining automated agent-environment interaction with human-in-the-loop annotation—enabled the construction of diverse benchmarks spanning GUI understanding, element grounding, and task completion.  
Extensive experiments demonstrated that our proposed \modelname significantly outperforms both open-source and proprietary baselines. On GUI understanding tasks, \modelname{32} achieves state-of-the-art accuracy across all platforms and difficulty levels, underscoring the benefits of GUI-specific supervision. In grounding benchmarks such as ScreenSpot-v2, ScreenSpot-Pro, and OSWorld-G, our models consistently establish new performance records, validating the effectiveness of fine-grained element-level alignment. Moreover, the strong results on L2-GUI grounding and online benchmarks further highlight the models’ robustness in realistic, interactive environments. 
Together, these contributions advance the frontier of computer use agents by bridging vision-language modeling with practical GUI interaction. While limitations remain in long-horizon reasoning and cross-application generalization, our results point toward the promise of scaling specialized training pipelines for robust, general-purpose computer use systems. We hope that \modelname and its released resources will serve as a solid foundation for future research in building capable, trustworthy, and deployable CUAs.

\section*{Limitations}
Our work develops a family of base agent models and establishes a comprehensive data pipeline to collect cross-platform GUI interaction trajectories. Despite that, several limitations remain. 

First, the quality of agent-collected data still lags behind expert crowdsourcing. Rule-based exploration often produces semantically weak trajectories, whereas expert-curated data is costly to scale. How to effectively couple automated collection with iterative model refinement to form a self-improving loop remains an open challenge.  

Second, while our study focuses on constructing robust base models for computer use agents across platforms, we have not incorporated advanced agentic mechanisms such as reflection, memory-based decision-making, or hierarchical planning. Furthermore, optimization strategies including Policy Reward Models (PRMs) or Reinforcement Learning are left unexplored. These techniques hold promise for enhancing reasoning, adaptability, and long-horizon decision making, but require strong base models as a prerequisite.  

Third, the current memory mechanism is rudimentary, treating previous operations as flat history. This shallow design limits long-horizon reasoning and may cause suboptimal executions. In this work, our primary goal is to develop base agent models with strong generalization and instruction-following abilities trained from large-scale GUI corpora. Developing lightweight yet effective memory systems that capture temporal dependencies is an important direction for future work.  

In summary, while our framework provides a unified foundation for training GUI agents, further work is needed to enable closed-loop learning and high-level reasoning capabilities. Addressing these limitations will be critical to realize fully autonomous and intelligent computer use agents.

% We would like to thank Xingpu Li, Wenjie Tan, and Zhenshi Xie for their valuable support on the training infrastructure, as well as for their assistance in diagnosing and resolving issues encountered during large-scale model training.  

\bibliography{iclr2025_conference}
\bibliographystyle{iclr2025_conference}

\appendix
\tcbset{colframe = cyan!40!black, 
% colback = white,
colback = cyan!1!white, colbacktitle =  cyan!40!black,
toptitle=1mm, 
bottomtitle=1mm, 
fonttitle=\fontsize{10pt}{12pt}\selectfont,  breakable, enhanced jigsaw, finish broken={\tcbset{colframe=white, bottom=0mm, bottomrule=0mm}},before upper={\tcbset{colframe=white, toprule=0mm}},after={\tcbset{colframe=blue!40!black, toprule=1mm, bottomrule=1mm}},}

\section{Appendix}
\label{sec:appendix}

\subsection{More Results}
\label{sec:more_results}
\begin{table*}[!b]
  \centering
  \caption{Performance on the MMBench-GUI L2 (GUI Element Grounding)~\citep{wang2025mmbenchgui}.}
  \small
  \resizebox{\textwidth}{!}{% Make the table fit the text width
  \begin{tabular}{llllllllllllll}
    \toprule
    \multirow{2}{*}{\textbf{Model}} 
      & \multicolumn{2}{c}{\textbf{Windows}} 
      & \multicolumn{2}{c}{\textbf{MacOS}} 
      & \multicolumn{2}{c}{\textbf{Linux}} 
      & \multicolumn{2}{c}{\textbf{iOS}} 
      & \multicolumn{2}{c}{\textbf{Android}} 
      & \multicolumn{2}{c}{\textbf{Web}} 
      & \multirow{2}{*}{\textbf{Avg}}\\
    \cmidrule(r){2-3}
    \cmidrule(r){4-5}
    \cmidrule(r){6-7}
    \cmidrule(r){8-9}
    \cmidrule(r){10-11}
    \cmidrule(r){12-13}
      & Basic & Adv. & Basic & Adv. & Basic & Adv. & Basic & Adv. & Basic & Adv. & Basic & Adv. & \\
    \midrule
    GPT-4o~\citeyearpar{hurst2024gpt} & 1.5 &  1.1  & 8.7 & 4.3 & 1.1 & 1.0 & 5.1 & 3.3 & 2.5 & 1.4 & 3.2 & 2.9 & 2.9 \\
    Claude-3.7~\citeyearpar{anthropic2025claude37} & 1.5   & 0.7   & 12.5 & 7.5 & 1.1 & 0.0 & 13.7 & 10.6 & 1.4 & 1.4 & 3.2 &2.3 & 4.7 \\
    Qwen-Max-VL~\citeyearpar{Qwen-VL} &  43.9  &  36.8  & 58.8 & {56.1} & 53.9 & 30.1 & 77.4 & 59.1 & 79.5 & 70.1 & 74.8 & 58.8 & 58.0 \\
    Aguvis-7B-720P~\citeyearpar{xu2024aguvis} &  37.3  &  21.7  & 48.1 & 33.3 & 33.5 & 25.0& 67.5 & 65.2 & 61.0 & 51.0 & 61.6 & 45.5 & 45.7\\
    ShowUI-2B~\citeyearpar{xu2024aguvis} &  9.2  &  4.4  & 24.1 & 10.4 & 25.1 & 11.7 &29.0  & 19.7& 17.4 & 8.7 & 22.9 & 12.7 & 16.0 \\
    OS-Atlas-Base-7B~\citeyearpar{wu2024osatlas}  &  36.9  &  18.8  & 44.4 & 21.7 & 31.4 & 13.3  & 74.8 & 48.8  & 69.6 & 46.8  & 61.3 & 35.4 & 41.4\\
    UGround-V1-7B~\citeyearpar{gou2024navigating} &  66.8  &  39.0  & 71.3 & 48.6 & 56.5 & 31.1 & {92.7} & 70.9 & {93.5} & 71.0 & {88.7} & 64.6 & 65.7 \\
    InternVL3-72B~\citeyearpar{zhu2025internvl3} &  {70.1}  & {42.6}   & {75.7}& 52.3 & 59.2 & {41.3} & {93.6} & {80.6} & 92.7 & {78.6} & {90.7} & {65.9} & {72.2}\\
    Qwen2.5-VL-72B~\citeyearpar{bai2025qwen25vl} &  55.7  &  33.8  & 49.9 & 30.1 & 40.3 & 20.9 & 56.1 & 28.2 & 55.6 & 25.4 & 68.4 & 45.8 & 41.8 \\
    Qwen2.5-VL-7B~\citeyearpar{bai2025qwen25vl} &  31.4  &  16.5  & 31.3 & 22.0 & 21.5 & 12.2 & 66.6 & 55.2 & 35.1 & 35.2 & 40.3 & 32.5 & 33.9 \\
    UI-TARS-1.5-7B~\citeyearpar{qin2025uitars} &  68.3 &  39.0  & 69.0 & 44.5 & {64.4} & 37.8 & 88.5 & 69.4 & 90.5 & 69.3 & 81.0 & 56.5  & 64.3 \\
    UI-TARS-72B-DPO~\citeyearpar{qin2025uitars} &  {78.6}  & {51.8}  & {80.3} & {62.7} & {68.6} &{51.5} & 90.8 &{81.2}& {93.0} & {80.0} & 88.1& {68.5} & {74.3} \\
    GUI-Owl-7B~\citeyearpar{ye2025mobile}  & \textbf{86.3} & 61.8 & 81.7 & \underline{64.5} & 74.4 & {61.7} & 94.9 & 83.0 & 95.8  &  \underline{83.7} & 93.2  & 72.7  &  80.5 \\
    GUI-Owl-32B~\citeyearpar{ye2025mobile}  & \underline{85.6} & \textbf{65.1} & \underline{84.9} & \textbf{67.1} & \underline{77.0} & \underline{63.3} & \underline{95.2} & \textbf{85.5} & \underline{96.1}  & \textbf{87.0}  & \textbf{95.5}  & \textbf{80.8}  & \textbf{83.0}  \\
    \hline
    ScaleCUA-3B &  78.6  & 46.0 & 79.4 & 52.9 & 73.3 & 49.0 & 93.0 & 73.3 &94.1  & 74.4 & 92.6 & 63.6 & 73.7 \\
    ScaleCUA-7B &  78.6  & 54.0 & 82.3 & 58.7 & 74.4 & 56.6 & 94.3 & 81.8 &\underline{96.1}  & 81.1 & 92.6 & 73.1 & 78.2 \\
    ScaleCUA-32B &  83.0  & \underline{62.9} & \textbf{88.1} & 64.2 & \textbf{81.2} & \textbf{65.8} & \textbf{95.9} & \underline{84.9} & \textbf{96.4}  & {81.7} & \underline{93.9} & \underline{76.3} & \underline{82.0} \\
    \bottomrule
  \end{tabular}
  }
  \vspace{-3mm}
  \label{tab:mmbenchgui-l2}
\end{table*}
\begin{table}[tbh]
  \centering
  \caption{Performance comparison on OSWorld-G~\citep{xie2025scalingcomputerusegroundinguser}.}
  \label{tab:osworld-g}
  \resizebox{\textwidth}{!}{% Make the table fit the text width
  \begin{tabular}{lcccccc}
    \toprule
    \multirow{2}{*}{Agent Model}   & \multirow{2}{*}{\makecell{\textbf{Text} \\ \textbf{Matching}}} & \multirow{2}{*}{\makecell{\textbf{Element} \\ \textbf{Recognition}}} & \multirow{2}{*}{\makecell{\textbf{Layout} \\ \textbf{Understanding}}} & \multirow{2}{*}{\makecell{\textbf{Fine-grained} \\ \textbf{Manipulation}}} & \multirow{2}{*}{\textbf{Refusal}} & \multirow{2}{*}{\textbf{Overall}} \\
    & & & & & & \\
    \midrule
    Gemini-2.5-Pro~\citeyearpar{comanici2025gemini}     & 59.8 & 45.5 & 49.0 & 33.6 & \textbf{38.9} & 45.2 \\
    Operator~\citeyearpar{cua2025}           & 51.3 & 42.4 & 46.6 & 31.5 & 0.0  & 40.6 \\
    Seed1.5-VL~\citeyearpar{guo2025seed1}         & \textbf{73.9} & \textbf{66.7} & \underline{69.6} & 47.0 & \underline{18.5} & \textbf{62.9} \\    
    OS-Atlas-7B~\citeyearpar{wu2024osatlas}        & 44.1 & 29.4 & 35.2 & 16.8 & 7.4  & 27.7 \\
    UGround-V1-7B~\citeyearpar{gou2024navigating}      & 51.3 & 40.3 & 43.5 & 24.8 & 0.0  & 36.4 \\
    Aguvis-7B~\citeyearpar{xu2024aguvis}          & 55.9 & 41.2 & 43.9 & 28.2 & 0.0  & 38.7 \\
    UI-TARS-7B~\citeyearpar{qin2025uitars}         & 60.2 & 51.8 & 54.9 & 35.6 & 0.0  & 47.5 \\
    UI-TARS-1.5-7B~\citeyearpar{qin2025uitars}     & \underline{70.1} & 57.9 & 59.7 & 51.7 & 0.0 & 56.0 \\
    UI-TARS-72B~\citeyearpar{qin2025uitars}        & 69.4 & 60.6 & 62.9 & 45.6 & 0.0  & 57.1 \\
    Qwen2.5-VL-3B~\citeyearpar{bai2025qwen25vl}      & 41.4 & 28.8 & 34.8 & 13.4 & 0.0  & 27.3 \\
    Qwen2.5-VL-7B~\citeyearpar{bai2025qwen25vl}      & 45.6 & 32.7 & 41.9 & 18.1 & 0.0  & 31.4 \\
    Qwen2.5-VL-32B~\citeyearpar{bai2025qwen25vl}     & 63.2 & 47.3 & 49.0 & 36.9 & 0.0  & 46.5 \\
    InternVL3.5-241B-A28B~\citeyearpar{wang2025internvl35} & 64.4 & 58.8 & 55.3 & 43.0 & 0.0  & 53.2 \\ 
    JEDI-3B~\citeyearpar{xie2025scalingcomputerusegroundinguser}            & 67.4 & 53.0 & 53.8 & 44.3 & 7.4  & 50.9 \\
    JEDI-7B~\citeyearpar{xie2025scalingcomputerusegroundinguser}            & 65.9 & 55.5 & 57.7 & 46.9 & 7.4  & 54.1 \\
    \midrule
    \modelname{3}         & 64.8 & \underline{61.8} & 64.0 & 43.6 & 0.0  & 55.7 \\
    \modelname{7}   &   67.8   & \underline{61.8}   &  64.8 &  \underline{49.7}   & 0.0  &  57.8  \\
    \modelname{32}   &  69.0   &   \textbf{66.7}  &  \textbf{70.0}    &  \textbf{51.0}  &   0.0   &   \underline{60.6}   \\
    \bottomrule
  \end{tabular}
  }
\end{table}
\begin{table}[t]
  \centering
  \caption{Performance comparison on AndroidControl~\citep{li2024effects}.}
  \label{tab:androidcontrol_comparison}
  \resizebox{0.85\textwidth}{!}{
  \begin{tabular}{lccc ccc}
    \toprule
    \multirow{2}{*}{\textbf{Agent Model}} &
    \multicolumn{3}{c}{\textbf{AndroidControl-Low}} &
    \multicolumn{3}{c}{\textbf{AndroidControl-High}} \\
    \cmidrule(lr){2-4} \cmidrule(lr){5-7}
    & \textbf{Type} & \textbf{Grounding} & \textbf{SR} 
    & \textbf{Type} & \textbf{Grounding} & \textbf{SR} \\
    \midrule
    Claude~\citeyearpar{AnthropicModelCA}       & 74.3 & 0.0 & 19.4 & 63.7 & 0.0 & 12.5 \\
    GPT-4o~\citeyearpar{hurst2024gpt}       & 74.3 & 0.0 & 19.4 & 66.3 & 0.0 & 20.8 \\
    SeeClick~\citeyearpar{cheng2024seeclick}     & 93.0 & 73.4 & 75.0 & 82.9 & 62.9 & 59.1 \\
    InternVL-2-4B~\citeyearpar{chen2024internvl}  & 90.9 & 84.1 & 80.1 & 84.1 & 72.7 & 66.7 \\
    Qwen2-VL-7B~\citeyearpar{wang2024qwen2vl}  & 91.9 & 86.5 & 82.6 & 83.8 & 77.7 & 69.7 \\
    Aria-UI~\citeyearpar{yang2024aria}      & --   & 87.7 & 67.3 & --   & 43.2 & 10.2 \\
    OS-Atlas-4B~\citeyearpar{wu2024osatlas}    & 91.9 & 83.8 & 80.6 & 84.7 & 73.8 & 67.5 \\
    OS-Atlas-7B~\citeyearpar{wu2024osatlas}    & 93.6 & 88.0 & 85.2 & 85.2 & 78.5 & 71.2 \\
    Aguvis-7B~\citeyearpar{xu2024aguvis}       & --   & --   & 80.5 & --   & --   & 61.5 \\
    Aguvis-72B~\citeyearpar{xu2024aguvis}      & --   & --   & 84.4 & --   & --   & 66.4 \\
    OS-Genesis-7B~\citeyearpar{sun2024osgenesis}    & 91.3   & --   & 74.2 & 66.2   & --   & 44.5 \\
    UI-TARS-2B~\citeyearpar{qin2025uitars}       & \textbf{98.1} & 87.3 & 89.3 & 81.2 & 78.4 & 68.9 \\
    UI-TARS-7B~\citeyearpar{qin2025uitars}      & \underline{98.0} & 89.3 & 90.8 & 83.7 & 80.5 & 72.5 \\
    UI-TARS-72B~\citeyearpar{qin2025uitars}     & \textbf{98.1} & 89.9 & 91.3 & 85.2 &81.5 & 74.7\\
    Qwen2.5-VL-3B~\citeyearpar{bai2025qwen25vl}   & -- & -- & 90.8 & -- & -- & 63.7 \\
    Qwen2.5-VL-7B~\citeyearpar{bai2025qwen25vl}   & -- & -- & 91.4 & -- & -- & 60.1 \\
    Qwen2.5-VL-32B~\citeyearpar{bai2025qwen25vl}   & -- & -- & \underline{93.3} & -- & -- & 69.6 \\
    Qwen2.5-VL-72B~\citeyearpar{bai2025qwen25vl}   & -- & -- & \textbf{93.7} & -- & -- & 67.4 \\
    InternVL3.5-241B-A28B~\citeyearpar{wang2025internvl35}   & 88.1 & 93.4 & 82.1 & 81.0 & 81.5 & 68.2 \\
    \midrule
    \modelname{3} & 91.4 & \underline{93.7} & 84.1 & 81.4 & \underline{83.9} & 70.3\\
    \modelname{7} & 93.3 & 93.1 & 86.0 & \underline{86.3} & \textbf{84.3} & \underline{74.8} \\
    \modelname{32} & 91.9 & \textbf{94.7} & 85.7 & \underline{85.7} & \textbf{87.3} & \textbf{75.9}  \\
    \bottomrule
  \end{tabular}
  } % End resizebox
\end{table}
\renewcommand{\arraystretch}{1.2} 
\begin{table*}[!t]
  \caption{Performance comparison on ScienceBoard~\citep{sun2025scienceboard}.}

  \label{tab:scienceboard}
  \centering
  \small
\resizebox{0.85\textwidth}{!}{
\begin{tabular}{lccccccc}
\toprule
\textbf{Model} & \textbf{Algebra} & \textbf{Biochem} & \textbf{GIS} & \textbf{ATP} & \textbf{Astron} & \textbf{Doc} & \textbf{Overall} \\
\midrule
GPT-4o~\citeyearpar{hurst2024gpt} & 3.2  & 0.0  & 0.0  & 0.0  & 0.0  & \underline{6.3}  & 1.6  \\
Claude-3.7-Sonnet~\citeyearpar{anthropic2025claude37} & 9.7  & \textbf{37.9} & \underline{2.9}  & 0.0  & 6.1  & \underline{6.3}  & \underline{10.5}  \\
Gemini-2.0-Flash~\citeyearpar{gemini2_0} & 6.5  & 3.5  & \underline{2.9}  & 0.0  & 0.0  & 6.1  & 3.2  \\
Qwen2.5-VL-72B~\citeyearpar{bai2025qwen25vl} & \textbf{22.6}  & \underline{27.6}  & \textbf{5.9}  & 0.0  & \underline{9.1}  & \textbf{12.5}  & \textbf{12.9} \\
InternVL3-78B~\citeyearpar{zhu2025internvl3} & 6.5  & 3.5  & 0.0  & 0.0  & 0.0  & \underline{6.3}  & 2.7  \\
UI-TARS-1.5-7B~\citeyearpar{qin2025uitars} & \underline{12.9}  & 13.8  & 0.0  & 0.0  & 6.1  & 0.0  & 5.5  \\
\midrule
\modelname{3} & 6.5  & 13.8  & 0.0  & 0.0  & 0.0  & 0.0  & 3.6  \\
\modelname{7} & 3.2  & 3.4  & 0.0  & 0.0  & 1.8  & 0.0  & 1.8  \\
\modelname{32} & 9.7  & 10.3  & 0.0  & 0.0  & \textbf{12.1}  & 0.0  & 5.9  \\
\bottomrule
\end{tabular}
}
\end{table*}
\begin{table}[!t]
  \centering
  \caption{Performance on General VLM Benchmarks. ScaleCUA-3B (25\%) denotes that, during training, the number of general-purpose data samples was set to 25\% of the GUI data samples (\eg, Understanding, Grounding, and Planning).}
  \small 
  \resizebox{\textwidth}{!}{
    % \begin{tabularx}{l*{5}{>{\centering\arraybackslash}X}>{\centering\arraybackslash}X}
    \begin{tabular}{lcccc}
    \toprule
    \textbf{Model} & \textbf{MathVista$_{\textbf{MINI}}$}~\citeyearpar{lu2023mathvista} & \textbf{OCRBench}~\citeyearpar{liu2024ocrbench} & \textbf{MMMU$_{\text{valid}}$}~\citeyearpar{yue2024mmmu} & \textbf{RealWorldQA}~\citeyearpar{xai-grok15v-2024} \\
    \midrule
    Qwen2.5-VL-3B~\citeyearpar{bai2025qwen25vl} & \textbf{62.3} & 797~(79.7) & 53.1 & \textbf{65.4} \\
    ScaleCUA-3B (0\%) & 52.8 & 819~(81.9) & 48.8 & 65.2 \\
    ScaleCUA-3B (25\%) & 58.6 & 823~(82.3) & 50.6 & \textbf{65.4} \\
    ScaleCUA-3B (50\%)& 58.7 & \textbf{824~(82.4)} & 52.4 & 65.1 \\
    ScaleCUA-3B (75\%) & 59.3 & 818~(81.8) & \textbf{55.6} & 65.2 \\
    ScaleCUA-3B (100\%) & 60.6 & 806~(80.6) & 53.4 & 63.5\\
    \hline
    Qwen2.5VL-7B~\citeyearpar{bai2025qwen25vl} & \textbf{68.2} & \textbf{864~(86.4)} & \textbf{58.6} & 68.5 \\
    \modelname{7} (50\%) & 65.4 & 852~(85.2) & 54.7 & \textbf{69.8} \\
    \hline
    Qwen2.5-VL-32B~\citeyearpar{bai2025qwen25vl} & \textbf{74.7} & \textbf{854~(85.4)} & \textbf{70.0} & 72.2 \\
    \modelname{32} (75\%) & 69.8 & 827~(82.7) & 61.9 & \textbf{72.3} \\
    % \hline
    % QWen2.5VL-72B\citep{bai2025qwen25vl} & & & \\
    % \modename{72} (50\%) &  & & \\
    \bottomrule
  % \end{tabularx}
  \end{tabular}
  }
  \label{tab:results_general_evaluation}
\end{table}
To fully demonstrate the potential of \modelname, we provide additional results on serveral benchmarks.

On \textbf{MMBench-GUI L2}~\citep{wang2025mmbenchgui}, which incorporates stratified grounding difficulty across major operating systems, \modelname{32} demonstrates performance comparable to state-of-the-art methods as shown in Table~\ref{tab:mmbenchgui-l2}. It achieves leading scores in the basic difficulty setting across several platforms including Android (96.4), Web (93.9), Linux (81.2), and macOS (88.1), while maintaining competitive results in the advanced difficulty setting (e.g., Web 76.3, Android 81.7). Furthermore, \modelname{7} and \modelname{3} achieve average scores of 78.2 and 73.7, respectively. They demonstrate particularly robust performance in the basic difficulty setting, especially on Windows, where both models score 78.6, and on iOS, with respective scores of 96.1 and 93.0.

On \textbf{OSWorld-G}~\citep{xie2025scalingcomputerusegroundinguser} for Ubuntu grounding, \modelname{32} demonstrates impressive results with an overall performance of 60.6 shown in Table~\ref{tab:osworld-g},  which includes strong marks in layout understanding (70.0), element recognition (66.7), and fine-grained manipulation (51.0). All of our models underperform on the Refusal subtask because we deliberately excluded the Refusal-specific training data provided by JEDI~\citep{xie2025scalingcomputerusegroundinguser}. Incorporating these examples may pose a risk of biasing the model toward emitting an await/refusal state in complex grounding scenarios. Such bias diminishes the agent’s propensity for active exploration within the environment, thereby degenerating its success rate in task completion.

On \textbf{AndroidControl}~\citep{li2024effects} which is an offline planning benchmark developed for the Android, all \modelname variants exhibit consistently strong performance demonstrated in Table~\ref{tab:androidcontrol_comparison}. On the AndroidControl-Low, \modelname{7} attains the highest task completion rate, whereas \modelname{32} achieves the most reliable grounding, indicating that the compact model favors execution efficiency while the larger capacity maximizes perceptual fidelity. As for AndroidControl-High, \modelname{32} demonstrates the highest success rate while showing the smallest degradation from Low to High. \modelname{3} and \modelname{7} achieve a favorable trade-off, sustaining solid performance across both low and high settings. The relatively small variance in type prediction across sizes suggests that residual failures arise more from long-horizon interaction and error accumulation than from intent misclassification or localization.

On \textbf{ScienceBoard}~\citep{sun2025scienceboard}, a planning benchmark designed for scientific professionals, our models show modest yet meaningful capability as shown in ~\ref{tab:scienceboard}. The \modelname{32} outperforms strong VLMs such as GPT-4o (1.6) while remaining below Qwen2.5-VL-72B (12.9) and Claude-3.7-Sonnet (10.5). Our model excels in domains demanding factual and visual-text reasoning over those requiring specialized symbolic workflows.

To evaluate the transfer learning capabilities of \modelname{32}, we augment our training with a diverse set of general-purpose SFT data. These data, sourced from the post-training corpus of InternVL3~\citep{zhu2025internvl3}, encompass a range of tasks, including OCR, mathematics, coding, reasoning-QA, and general multimodal understanding. We then assess performance on four standard \textbf{General Multimodal Benchmarks} shown in Table~\ref{tab:results_general_evaluation}. These benchmarks jointly evaluate skills such as mathematical and commonsense reasoning, text comprehension, and open-domain visual question answering, which are also fundamental for computer-use agents. The ``\modelname{3} (25\%)" specifies the proportion of this general-purpose data relative to the core GUI data used in training. 

Based on Table~\ref{tab:results_general_evaluation}, several consistent trends emerge regarding the interaction between the proportion of general-purpose data and agent performance on general VLM benchmarks. First, incorporating moderate amounts of general-purpose data (e.g., 25–50\% relative to GUI-specific data) yields notable gains over the 0\% setting, particularly on MathVista and MMMU$_{\text{valid}}$, suggesting that exposing the agent to broader multimodal reasoning tasks improves its mathematical and cross-domain inference ability. For instance, \modelname{3} rises from 52.8 to 58.7 on MathVista and from 48.8 to 52.4 on MMMU when increasing general data to 50\%, while maintaining stable performance on RealWorldQA. Second, the results indicate a saturation effect: pushing the general data ratio to 75\% or 100\% offers only marginal or inconsistent benefits. Third, scaling model capacity amplifies the positive effect of general data. 
However, our 7B and 32B models still exhibit a substantial performance gap compared to the baseline on general benchmarks, indicating that the proportion of general-purpose data could be further increased. Such adjustments must also consider their potential impact on the computer-use capability of agent models.

\subsection{Public Data Used in Training}
Table~\ref{tab:public_data} summarizes the public datasets used for training \modelname. Please note that the reported statistics refer to the portion of each dataset actually utilized in our experiments, rather than the original sizes of the source datasets.
% \begin{table}[t]
% \centering
% \small
% \begin{tabular}{l|l|c}
% \mytoprule
% \textbf{Task} & \textbf{Data source} & \textbf{\#Elements / Trajectory} \\
% \mymidrule
% \multirow{9}{*}{Grounding} 
%  & SeeClick~\citeyearpar{cheng2024seeclick} & 271K \\
%  & GUIEnv~\citeyearpar{chen2024guicourse} & 328K \\
%  & GUIAct~\citeyearpar{chen2024guicourse} & 67K \\
%  & WebUI~\citeyearpar{wu2023webui} & 57K \\
%  & Widget Captioning~\citeyearpar{li2020widget} & 101K \\
%  & RicoSCA~\citeyearpar{li2020mapping} & 173K \\
%  & UI RefExp~\citeyearpar{bai2021uibert} & 16K \\
%  & RICO Icon~\citeyearpar{deka2017rico} & 16K \\
%  & OmniACT~\citeyearpar{kapoor2024omniact} & 7K \\
%  & Jedi~\citeyearpar{xie2025scalingcomputerusegroundinguser} & 7K \\
% \cline{2-3}
%  & \textbf{Total} & \textbf{1.036M} \\
% \mymidrule
% \multirow{8}{*}{Planning} 
%  & MM-Mind2Web~\citeyearpar{deng2023mind2web} & 1,009 \\
%  & GUIAct~\citeyearpar{chen2024guicourse} & 2,482 \\
%  & MiniWoB++~\citeyearpar{Tianlin2017miniwob} & 2,762 \\
%  & AitZ~\citeyearpar{zhang2024aitz} & 1,987 \\
%  & AndroidControl~\citeyearpar{li2024effectsdatascalecomputer} & 13,594 \\
%  & GUI Odyssey~\citeyearpar{lu2024gui_odyssey} & 7,735 \\
%  & AMEX~\citeyearpar{chai2024amex} & 2,991 \\
%  & AitW~\citeyearpar{rawles2023androidinthewild} & 2,346 \\
%  & AitW~\citeyearpar{rawles2023androidinthewild} & 2,346 \\
% \cline{2-3}
%  & \textbf{Total} & \textbf{35K} \\
% \mybottomrule
% \end{tabular}
% \caption{Data sources used for grounding and planning tasks.}
% \label{tab:data_sources}
% \end{table}

\begin{table}[t]
\caption{Public data sources used for training our models. The ``*'' indicates that we count the number of samples we use rather than the full size of the original dataset because we have processed several datasets and filtered some noisy examples. Some statistics are calculated from Aguvis~\citeyearpar{xu2024aguvis}. }
\label{tab:public_data}
\centering
\small
\resizebox{0.75\textwidth}{!}{
\begin{tabular}{l|l|l|c}
\mytoprule
\textbf{GUI Task} & \textbf{Data source} & \textbf{Platform} & \textbf{\#Elements / \#Steps} \\
\mymidrule
\multirow{10}{*}{Grounding} 
 & SeeClick~\citeyearpar{cheng2024seeclick} & Website & 271K \\
 & GUIEnv~\citeyearpar{chen2024guicourse} & Website & 328K \\
 & GUIAct~\citeyearpar{chen2024guicourse} & Website & 67K \\
 & WebUI~\citeyearpar{wu2023webui} & Website & 57K \\
 & Widget Captioning~\citeyearpar{li2020widget} & Mobile & 101K \\
 & RicoSCA~\citeyearpar{li2020mapping} & Mobile & 173K \\
 & UI RefExp~\citeyearpar{bai2021uibert} & Mobile & 16K \\
 & RICO Icon~\citeyearpar{deka2017rico} & Mobile & 16K \\
 & OmniACT~\citeyearpar{kapoor2024omniact} & Desktop \& Website & 7K \\
 & UGround~\citeyearpar{gou2024navigating}* & Website & 1404K \\
 & OS-Atlas~\citeyearpar{wu2024osatlas}* & Desktop & 799K \\
 & JEDI~\citeyearpar{xie2025scalingcomputerusegroundinguser}* & Desktop & 550K \\
\cline{2-4}
 & \textbf{Total} & -- & \textbf{3789K} \\
\mymidrule
\multirow{8}{*}{Planning} 
 & MM-Mind2Web~\citeyearpar{deng2023mind2web} & Website & 7.8K \\
 & GUIAct~\citeyearpar{chen2024guicourse} & Website & 16.6K \\
 & MiniWoB++~\citeyearpar{Tianlin2017miniwob} & Website & 9.9K \\
 & AitZ~\citeyearpar{zhang2024aitz} & Mobile & 11.9K \\
 & AndroidControl~\citeyearpar{li2024effectsdatascalecomputer} & Mobile & 74.8K \\
 & GUI Odyssey~\citeyearpar{lu2024gui_odyssey} & Mobile & 118.3K \\
 & AMEX~\citeyearpar{chai2024amex} & Mobile & 35.6K \\
 & AitW~\citeyearpar{rawles2023androidinthewild} & Mobile & 19.0K \\
 & PC Agent-E~\citeyearpar{he2025efficientagenttrainingcomputer} & Desktop & 27.8K \\
\cline{2-4}
 & \textbf{Total} & -- & \textbf{321.7K} \\
\mybottomrule
\end{tabular}
}
\end{table}

\subsection{Action Space}
\label{sec:action_space}

\begin{table}[bthp]
\footnotesize
\caption{Actions space.}  
\label{tab:action_space}
\centering
\resizebox{1\linewidth}{!}{%
\begin{tabular}{l|c|p{8cm}}
\toprule
Action & Platforms & Description \\
\midrule
\texttt{click(x, y, clicks, button)} & All & Perform a mouse click at coordinates (x, y) using the specified button and number of clicks. \\
\texttt{write(message)} & All &  Input the given message. \\
\texttt{wait(seconds)} & All & Pause execution for the specified number of seconds. \\
\texttt{response(answer)} &  All & Submit a response to the environment or task prompt. \\
\texttt{terminate(status)} &  All & Terminate the current task with a given completion status. \\
\texttt{scroll(clicks, x, y)} & Desktop & Performs a scroll of the mouse scroll wheel at position (x, y). \\
\texttt{doubleClick(x, y, button)} & Desktop \& Web & Perform a double click at coordinates (x, y) with the specified button. \\
\texttt{rightClick(x, y, button)} & Desktop \& Web & Perform a right click at coordinates (x, y) with the specified button. \\
\texttt{hotkey(*args)} & Desktop \& Web & Trigger a keyboard shortcut composed of one or more keys. \\
\texttt{moveTo(x, y)} & Desktop \& Web & Move the mouse pointer to the specified (x, y) position. \\
\texttt{dragTo(x, y, button)} & Desktop \& Web & Drag the mouse to (x, y) while holding the specified button. \\
\texttt{press(keys, presses)} & Desktop \& Web & Press the specified key(s) a given number of times. \\
\texttt{keyDown(key)} & Desktop \& Web & Press and hold a key without releasing it. \\
\texttt{keyUp(key)} & Desktop \& Web & Release a previously held key. \\
\texttt{swipe(from, to, direction, amount)} & Mobile \& Web & Swipe from a start to end point in the specified direction with a given intensity or distance. \\
% \texttt{go\_forward()} & Web & Navigate forward in the browser history. \\
% \texttt{go\_backward()} & Web & Navigate backward in the browser history. \\
% 
\texttt{navigate\_home()} & Mobile & Return to the mobile home screen. \\
\texttt{navigate\_back()} & Mobile & Navigate back to the previous screen on mobile. \\
\texttt{long\_press()} & Mobile & Perform a long-press gesture on the current focus or location. \\
\texttt{open\_app(app\_name)} & Mobile & Launch a mobile application by its name. \\
\bottomrule
\end{tabular}
}
\end{table}
To enable robust cross-platform control, we define a unified action space that abstracts low-level GUI actions into a concise yet expressive set of semantic commands. As shown in Table~\ref{tab:action_space}, this action space is designed to be platform-aware yet semantically consistent, allowing our agents to operate seamlessly across Desktop (Windows, macOS, Ubuntu), Mobile (Android, iOS), and Web platforms. The action set includes universally supported operations such as \texttt{click}, \texttt{write}, \texttt{wait}, and \texttt{terminate}, which are shared across all platforms. It also accommodates platform-specific interactions, including \texttt{swipe} and \texttt{long\_press} for mobile devices, and fine-grained mouse or keyboard controls such as \texttt{doubleClick}, \texttt{rightClick}, \texttt{dragTo}, and \texttt{hotkey} for desktop and web interfaces. To handle modern interactive elements, \texttt{swipe} operation has also been implemented for Web.
By standardizing the operation interface through a shared action space, we simplify training and inference while supporting both generalization and specialization. Each action is defined with explicit arguments (e.g., coordinates, keypresses), enabling precise control and compatibility with structured outputs in grounding, direct-action, and reasoned-action inference modes. This design facilitates modular training, policy transfer, and scalable data annotation, forming a critical foundation for developing universal GUI agents.

\subsection{Error Case Analysis}
\begin{figure*}[!bh]
  \centering
  \includegraphics[width=0.9\linewidth]{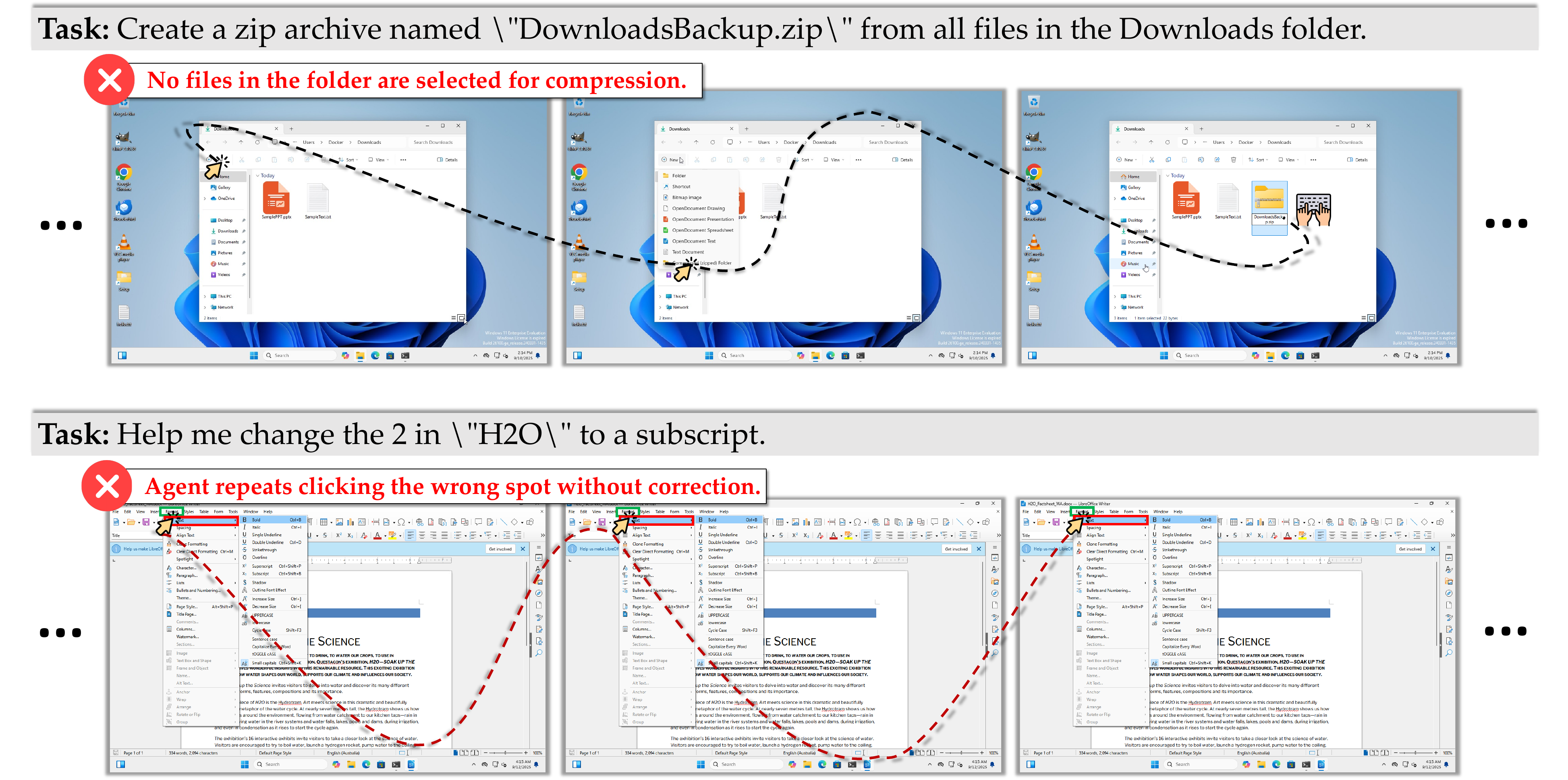}
    \vspace{-0.5em}
\caption{Error cases on the Windows platform. The first case shows \modelname creating an archive without having selected any files, revealing that it sometimes fails to follow the full instruction and only completes a sub-step. The second case shows \modelname persistently repeating the same action until the step limit, when it misses the correct element and the screen remains unchanged.}
\label{fig:waa_error_case}
  \vspace{-0.75em}
\end{figure*}

\begin{figure*}[!tbh]
  \centering
  \includegraphics[width=0.9\linewidth]{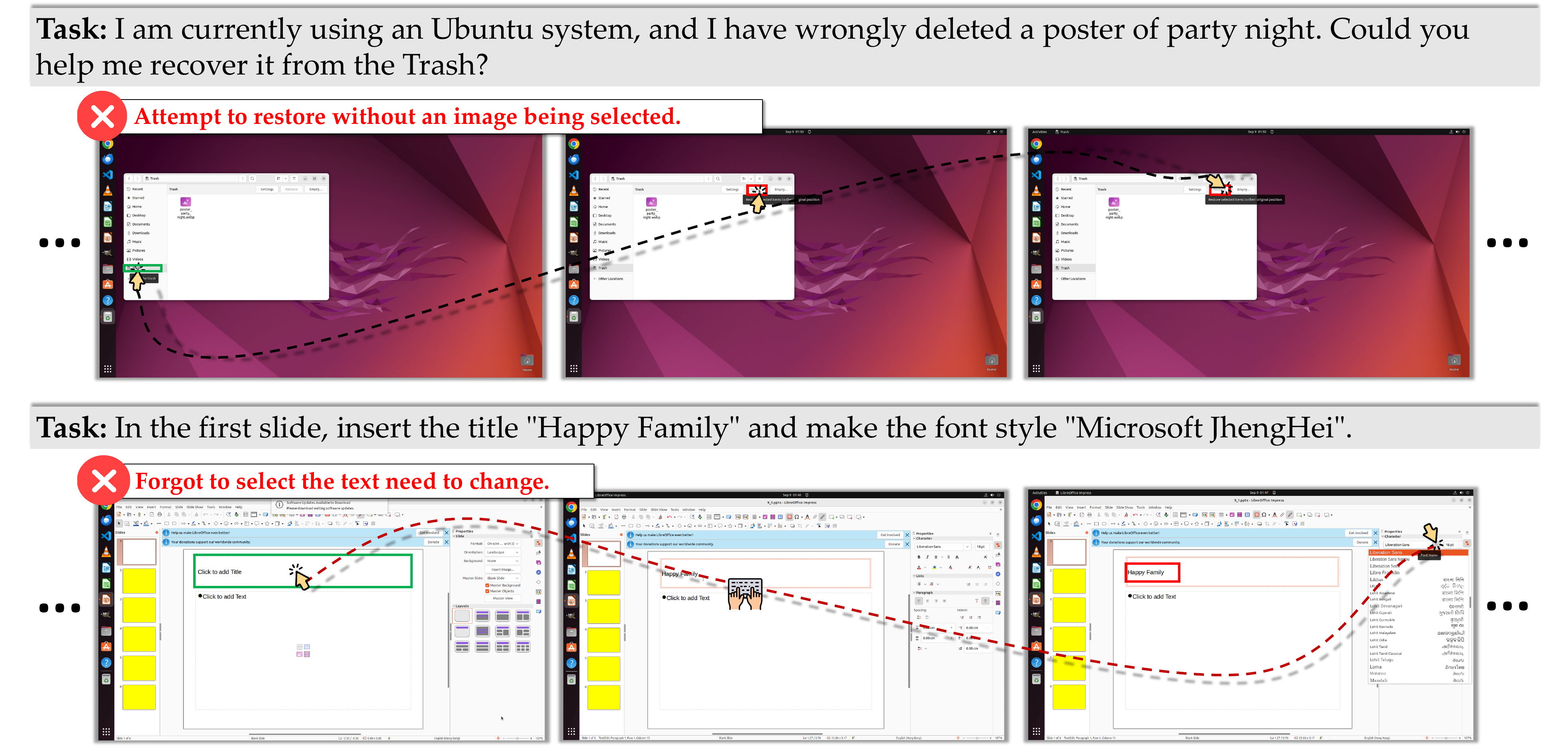}
    \vspace{-0.5em}
\caption{Error cases on the Ubuntu platform. \modelname repeatedly fails tasks because it does not comprehend procedural prerequisites. The agent attempts to execute a final command without first performing the necessary intermediate step of selecting the target object. For instance, it tries to restore a file without selecting it from the trash or alter a font without highlighting the text. Critically, this operational flaw generates no explicit error message, trapping the agent in a repetitive loop of ineffective actions.}
\label{fig:ubuntu_error_case}
  \vspace{-0.75em}
\end{figure*}
% android error case
\begin{figure*}[!tbh]
  \centering
  \includegraphics[width=0.9\linewidth]{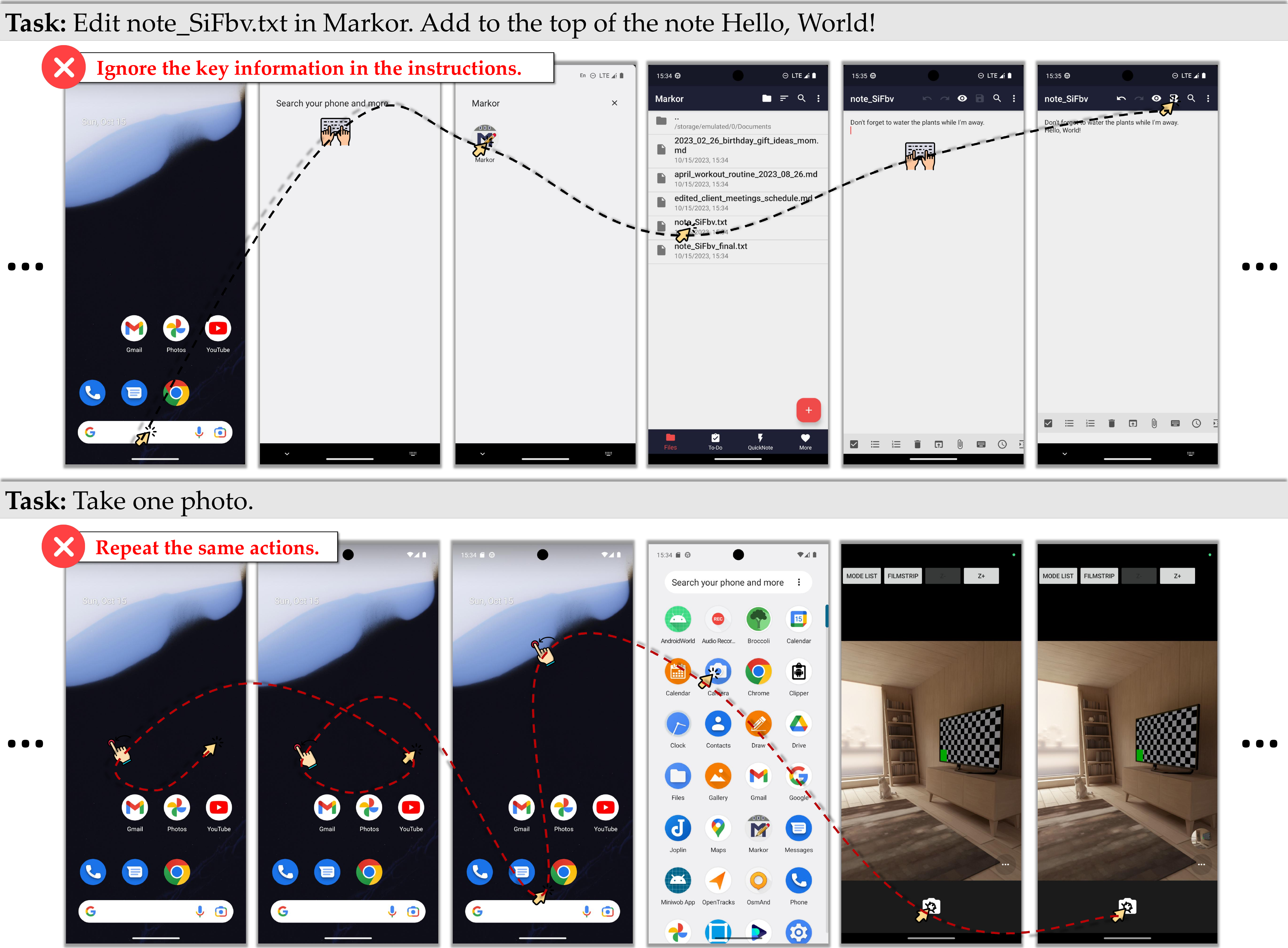}
    \vspace{-0.5em}
\caption{Error cases on the Android platform. The first case shows an instruction requiring content to be inserted at the top of a document; however, \modelname opens the file and inserts directly at the current cursor location, ignoring the positional prerequisite. The second case shows that when the UI exhibits no obvious state change after an operation, \modelname repeats the same action multiple times, causing tasks such as taking a photo to fail.}
\label{fig:android_error_case}
  \vspace{-0.75em}
\end{figure*}
We here provide several error cases across different platforms to analyze the limitations of our \modelname.

On desktop platforms, \modelname frequently violates \textit{procedural prerequisites} shown in  Fig~\ref{fig:waa_error_case} and Fig~\ref{fig:ubuntu_error_case}, such as attempting to compress files without selecting them or changing font styles without highlighting the target text. These issues stem from an incomplete understanding of interface states and sub-task dependencies. Moreover, a significant limitation of \modelname emerges when actions result in silent failures, characterized by a lack of discernible state transition. In such instances, the model tends to perseverate on the unsuccessful operation, revealing the absence of a robust error-recovery mechanism. This issue underscores the critical requirement for fine-grained perception and a robust understanding of element state to interact with context-dependent UI elements, such as focus and selection.
% web error case
\begin{figure*}[!htbh]
  \centering
  \includegraphics[width=\linewidth]{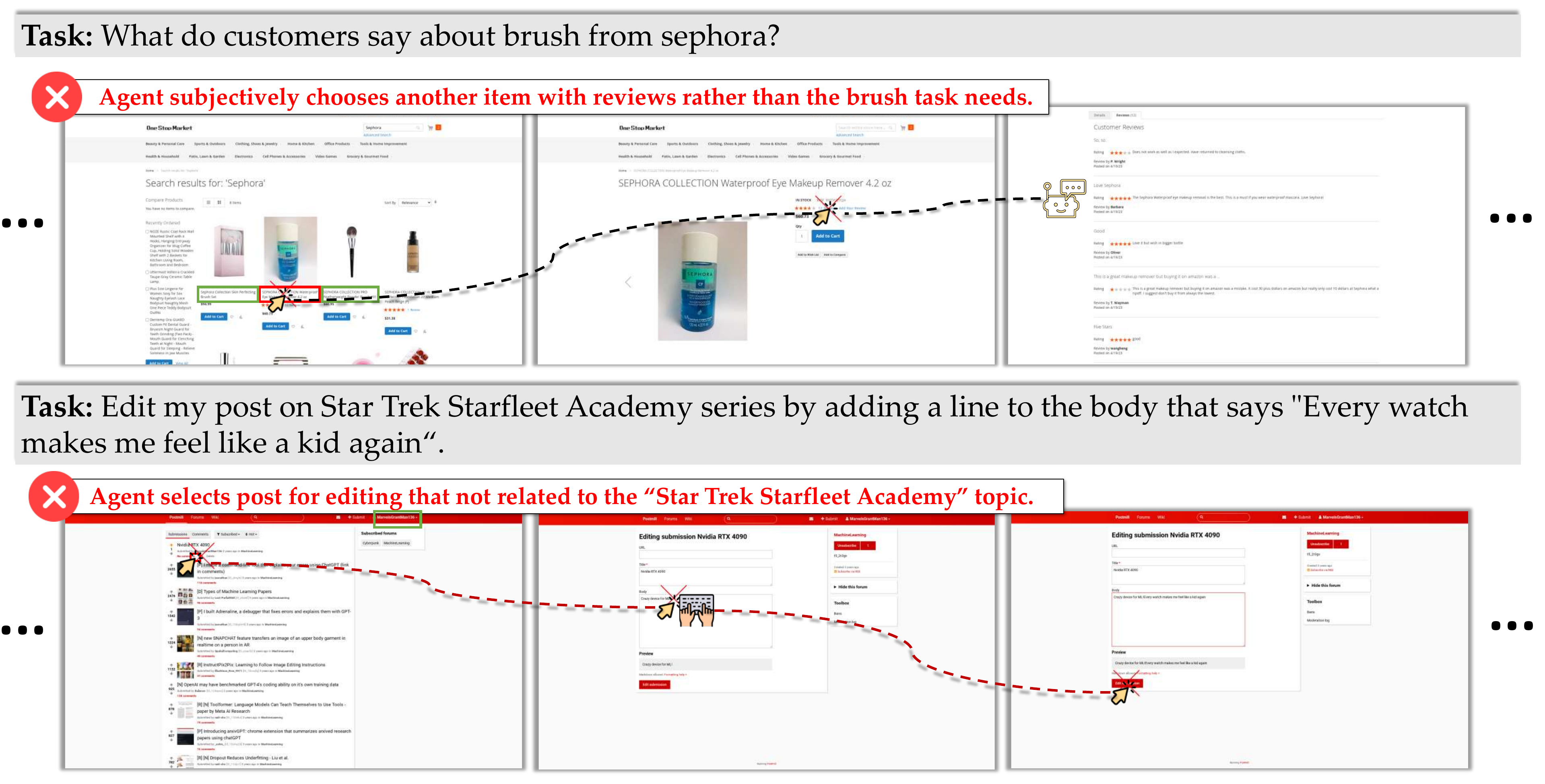}
    \vspace{-0.5em}
\caption{Error cases on the Web platform. The first case shows \modelname made subjective analytical assumptions, presuming the product necessarily contained reviews, while disregarding the explicitly specified product category in the task instructions. The second case shows \modelname struggles with complex tasks in complex initial environments (where numerous posts already exist on the starting interface). When faced with multifaceted requirements (needing to identify both "my" posts and posts on a specified topic), it neglected the explicitly stated topic in the instructions, instead selecting only posts visible in the current observation space that belonged to me.}
\label{fig:web_error_case}
  \vspace{-2em}
\end{figure*}

For the Android platform, there exist precision and positional challenges demonstrated in Fig~\ref{fig:android_error_case}. In the first case, the instruction explicitly requires appending text to the top of a file within a note-taking application (Markor). However, the agent fails to recognize this positional constraint, instead inserting content at the current cursor location without adjusting it. This suggests that \modelname lacks a fine-grained understanding of positional semantics in natural language instructions, as well as the ability to reason about UI state changes like cursor positioning. In the second case, the agent is instructed to "Take one photo." Despite correctly launching the camera and triggering the shutter once, the agent erroneously repeats the same actions multiple times. This behavior stems from a failure to detect visual feedback or confirm state transitions (e.g., a captured photo thumbnail), leading to unnecessary repetition. These failure modes indicate two key limitations: (1) insufficient grounding of spatial and contextual cues embedded in task descriptions, and (2) inadequate visual state tracking, particularly under conditions where UI feedback is subtle. Addressing these issues may require enhanced visual reasoning modules, memory-based state modeling, or task-guided grounding refinements.

Empirical analysis of trajectories from web platform reveals that \modelname may struggle with semantic disambiguation. \modelname often selects visually salient but instruction-inconsistent elements (e.g., wrong product category or unrelated post) as presented in Fig~\ref{fig:web_error_case}, revealing a bias toward superficial cues over explicit constraints like ownership (“my post”) or topical relevance (“Starfleet Academy”).  

To mitigate these issues, three avenues may show promise:
(1) \textbf{Reflection and State Verification.} Integrating lightweight screen-change detectors and visual precondition checkers can allow agents to validate action effects and avoid ineffective loops.
(2) \textbf{Reinforcement Learning with Recovery Signals.} Reward structures should penalize redundant, non-progressive behaviors and incentivize predicate satisfaction (e.g., ``text selected'', ``correct tab active'') before proceeding.
(3) \textbf{Memory-Augmented Planning.} By introducing episodic memory to recall past interactions (e.g., whether a menu opened successfully), the agent can reason across time and avoid retrying failed subgoals.

\subsection{The Details of Data Curation}
\label{sec:the_details_of_data_curation}
\subsubsection{Data Sources}
\label{sec:data_sources}
We systematically collect GUI data across diverse platforms to construct \ourdataset, including desktop, mobile, and web environments. As shown in Table.~\ref{tab:data_sources}, \ourdataset spans 7 major operation systems: Windows, Ubuntu, macOS, iOS, iPadOS, Android, and Web. Each platform features a broad spectrum of frequently used applications designed for productivity, communication, entertainment, browsing, and utilities.

On desktop platforms, Windows includes both native and third-party applications such as Microsoft Office Suite, Adobe Creative Cloud, Visual Studio, and system utilities, offering a comprehensive view of traditional GUI layouts. Ubuntu and macOS incorporate open-source and system software, including LibreOffice, GIMP, Terminal, Finder, and Safari.

Mobile data is collected from the iOS and Android platforms. The data from the iOS platform includes system applications such as Settings, Safari, Calendar, and Health, as well as third-party applications including Weibo, Notability, and Spotify. The Android platform, by virtue of its open ecosystem, serves as the greatest diversity of data sources, encompassing both system applications and a broad array of commercial software from domains such as productivity, e-commerce, social media, and multimedia (e.g., WeChat, Taobao, TikTok, and Google Suite).

For tablet interfaces, our data collection primarily focused on iPadOS, encompassing a selection of its most frequently utilized system applications.

As for Web, we collected pages from over 200 frequently accessed websites spanning e-commerce, social media, education, government services, travel, and developer tools. These sources encompass major websites such as Amazon, YouTube, Reddit, Wikipedia, Coursera, and GitHub, with data captured through both static DOM snapshots and dynamic interaction traces.

The collected dataset constitutes a high-coverage, cross-platform corpus of real-world graphical interfaces which endows the model with comprehensive domain knowledge and leads to significantly improved generalization.

\subsubsection{GUI Understanding}
To support the development of general-purpose computer use agents, we construct a large-scale corpus for GUI understanding that encompasses both element-level and screenshot-level semantics. This corpus is designed to facilitate fine-grained perception and reasoning over static and dynamic user interfaces.

For element-level understanding, we define five task formulations targeting visual appearance, spatial layout, textual grounding, and semantic functionality. First, we introduce the \textit{Element Appearance Captioning} task, which requires the model to describe visual features (e.g., shape, color, borders) of a given GUI component. These attributes often signal affordances and can help distinguish between interactive and static elements. Second, we incorporate \textit{Referring OCR}, a referring task where the model extracts the textual content within a specified bounding box, enabling alignment between visual context and embedded text. Third, to capture spatial organization, the \textit{Element Layout Understanding} task asks the model to predict both absolute screen coordinates and relative positions with respect to nearby components. Fourth, to understand the operational roles of components, we define the \textit{Element Functionality Captioning} task, where the model infers the intended function of a labeled element within its surrounding interface. Finally, we propose a \textit{User Intention Prediction} task, where the model is asked to infer the user's likely goal based on contextual clues and ongoing interactions.

For screenshot-level understanding, we formulate two tasks that promote global comprehension. The \textit{Interface Captioning} task prompts the model to generate a high-level textual description summarizing the overall structure, visual hierarchy, and content of the interface. This encourages holistic reasoning and layout recognition. Complementarily, the \textit{Screen Transition Captioning} task focuses on temporal changes by asking the model to describe the differences between two consecutive screenshots. This enables the model to understand GUI dynamics, such as state updates, navigation events, or content refreshes.

Together, these tasks define a comprehensive benchmark for GUI understanding. We leverage vision-language models to automatically generate annotations for both element-level and screenshot-level tasks, using visual context, structural metadata, and interaction histories. This corpus provides the foundation for training agents capable of fine-grained perception, robust grounding, and high-level reasoning in complex GUI environments.
\begin{center} 
\resizebox{\textwidth}{!}{
    \begin{minipage}{\textwidth} 
    \scriptsize 
    \renewcommand{\arraystretch}{1.2}
    \captionof{table}{The main sources of GUI corpora across different platforms.} 
    \label{tab:data_sources}
     \begin{tabularx}{\textwidth}{l|X} 
     \Xhline{1.5pt}
     \multicolumn{1}{l|}{\textbf{Platform}}
     & \multicolumn{1}{c}{\textbf{Application}}  \\
     \hline
    Windows & File Explorer, OS, Chrome, Microsoft Edge, Word, Excel, PowerPoint, LibreOffice Calc, LibreOffice Impress, LibreOffice Writer, Maps, Camera, Calculator, Microsoft Store, Clock, Photos, Outlook, Media Player, VLC Media Player, Calendar, Paint, Paint 3D, QQ Music, KuGou Music, Spotify, Tencent QQ, Visual Studio Code, Dev-C++, Microsoft Solitaire \& Casual Game, Pycharm, Android Studio, Vmware Workstation Pro, Vmware Fusion, Adobe Photoshop, Adobe Premiere Pro, Adobe Illustrator, Blender, FL Studio, Unreal Engine, DaVinci Resolve, AutoCAD, SolidWorks, Inventor, Vivado, MATLAB, Origin, Stata, Eviews \\
    \hline
    Ubuntu & Files, OS, Firefox, Chrome, LibreOffice Calc, LibreOffice Impress, LibreOffice Writer, OneNote, GIMP, Slack, Thunderbird, Visual Studio Code, Zotero    \\
    \hline
    MacOS & Finder, OS, Safari, Chrome, Pages, Numbers, Keynote, Calculator, Maps, Notes, Calendar, Contacts, Reminders, Apple Music, Podcasts, Weather, Stocks, Freeform, Terminal, Clock, Pycharm, Android Studio, App Store, Mail, Visual Studio Code \\
    \hline
    iOS & Weather, Maps, Find My, Settings, Stocks, Safari, Mail, Calendar, App Store, Home, Camera, Files, Wallet, Contacts, Shortcuts, Clock, Twitter, Weibo, Outlook, Reddit, Instagram, Notes, Keynote, Reminders, Notability, GoodNotes, Rednote, Translate, Calculator, Voice Memos, Shadowrocket, Music, Podcasts, Spotify, iTunes Store, Apple TV, Books, Zhihu, Health
    \\
    \hline
    iPadOS & Weather, Settings, Safari, Camera, Goodnotes, Translate, Notes, Freeform, Chrome\\
    \hline
    Android & Settings, Clock, Desktop Clock, Calendar, Contacts, Files,  Camera, LinkedIn, Weibo, Twitter, Tieba, Reddit, Zoom, Gmail, Duolingo, Xueersi, Wikipedia, XuetangX, edX, Coursera, Skillshare, ZLibrary, To Do, Word, Excel, PowerPoint, OneNote, Taskade, Notion, TickTick, Google Maps, AMap, Tencent Map, Qunar, Trip.com, Ctrip, Qunar, LY.com, Fliggy, Zhixing Train Tickets, Map.me, Booking, Amazon, eBay, Taobao, Alipay, Poizon, VIPShop, 58.com, Beike, Anjuke, Zhuanzhuan, Douyin Mall, Shihuo, Nike, Bilibili, Bilibili CN, QQ Music, himalaya, Classical Music, News, Toutiao, Sohu News, NetEase News, Hupu, Huya, Sohu Video, Pi Music Player, NetEase Cloud Music, Kuaishou, Kugou, WeSing, Douban, Xiaohongshu, Zhihu, Qidian, Xiaoheihe, Prime Video, CNN, Quora, Cantook, Spotify, Apple Music, YouTube, Fitness, Health, JD Health, Translate, Moji Weather, App Store, Google Chrome, BlueCoins, VPN, Shadowrocket, Surfboard, Speedtest, Meitu, Jianying, Canva, Procreate, Pinterest, GitHub, DeepSeek, Grok
    \\
    \hline
    Web & 5i5j(sh.5i5j.com), AccuWeather(accuweather.com), adidas China(adidas.com.cn), Adobe(adobe.com), Amazon(amazon.com), American Kennel Club(akc.org), Apple(apple.com), arXiv(arxiv.org), BabyCenter(babycenter.com), Baidu(baidu.com), Baidu Baike(baike.baidu.com), Baidu Tieba(tieba.baidu.com), Beihang University(buaa.edu.cn), Bilibili(bilibili.com), BoardGameGeek(boardgamegeek.com), BoardMix(boardmix.cn), Booking.com(booking.com), Budget(budget.com), Cambridge Dictionary(dictionary.cambridge.org), Cars.com(cars.com), CNBlogs(cnblogs.com), CNN(cnn.com), CoinMarketCap(coinmarketcap.com), Coursera(coursera.org), CSDN(csdn.net), Ctrip(ctrip.com), Damai(damai.cn), Dianping(dianping.com), Dior(dior.com), Douban(douban.com), Douyin(douyin.com), Drugs.com(drugs.com), eBay(ebay.com), Britannica(britannica.com), ePay(epay.com), Epicurious(epicurious.com), Facebook(facebook.com), Fastly(fastly.com), FedEx(fedex.com), Fliggy(fliggy.com), Food Network(foodnetwork.com), Gaode Maps(gaode.com), Gmail(gmail.com), GitHub(github.com), Google Finance(finance.google.com), Google Maps(map.google.com), Google Scholar(scholar.google.com), GOV.UK(gov.uk), Healthline(healthline.com), Hugging Face(huggingface.co), Hupu(hupu.com), IGN(ign.com), IMDb(imdb.com), Indeed UK(uk.indeed.com), iQiyi(iqiyi.com), JD.com(jd.com), JetBrains(jetbrains.com), KAYAK(kayak.com), Kohl's(kohls.com), Last.fm(last.fm), LeetCode(leetcode.cn), LinkedIn(linkedin.com), Marriott(marriott.com), Microsoft Azure(azure.microsoft.com), Microsoft Office(office.com), ModelScope(modelscope.cn), MSN(msn.com), NBA(nba.com), National Relocation(nationalrelocation.com), NetEase Cloud Music(music.163.com), Newegg(newegg.com), OpenStreetMap(openstreetmap.org), PayPal(paypal.com), PJLab GitLab(gitlab.pjlab.org.cn), QQ(qq.com), QQ Music(y.qq.com), QS China(qschina.cn), Reddit(reddit.com), Redfin(redfin.com), REI(rei.com), Rotten Tomatoes(rottentomatoes.com), Ryanair(ryanair.com), Samsung(samsung.com), Shimo(shimo.im), Sina News(news.sina.com.cn), Skype(skype.com), SpotHero(spothero.com), Stack Overflow(stackoverflow.com), Steam Store(store.steampowered.com), Student.com(student.com), TensorFlow(tensorflow.org), Tencent Docs(docs.qq.com), Tencent Video(v.qq.com), The Weather Channel(weather.com), The Weather Network(theweathernetwork.com), Thumbtack(thumbtack.com), Ticket Center(ticketcenter.com), Trip.com US(us.trip.com), TripAdvisor(tripadvisor.com), UNIQLO China(uniqlo.cn), United Airlines(united.com), University of Cambridge(cam.ac.uk), University of Michigan(umich.edu), Vmall(vmall.com), Virginia DMV(dmv.virginia.gov), WebArena Forum(wa\_forum), WebArena GitLab(wa\_gitlab), WebArena Shopping(wa\_shopping), WebArena CMS(wa\_shopping\_admin), WebMD(webmd.com), Weibo(weibo.com), Wikipedia(wikipedia.org), WolframAlpha(wolframalpha.com), X(x.com), Xiaohongshu(xiaohongshu.com), Yahoo Finance(finance.yahoo.com), Yahoo Sports(sports.yahoo.com), Yelp(yelp.com), YouTube(youtube.com), Zhihu(zhihu.com), Zhaopin(i.zhaopin.com), Zhaopin Landing Page(landing.zhaopin.com), Zhipin(zhipin.com) and  $\sim$ 0.2M URLs selected from TOP-1M URLs(\textit{https://tranco-list.eu/})  \\
    \bottomrule
    \end{tabularx}
    \end{minipage}%
}
\end{center}

\subsubsection{Metadata Extraction}
\textbf{Windows Platform.} To facilitate the automated analysis and interaction with graphical user interfaces (GUIs), we design and implement a framework for extracting UI metadata on the Windows operating system. The core of this framework leverages the UI Automation~(UIA) technology to perform a depth-first traversal of an application's A11y Trees, initiated from the foreground window identified via native \texttt{Win32}\footnote{https://learn.microsoft.com/en-us/windows/win32/} API calls. Subsequently, the collected raw data undergoes a multi-stage filtering and refinement pipeline to ensure its relevance and actionability. This pipeline first performs a geometric validity check to filter out improperly sized or off-screen controls, followed by a visibility and occlusion analysis to retain only the topmost, unobscured elements. Furthermore, a semantic pruning module uses a predefined keyword list (e.g., ``close'', ``save'') to remove controls that might cause task interruption, while a system component exclusion module discards elements within standard OS regions like the taskbar based on their absolute coordinates. Each element that successfully passes through this pipeline is then abstracted into a structured JSON object. This object encapsulates its multi-dimensional attributes, including identity properties~(\texttt{control\_type}, \texttt{name}), state information~(\texttt{is\_enabled}), spatial coordinates~(\texttt{bbox}), and descriptive text~(\texttt{description}, \texttt{tooltip}). The aggregation of these objects yields a comprehensive metadata representation of the UI, establishing the foundation for subsequent automated tasks.

\textbf{Ubuntu Platform.} To extract Ubuntu metadata, we process an XML string representation of the A11y Trees, leveraging Python's built-in \texttt{xml} library for parsing\footnote{https://docs.python.org/3/library/xml.etree.elementtree.html}. The process commences by parsing the raw XML data into a tree structure. Following this, we linearize these nodes into structural elements. Specifically, for each node in this set, we programmatically extract key attributes, including its tag (representing the element's role), name, class, and description. To capture the semantic content robustly, the element's text is derived either directly from its text content or inferred from its value attribute, particularly for input fields. Positional and dimensional data are extracted from screencoord and size attributes, which together define the element's bounding box. The final output is a structured, tab-separated string where each line represents a single UI element. This entry is composed of seven fields: (1) \texttt{tag} indicating the UI type, (2) \texttt{name} for the element's given name, (3) \texttt{text} capturing its content or value, (4) \texttt{class} specifying its component class, (5) \texttt{description} for accessibility-related details, (6) \texttt{position} as a top-left (x, y) coordinate, and (7) \texttt{size} as a width and height pair. In essence, this process distills raw, platform-specific A11y Trees into a flattened, semantically-annotated dataset, providing a crucial foundation for downstream understanding, grounding tasks.

\textbf{MacOS Platform.} We extract UI metadata from macOS applications by leveraging the macOS Accessibility API, primarily via the \texttt{ApplicationServices}\footnote{https://developer.apple.com/documentation/applicationservices} frameworks. It allows structured traversal of the A11y Trees by programmatically accessing on-screen UI windows and querying attributes such as \texttt{AXPosition}, \texttt{AXSize}, \texttt{AXRole}, \texttt{AXTitle}, \texttt{AXValue}, and \texttt{AXDescription}. To initiate the process, we identify top-level windows from the system window list using \texttt{CGWindowListCopyWindowInfo}, filter for visible application windows, and create AX references using \texttt{AXUIElementCreateApplication}. A recursive collection strategy is then applied, traversing each window's A11y Trees up to a bounded depth while filtering out off-screen or irrelevant elements. To ensure semantic clarity, we enrich metadata by inferring contextual labels for interactive elements (e.g., \texttt{AXButton}, \texttt{AXTextField}) based on their surrounding static text, spatial layout, and role. Further, we apply spatial deduplication heuristics to eliminate overlapping or redundant elements, and merge content-bearing \texttt{AXStaticText} regions with their parent interactive widgets when appropriate. The final output is a flattened list of UI elements, each annotated with role, text content, description, and bounding box information. Structurally, each metadata entry consists of: (1) \texttt{role} indicating UI type (e.g., \texttt{AXButton}), (2) \texttt{text} and \texttt{description} capturing semantic content, (3) a \texttt{bbox} dictionary with \texttt{x}, \texttt{y}, \texttt{width}, and \texttt{height}, and (4) optionally a list of \texttt{children} for nested components. This pipeline enables robust and interpretable extraction of macOS GUI structures, supporting downstream tasks such as screen annotation, interaction modeling, and agent behavior learning. Additionally, due to the limited accessibility information exposed by some system-level macOS applications or the difficulty in filtering non-visible elements, we incorporate \texttt{omniparser-v2} as a complementary mechanism to refine and validate extracted elements based on screenshot alignment and bounding box overlap.

\textbf{Mobile Platform.} For Android, we begin by using UIAutomator2\footnote{https://uiautomator2.readthedocs.io/en/latest/} to dump the current app’s accessibility hierarchy as XML and parse it into an in-memory \texttt{lxml} tree. In a depth-first walk, we record each node’s \texttt{class}, \texttt{resource\-ID}, \texttt{text} and \texttt{content description}, and parse its \texttt{bounds} string (e.g.\ \texttt{"[x1,y1][x2,y2]"}) into integer coordinates to build robust locators and raw geometry. During this pass we filter out any control that is off-screen, too small (for example, width~$<5$\,px or height~$<15$\,px), devoid of both text/description and interaction flags (\texttt{clickable}, \texttt{focusable}, \texttt{scrollable}, or \texttt{long-clickable}), or fully occluded by its parent—leaving only truly visible, actionable elements. For each remaining node, we generate a concise label by combining up to the first ten words of its text or description with its UI role (e.g.\ “Button” or “EditText”) and infer possible actions (click, swipe, long press, write). In a second sweep, we detect exactly which elements support taps, focus moves, scrolling, or long presses, then wrap each into a structured record containing its unique identifier, bounding‐box coordinates, a summary of core attributes (ID, text, type, state flags like \texttt{enabled} and \texttt{visible\_to\_user}), and the full raw attribute map (\texttt{package}, \texttt{index}, \texttt{checkable}/\texttt{checked}, \texttt{password}, etc.). Finally, we serialize this collection as a flat JSON array or tab-separated lines, producing a complete, coordinate-aware metadata set that underpins precise mobile UI analysis and automated testing. 
For iOS, we feed the screenshot directly into OmniParser V2~\citep{yu2025omniparser}, which parses the page elements—extracting their type, bounding box, interactivity, content, and so on—and uses this information as metadata.

\textbf{Web Platform.} 
Our web metadata extraction pipeline employs Selenium WebDriver\footnote{https://www.selenium.dev/} with ChromeDriver\footnote{https://www.google.cn/chrome} to automate web interaction trajectory acquisition using a random walk algorithm. At each step, it leverages browser-native rendering to ensure visual fidelity while capturing the current page's element metadata, including coordinates, descriptions, types, and special attribute information.
The pipeline executes a JavaScript parsing pipeline via Chrome DevTools Protocol (CDP) that implements a comprehensive element classification and filter methodology. Clickable elements are identified through a multi-criteria approach combining semantic \textbf{HTML tags} (\texttt{<a>}, \texttt{<button>}, \texttt{<input>}, \texttt{<select>}, \texttt{<textarea>}, \texttt{<option>}, \texttt{<video>}), \textbf{CSS properties} (\texttt{cursor:pointer}, since CSS properties cascade to child elements, we only treat an element as clickable if its parent lacks cursor:pointer, ensuring accurate detection of standalone clickable elements), \textbf{JavaScript click event listeners}, and \textbf{element attributes} (\texttt{onclick}, \texttt{ondblclick}, \texttt{roles} contain button, option, tab); Non-interactive elements are systematically classified as text objects, media objects, or structural panels through DOM hierarchy analysis. All elements undergo rigorous validation including \textbf{geometric verification} using \texttt{getBoundingClientRect()} to filter occluded components, \textbf{visibility validation} through CSS property checks (\texttt{display:none}, \texttt{visibility:hidden}, \texttt{opacity:0}), and \textbf{active validation} via \texttt{document.elementFromPoint()} center-point sampling to confirm visual prominence and top-layer activity. Finally, we perform a set difference operation with the elements from the last step to filter out the set of new elements for the random walk.
Text description metadata aggregation incorporates content from over 12 attributes including \texttt{textContent}, \texttt{innerText}, \texttt{value}, \texttt{alt}, \texttt{title}, and \texttt{aria-label}, normalized through whitespace compression algorithms. The framework implements multiple integrity safeguards including dynamic language detection via langdetect\footnote{https://github.com/Mimino666/langdetect}, sensitive lexicon pattern matching, and visual anomaly detection with adaptive boundary refinement. Cross-resolution robustness is achieved through randomized viewport initialization spanning device pixel ratios (1.4--2.1) and common resolutions~(720p, 1080p, 2K, 4K, 2560$\times$1600), stabilized via CSS viewport normalization techniques. This comprehensive web trajectory metadata extraction pipeline ensures exceptional data integrity, security, diversity, granularity, and accuracy, thereby establishing a robust foundation for instruction construction and model training.

\subsubsection{GUI Grounding}
\label{sec:gui_grounding}
GUI grounding is a fundamental capability for computer use agents, enabling them to associate the natural language instruction with a corresponding region of interest. Effective grounding determines whether the agent can interact with the correct interface components, directly impacting its ability to complete downstream tasks. In fact, a grounding-only agent can be paired with a general-purpose planner (e.g., GPT-4o~\citep{hurst2024gpt}) to complete tasks via a modular style.

To support various grounding demands, we construct a multi-format GUI grounding corpus with three distinct supervision targets: point grounding, bounding box grounding, and action grounding. Point grounding requires the model to identify a single pixel-level location, typically the center of a button, icon, or control, that corresponds to a user instruction. Bounding box grounding extends this capability by predicting rectangular regions that encapsulate target elements, which is particularly useful for operations involving region selection, such as dragging or editing. Action grounding combines spatial localization with operational semantics by producing an executable command, such as \texttt{click(x=105, y=23)}, that aligns with the intended interaction.
As for the annotation, we reuse structured annotations generated during the GUI understanding stage. Specifically, appearance, spatial, and functional descriptions of each UI element provide rich supervision signals. The center point and bounding box coordinates are extracted directly from UI layout metadata or visual parsing modules. Action-level grounding pairs these spatial targets with predefined atomic operations based on the element’s inferred function. 
In addition, we explore data augmentation strategies to expand the grounding corpus. Specifically, we filter out previously annotated elements from the metadata and use prompt templates combined with GPT-4o to generate a larger set of grounding annotations. This augmented data is designed to improve the model’s generalization ability across diverse GUI layouts and interaction patterns. This annotated corpus serves as a foundation for learning robust visual-linguistic alignment and facilitates both direct interaction and integration with high-level task planners.

\subsubsection{Weak-Semantic Trajectory}
\label{sec:weak_traj}
While the trajectories collected by rule-based agents do not correspond to explicit task objectives, we incorporate heuristics into the exploration process to encourage transitions into deeper and less frequently visited interface states. This results in more diverse and representative interaction sequences, which are critical for training agents to generalize across complex GUI structures.

To further exploit the potential of these unsupervised trajectories, we segment long interaction sequences into shorter, weakly semantic sub-trajectories. The segmentation is based on screenshot similarity: when a current screen is visually similar to a previous one, it often indicates that the agent has reached a terminal or redundant interface state with minimal novelty in further interactions. These similarity-based boundaries serve as natural points for restarting exploration, thereby improving coverage and trajectory diversity.

We refer to the resulting sequences as \textit{weak-semantic trajectories}, as they preserve partial continuity and structural coherence without being aligned to manually defined tasks. Despite their lack of strong supervision, such trajectories often reflect meaningful UI flows, especially when the agent is biased toward newly rendered elements.

We hypothesize that exposure to weak-semantic trajectories can help the agent internalize common patterns of GUI interaction and enhance its planning ability. If validated, this approach may offer a cost-effective alternative to large-scale manual annotation, accelerating the evolution of more capable computer use agents through low-cost, high-coverage exploration.

\subsubsection{Human-Curated Trajectory}
In addition to rule-driven exploration, we incorporate human-curated trajectories to address the limitations of automatically collected data. While rule-driven agents enable scalable collection, they inherently exhibit stochasticity and often fail to uncover certain goal-directed operations, especially for tasks requiring deep or context-specific interactions. Moreover, although weak-semantic trajectories segmented from raw explorations provide partial structure, their action sequences are not always aligned with human reasoning. As a result, they may contain fragmented or noisy behaviors that limit their utility for downstream training.

To overcome these limitations, we design a human-in-the-loop protocol for collecting high-quality task trajectories. We begin by constructing a seed task set, categorizing applications into common use domains such as daily utilities, entertainment, and productivity. For each domain, we identify representative applications and select frequently used functions based on user documentation and empirical analysis. Annotators are then instructed to convert these functions into clear, goal-oriented task descriptions, ensuring linguistic clarity and operational feasibility.
Using our unified cross-platform recording system, human experts remotely interact with each application environment encapsulated within a Docker container. This design provides process isolation, avoids side effects such as misoperation. Annotators are able to finish tasks in a natural and fluent manner, producing coherent action trajectories that reflect realistic usage patterns across platforms.

These curated trajectories serve as high-quality supervision for training agents with accurate planning and execution capabilities. They complement the broader, noisier dataset collected via automation, and provide reference paths that guide model alignment with human intent and behavior.

\subsubsection{Annotation Schemes}
\label{sec:anno_scheme}
In our data acquisition, we collect screenshots along with their metadata, which includes all potentially interactive elements on the page. 
Since different exploration paths can lead to the same state and common states like the homepage are visited frequently, we employed image feature similarity to deduplicate these screenshots. 
This yields a unique set of interface screenshots paired with their corresponding metadata.
To reduce redundancy and mitigate noise within the metadata, we randomly sample 25 to 40 elements per screenshot. 
These elements are then semantically filtered using GPT-4o to ensure both efficacy and diversity. 

For each retained element, we mark its position on the image using a red box with an arrow. By combining with associated metadata, we prompt GPT-4o to generate appearance and position descriptions, and Claude-3.7-Sonnet~\citep{anthropic2025claude37} to generate functional descriptions. 
These serve as ground truth annotations for our Element Appearance Captioning, Element Layout Understanding, and Element Functionality Captioning, respectively.
These descriptions are further used to construct grounding tasks, where the appearance and position descriptions are used for non-action grounding and the function description is used for action-based grounding.
To simulate all possible positions of elements and accommodate a wider range of usage scenarios, we perform data augmentation. This includes simulating higher resolutions by stitching two images together, as well as cropping elements and pasting them onto solid-color backgrounds or real-world backgrounds from images captured by the author's own device.

For each unique interface screenshot, GPT-4o is also used to generate an overall caption. If the image was not the final step of a trajectory, we additionally provided GPT-4o with the subsequent screenshot along the same exploration path to summarize the UI changes and infer the intention. 
These are used for Screen Transition Captioning and User Intention Prediction tasks.

For all trajectories, we provide Claude-3.7-Sonnet with the current and next screenshots, as well as a cropped image of the interacted element, to infer both the step-level instruction and the reasoning process.
For weakly semantic trajectories that primarily involve navigation across pages, we generate high-level task objectives. To do this, we provide Claude-3.7-Sonnet with the first and last screenshots of the trajectory to synthesize a navigation-related task goal.
Considering that different annotators have varying styles of writing instructions and different operational habits, we implement two types of augmentations for trajectories to improve model generalization. The first is instruction augmentation, where we prompt the model to generate task instructions in diverse styles, aiming to cover all possible user scenarios. The second is trajectory augmentation, for which we prompt the model to generate several step-level instructions and the reasoning process based on the trajectory. This can help mitigate the noise introduced by model labeling. All prompts used for annotation are provided in the Appendix~\ref{sec:prompt_template_anno}.

\subsubsection{More Details of Data Distribution}
To support the development of general-purpose GUI agents, we curate a large-scale training corpus spanning three major task domains: \textit{Understanding}, \textit{Grounding}, and \textit{Task Planning}. As shown in Table~\ref{tab:data_distribution}, our dataset comprises 471.4K examples for GUI Understanding, focusing on capabilities such as OCR, UI Comprehension, and Layout Comprehension, Screen Transition Reasoning sourced from 355.5K unique screenshots. For GUI Grounding, we annotate over 17.1M examples across multiple formats including point, bounding box, and action—grounded on 1.6M images. These annotations enable precise visual-textual alignment across diverse platforms and screen configurations.

\begin{figure*}[!htb]
  \centering
  \includegraphics[width=0.6\linewidth]{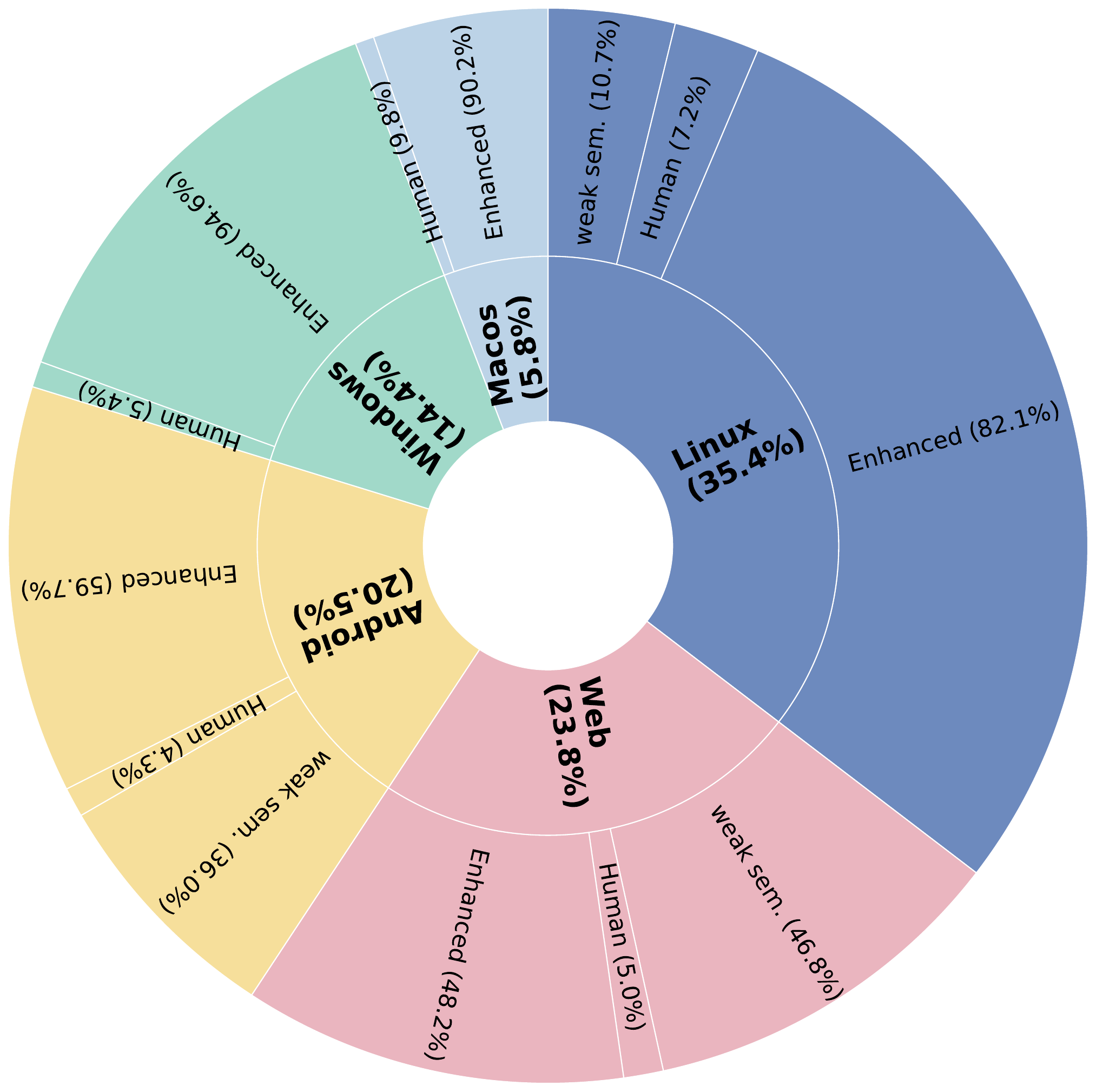}
    \vspace{-0.25em}
\caption{A hierarchical visualization of the trajectory composition.}
\label{fig:trajectory_composition}
  \vspace{-0.5em}
\end{figure*}

For Task Planning, we provide 19K unique trajectories categorized into three types: Weak Semantics, Human-Curated, and Enhanced. The Human-Curated Trajectories consist of 4K high-quality goal-oriented demonstrations collected by experts, while the Enhanced Trajectories incorporate diverse instructions and augmented reasoning steps, totaling 48.2K examples. Weak Semantics Trajectories derived from rule-based exploration with limited semantic guidance are used for supplementing navigation knowledge. Fig.~\ref{fig:trajectory_composition} provides a hierarchical view of the trajectory composition across platforms and types. By integrating agent-generated and expert-curated signals, we ensure both data diversity and quality.

In conclusion, our \ourdataset delivers the largest GUI grounding dataset to date, coupled with substantial understanding and planning examples. Its platform coverage and hierarchical task composition form a comprehensive foundation for training robust, cross-platform GUI agents. The performance of \modelname validates the quality of \ourdataset, and highlights future directions in data-centric agent training.
\begin{table}[thbp]
\centering
\small
\caption{Distribution of examples in our training corpus.}
\label{tab:data_distribution}
\scalebox{0.9}{
\begin{tabular}{llcc}
\toprule
\textbf{Task Domain} & \textbf{Tasks} & \textbf{\#Images} & \textbf{\#Examples} \\
\midrule
\multirow{6}{*}{Understanding} & Element Appearance Captioning, & \multirow{6}{*}{355.5K} & \multirow{6}{*}{471.4K} \\
 & Referring OCR, &  &  \\
 & Element Layout Understand, &  &  \\
 & Element Functionality Captioning, &  &  \\
 & User Intention Prediction, &  &  \\
 & Interface Captioning, &  &  \\
 & Screen Transition Captioning &  &  \\
\midrule
\multirow{2}{*}{Grounding} & Bounding Box & \multirow{2}{*}{1.6M} & \multirow{2}{*}{17.1M} \\
 & Point, Action &  &  \\
\midrule
\multirow{3}{*}{Task Planning} & Weak Semantics Trajectories & 5.5K & 15.0K \\
 & Human-Curated Trajectories & 29.3K & 4.0K \\
 & Enhanced Trajectories & 29.3K & 48.2K \\
\bottomrule
\end{tabular}
        }
\end{table}

\subsection{Data Visualization}
\subsubsection{GUI Understanding}
To qualitatively demonstrate our data in GUI understanding tasks, we provide examples that cover both element-level and screenshot-level understanding. 
At the element level, we have designed five distinct tasks regarding individual GUI elements. Table~\ref{tab:vis_understanding_elements} showcases specific examples of these tasks.
At the screenshot level, we focus on the ability to comprehend the entire GUI interface globally and its dynamic changes. Table~\ref{tab:vis_understanding_screenshot} provides examples for these two tasks.

\begin{table}[h!]
\centering
\caption{Task visualization for element-level understanding.}
\label{tab:vis_understanding_elements}
\begin{tabular}{|c|m{7.5cm}|} 
\mytoprule

\multicolumn{2}{|l|}{\textbf{Examples of Five Types of Tasks.}} \\
\mymidrule

\raisebox{-0.5\height}{\includegraphics[width=0.4\textwidth]{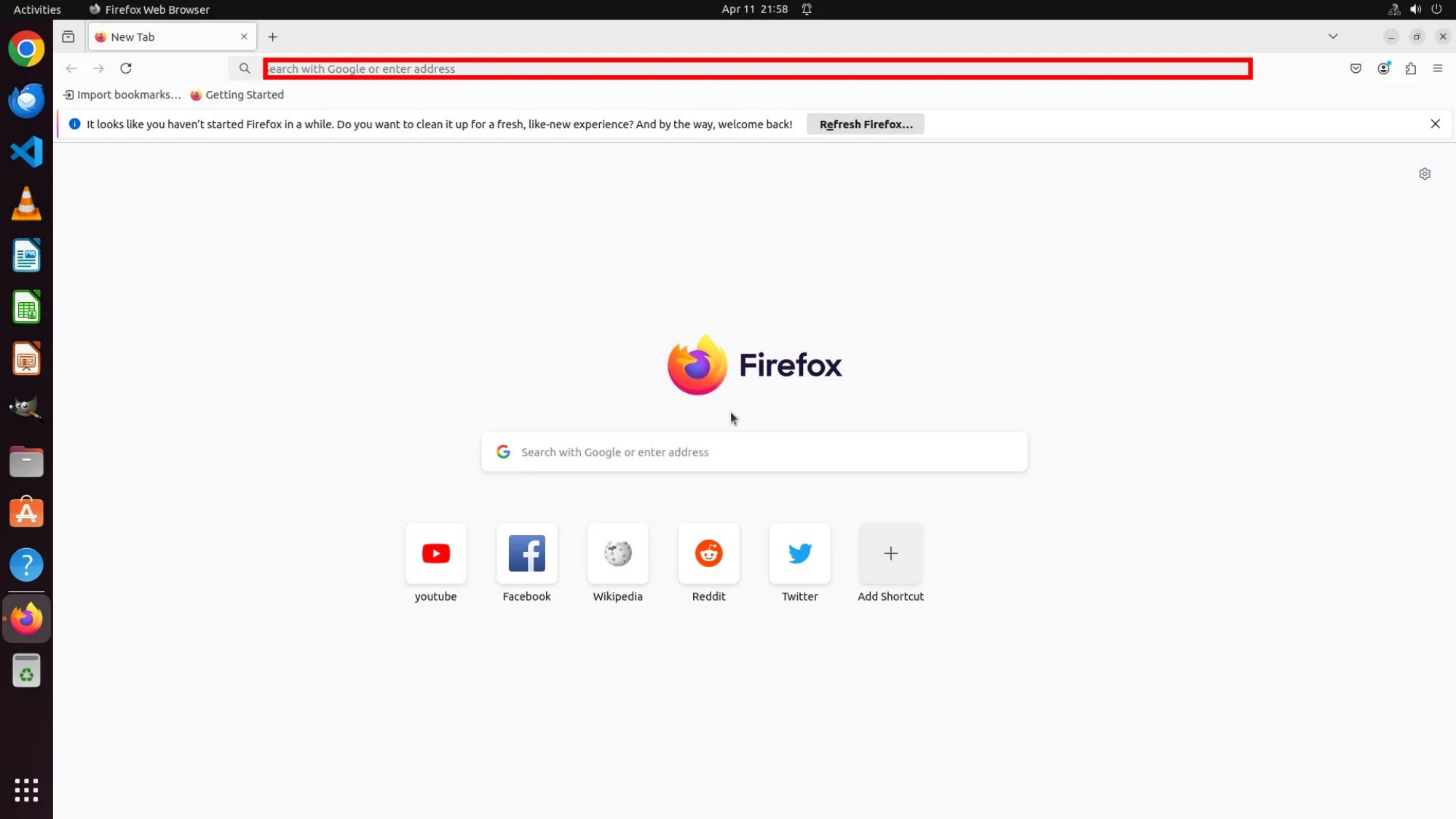}} &
    \underline{[Element Appearance Captioning]}
    
    \textbf{Q:} Please describe the appearance of the element marked in the image.

    \textbf{A:} A rectangular search bar with a white/light gray background and placeholder text 'Search with Google or enter address' in light gray font. The search bar has smooth, rounded corners and a subtle border. It appears to be empty with no entered text.
\\
\mymidrule

\raisebox{-0.5\height}{\includegraphics[width=0.4\textwidth]{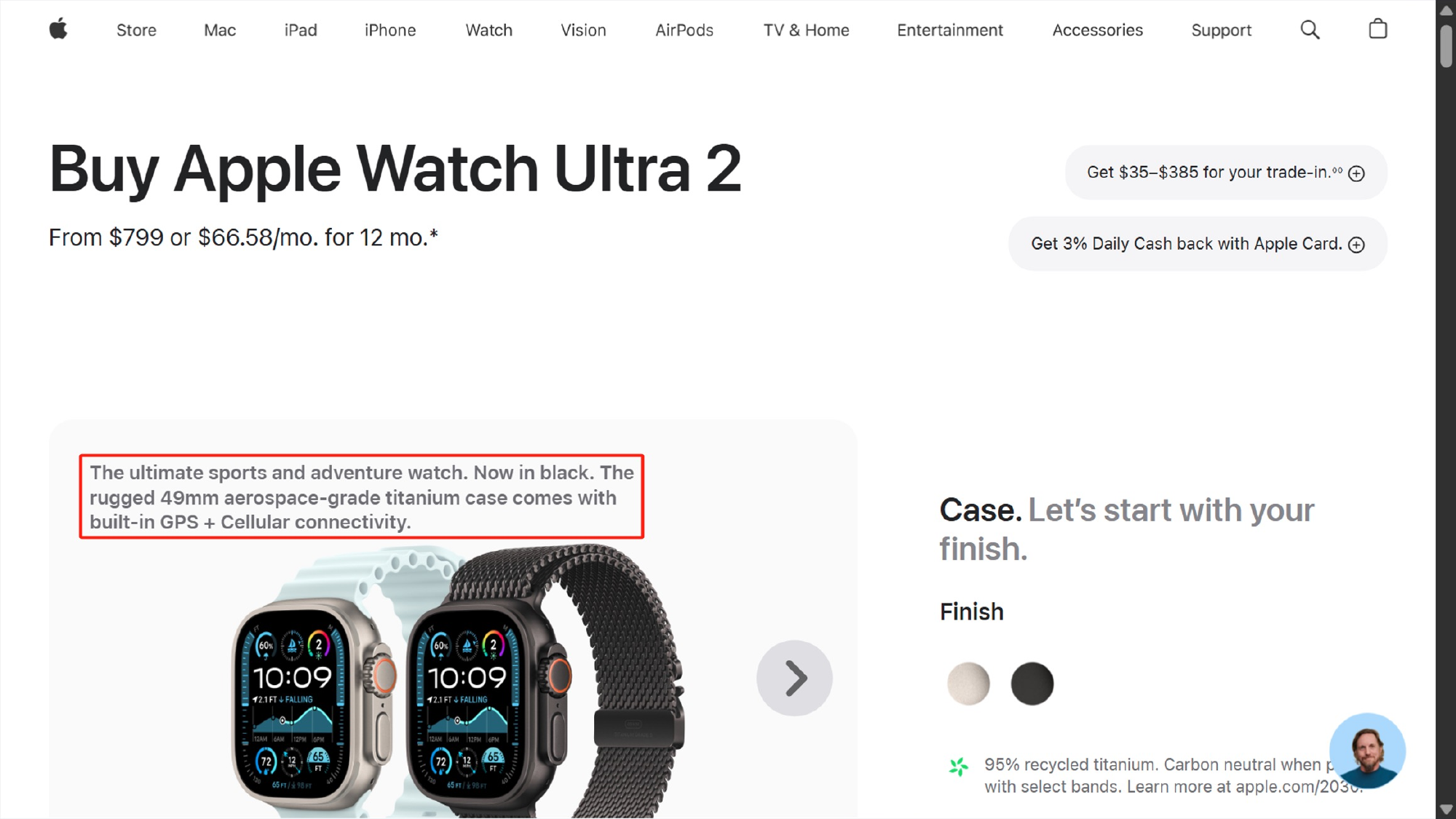}} &
    \underline{[Referring OCR]}
    
    \textbf{Q:} Please output the text content correctly responding to this term: \textless box\textgreater[[61, 563, 435, 651]]\textless/box\textgreater

    \textbf{A:} A rectangular search bar with a white/light gray background and placeholder text 'Search with Google or enter address' in light gray font. The search bar has smooth, rounded corners and a subtle border. It appears to be empty with no entered text.
\\
\mymidrule

\raisebox{-0.5\height}{\includegraphics[width=0.4\textwidth]{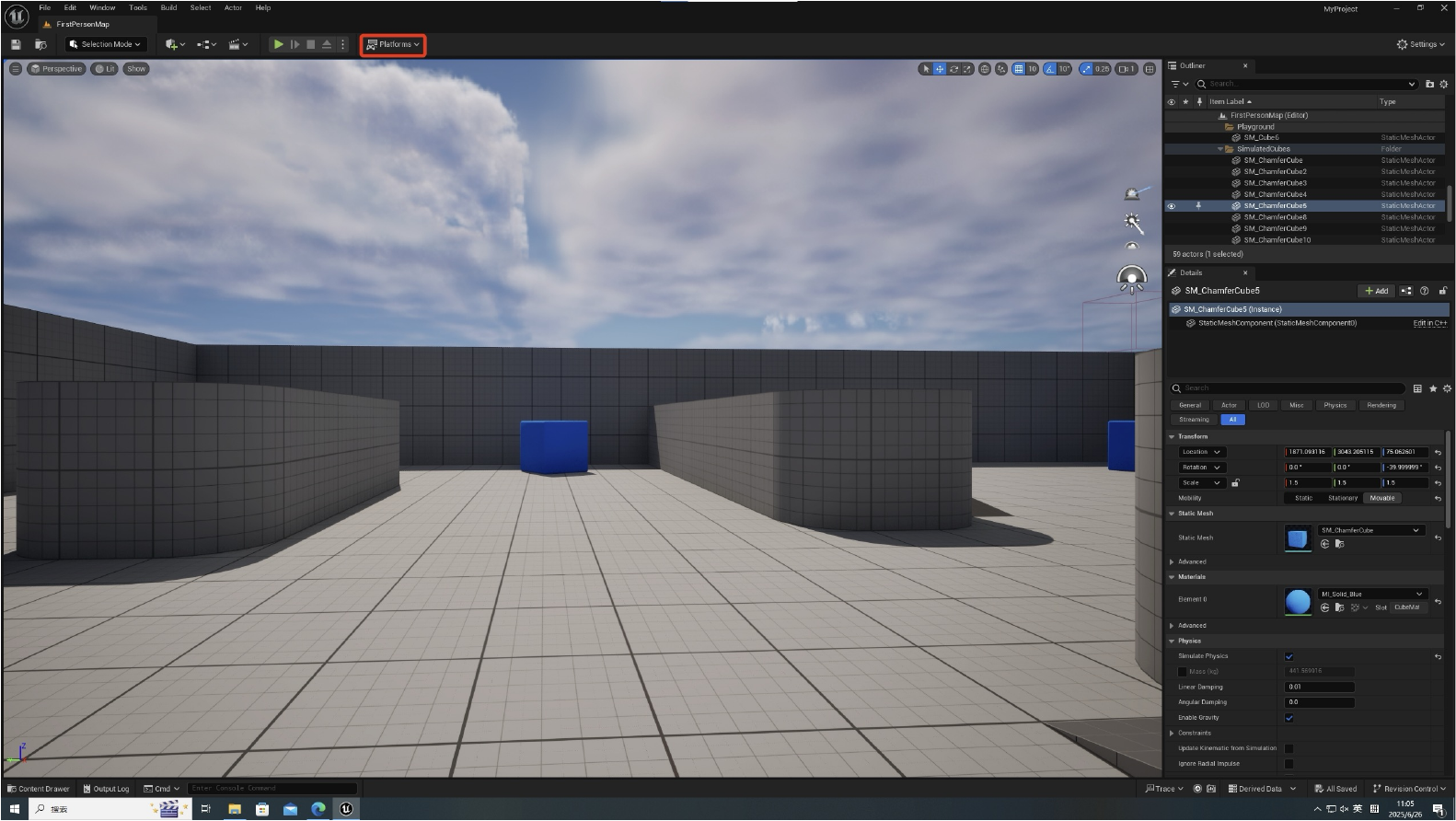}} &
\underline{[Element
Layout Understanding]}

\textbf{Q:}
Describe the position of the highlighted elements in the image, such as their location relative to other objects, alignment, and any spatial relationships.

\textbf{A:} Located in the top toolbar of the application window, positioned between the interface navigation controls on the left and the viewport controls on the right. It sits approximately in the upper third of the screen, within the main toolbar area.
\\
\mymidrule

\raisebox{-0.5\height}{\includegraphics[width=0.4\textwidth]{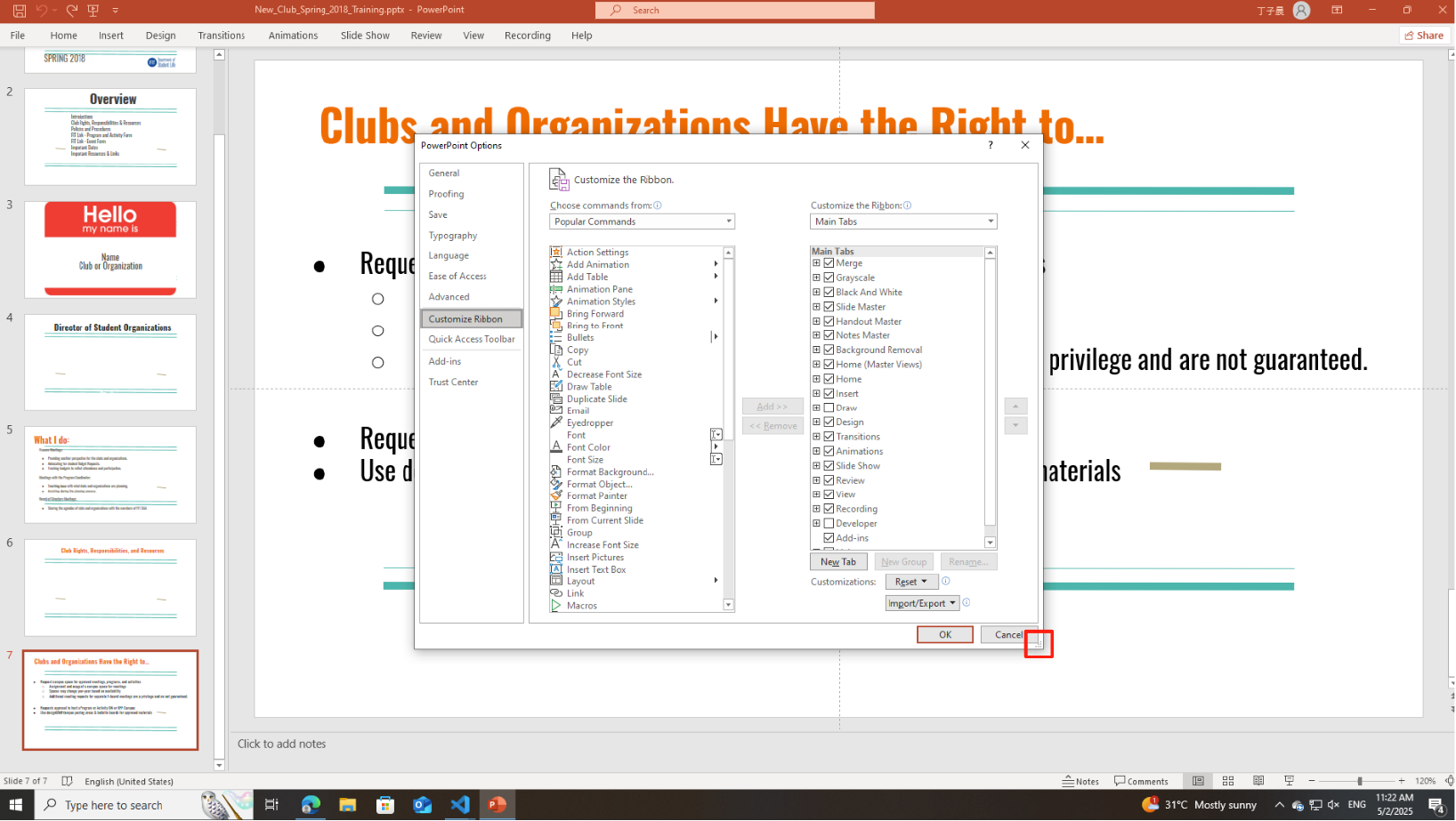}} &
\underline{[Element Functionality Captioning]}

\textbf{Q:}
Describe the position of the highlighted elements in the image, such as their location relative to other objects, alignment, and any spatial relationships.

\textbf{A:} Use the size grip at the corner of the PowerPoint Options window to change its size
\\
\mymidrule

\raisebox{-0.5\height}{\includegraphics[width=0.2\textwidth]{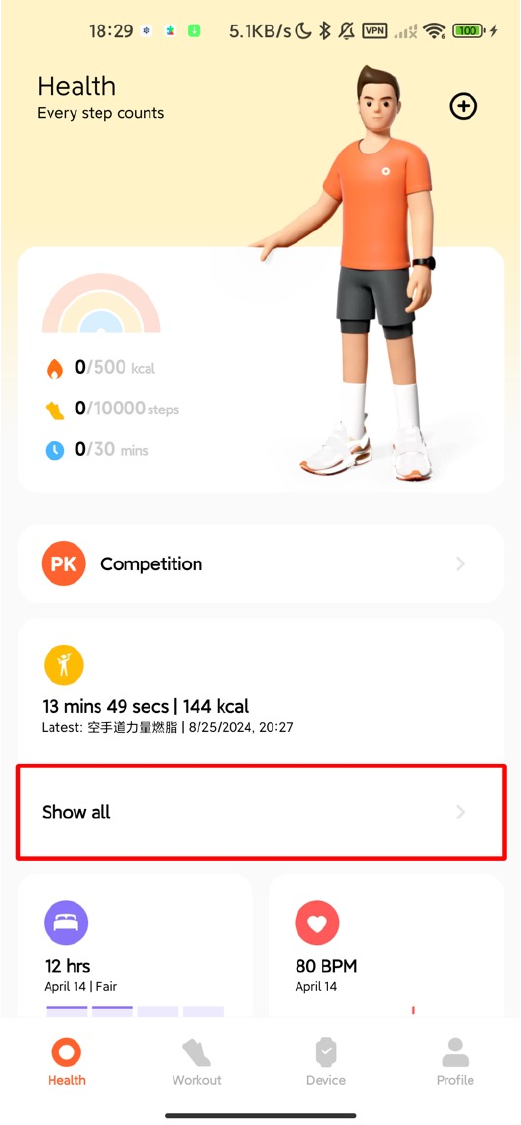}} &
\underline{[User Intention Prediction]}

\textbf{Q:}
Analyze the current image and the provided action 'click(x=540, y=1686)', then predict the user's intent based on these inputs

\textbf{A:} The user likely intends to view detailed information or a complete list of their past workout activities, specifically the "Latest: Karate Power Fat Burn" session recorded on 8/25/2024 at 20:27, as the label "Show all" suggests an expansion or summary of historical data.
% \\
\\
\mybottomrule

\end{tabular}
\end{table}

\begin{table}[ht!]
\centering
\caption{Task visualization for screenshot-level understanding.}
\label{tab:vis_understanding_screenshot}
\begin{tabular}{|c|m{8.5cm}|} 
\mytoprule

\multicolumn{2}{|l|}{\textbf{Examples of Two Types of Tasks.}} \\
\mymidrule

\raisebox{-0.5\height}{\includegraphics[width=0.3\textwidth]{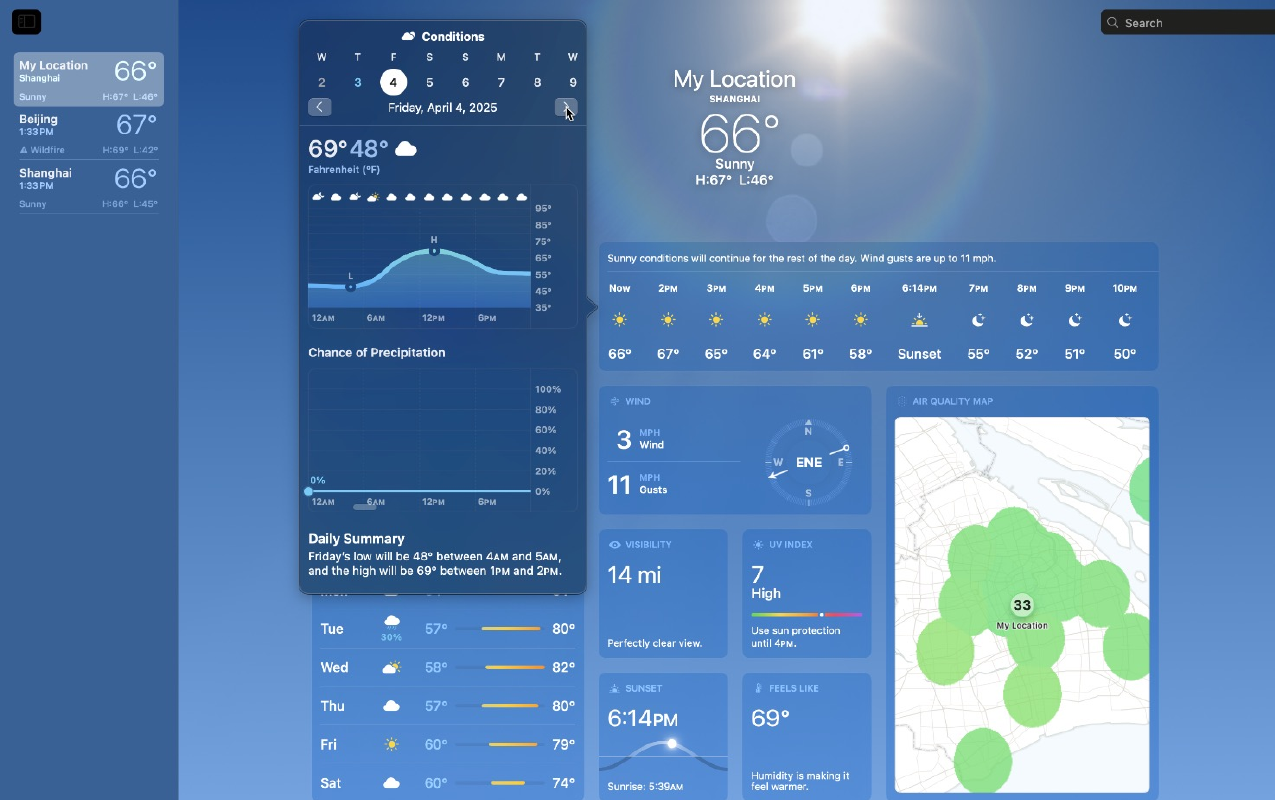}} &
    \underline{[Interface Captioning]}
    
    \textbf{Q:} Provide a detailed description in the current image.

    \textbf{A:}The image shows a weather application interface displaying the forecast for Shanghai. The current temperature is 66°F (about 19°C) with sunny conditions. The interface shows a detailed forecast for Friday, April 4, 2025, with a high of 69°F and a low of 48°F. The hourly forecast indicates temperatures will range from 66°F to 50°F throughout the day, with sunset occurring at 6:14 PM. Wind conditions show a speed of 3 MPH with gusts up to 11 MPH. The interface also displays additional weather information including visibility (14 mi), UV index (7 - High), and a five-day forecast showing temperatures for upcoming days. The air quality map shows green areas indicating good air quality around the user's location.
\\
\mymidrule

\raisebox{-0.5\height}{\includegraphics[width=0.3\textwidth]{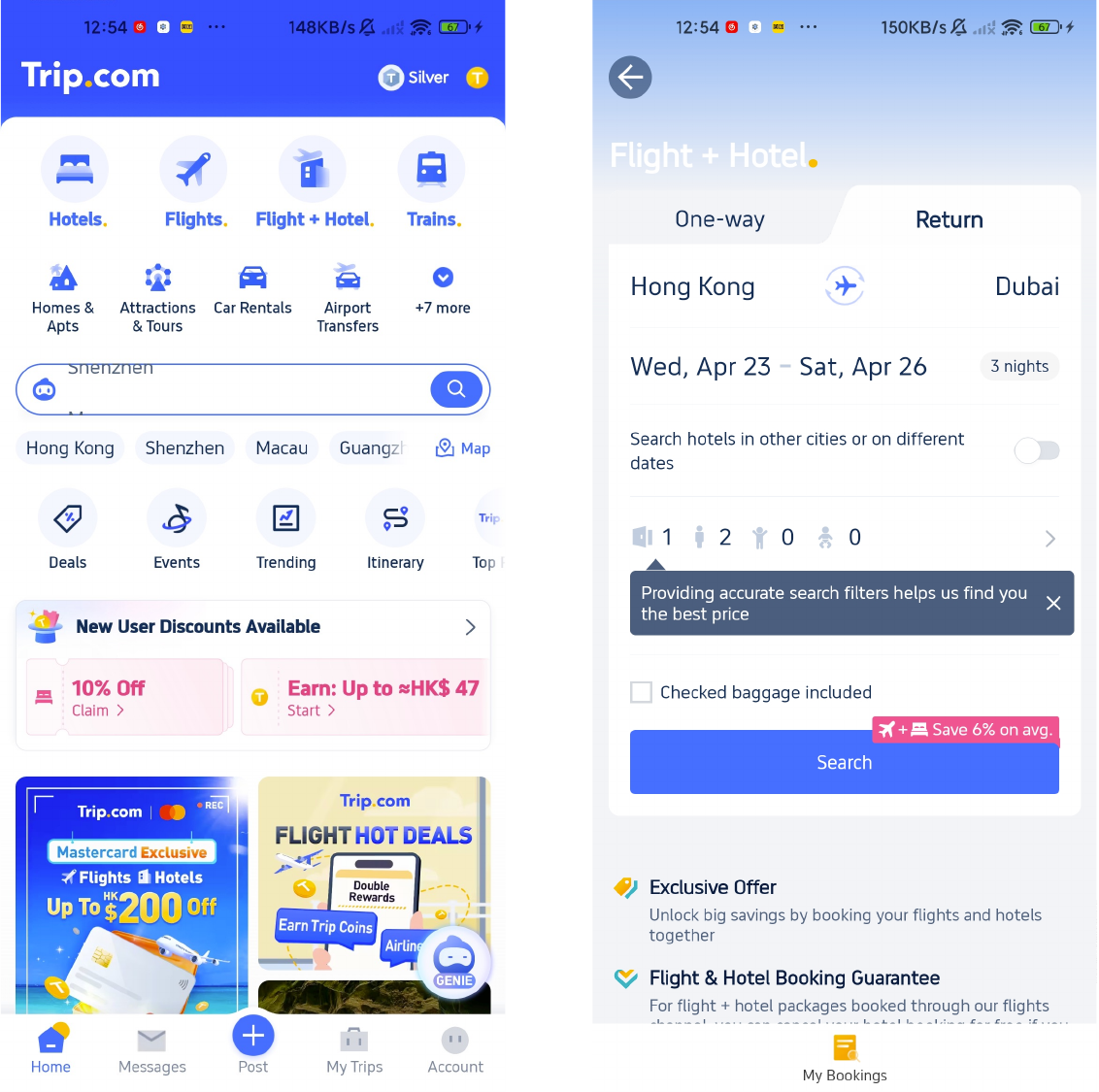}} &
    \underline{[Screen Transition Captioning]}
    
    \textbf{Q:} Analyze the differences between two consecutive GUI screenshots. Describe the initial state, highlight the changes in the second screenshot.

    \textbf{A:}In the first screenshot, we see the Trip.com mobile app homepage with various booking options including Hotels, Flights, Flight + Hotel, Trains, and other services. After tapping on Flight + Hotel option, the second screenshot shows the Flight + Hotel booking page where users can input their travel details. The page displays a one-way/return toggle (with 'Return' selected), origin and destination fields set to Hong Kong and Dubai respectively, travel dates (Wed, Apr 23 - Sat, Apr 26 for 3 nights), traveler count settings, and a search button at the bottom.
\\
\mymidrule

\end{tabular}
\end{table}

\begin{table}[!h]
\centering
\caption{Task visualization for GUI grounding}
\label{tab:vis_grounding}
\begin{tabular}{|c|m{8.2cm}|} 
\mytoprule

\multicolumn{2}{|l|}{\textbf{Examples of Three Modes.}} \\
% \midrule
\mymidrule

\raisebox{-0.5\height}{\includegraphics[width=0.35\textwidth]{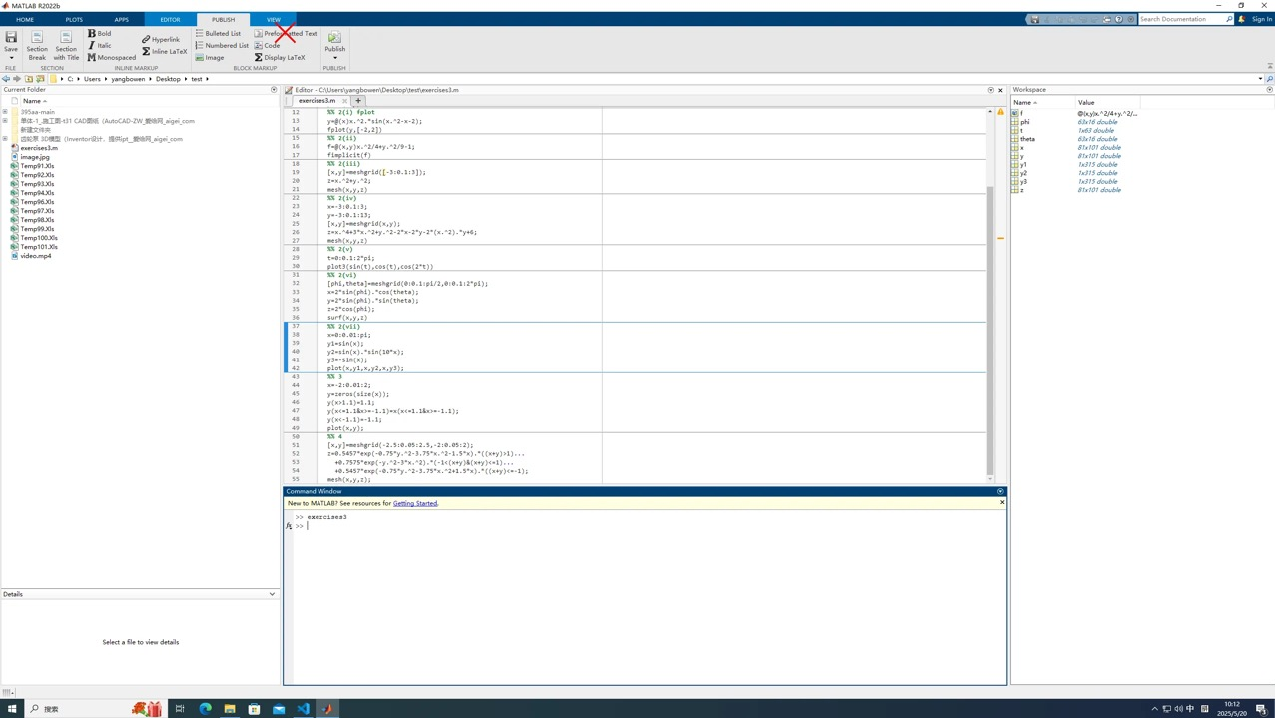}} & 
\underline{[Point Grounding]}

\textbf{Q:}
Return the point within this UI element: \textless ref\textgreater Preformatted Text button in the EDITOR tab's formatting toolbar that allows users to insert pre-formatted text tags in MATLAB's editor.\textless /ref\textgreater

\textbf{A:}
\textless ref\textgreater Preformatted Text button in the EDITOR tab's formatting toolbar that allows users to insert pre-formatted text tags in MATLAB's editor.\textless /ref\textgreater\textless point\textgreater[[223, 45]]\textless /point\textgreater

\\
\mymidrule
\raisebox{-0.5\height}{\includegraphics[width=0.35\textwidth]{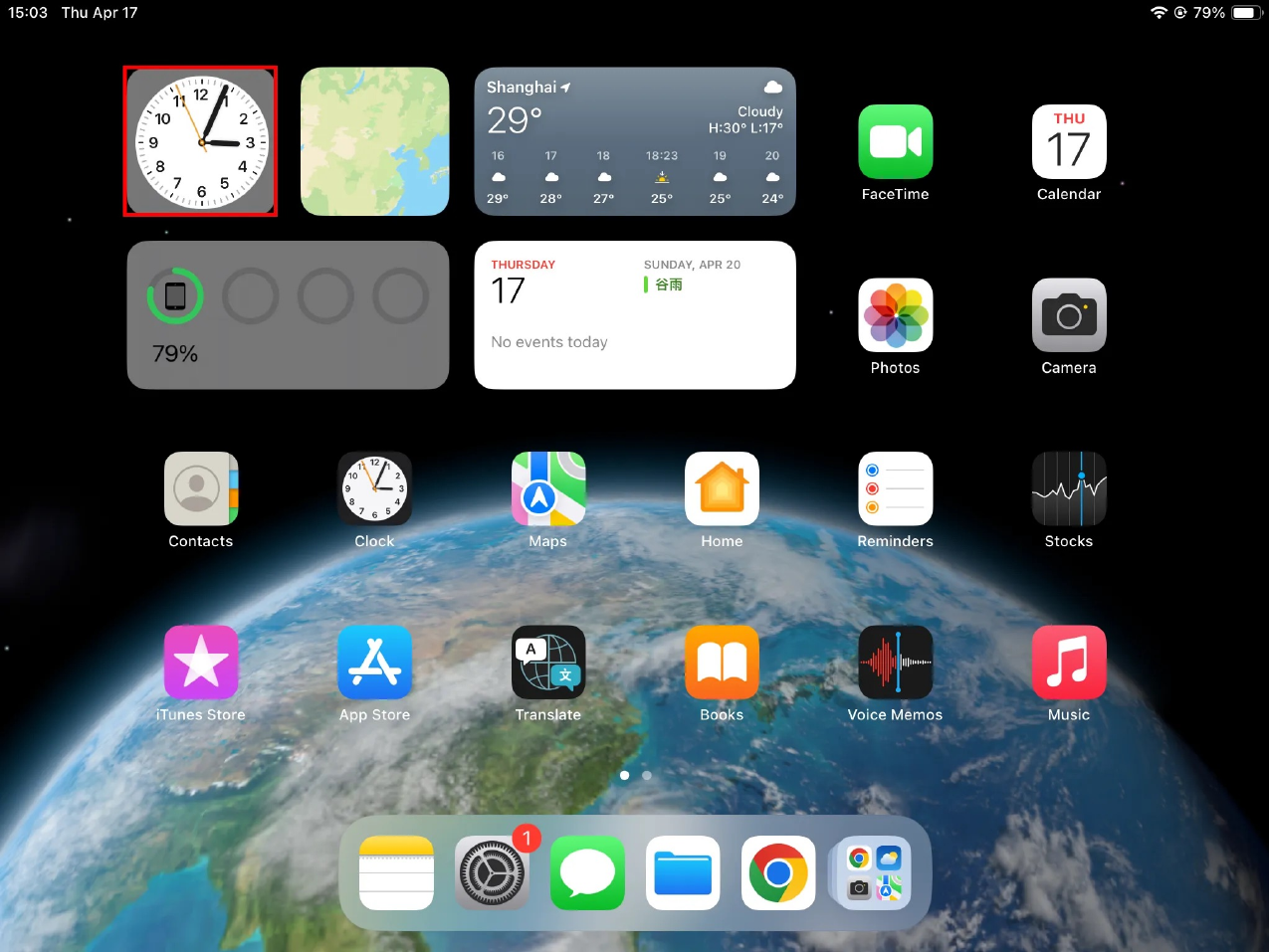}} & 
\underline{[Bbox Grounding]}

\textbf{Q:}
Indicate the location with a bounding box to this UI element: \textless ref\textgreater A white-faced analog clock with black numerals (1-12) and three hands, placed in the upper left corner.\textless ref\textgreater

\textbf{A:}
\textless ref\textgreater A white-faced analog clock with black numerals (1-12) and three hands, placed in the upper left corner.\textless ref\textgreater\textgreater[[97, 69, 218, 227]]\textless bbox\textgreater

\\
\mymidrule

\raisebox{-0.5\height}{\includegraphics[width=0.35\textwidth]{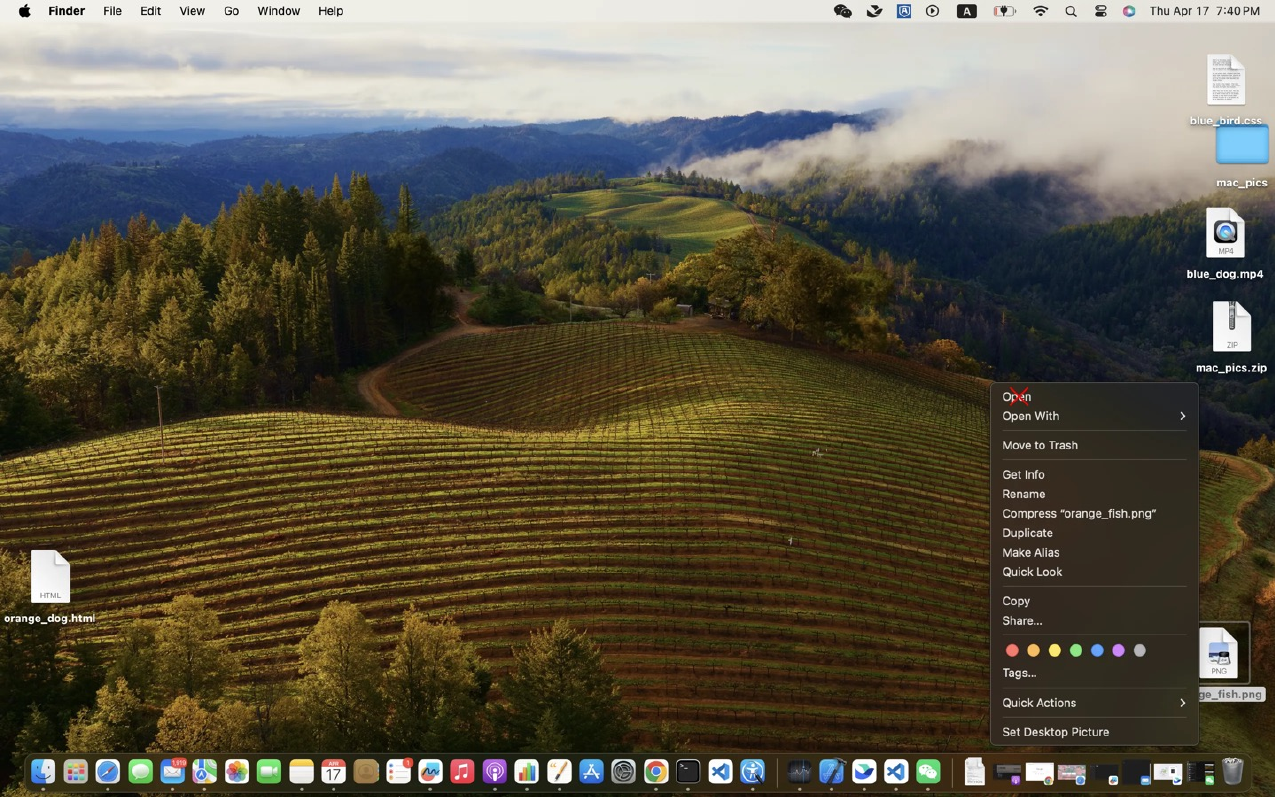}} &
\underline{[Action Grounding]}

\textbf{Q:}
Click the 'Open' option to open the selected file

\textbf{A:} 
\textless action\textgreater
click(x=0.7983, y=0.4967)
\textless/action\textgreater
\\
\mybottomrule

\end{tabular}
\end{table}

\subsubsection{GUI Grounding}
To enable precise interaction, a model must be able to map natural language instructions to specific elements on an interface. 
As mentioned in Section~\ref{sec:agentmode}, we support three distinct GUI grounding modes to accommodate a variety of scenarios and Table~\ref{tab:vis_grounding} provides a visual illustration of these three modes.

\subsubsection{Weak-Semantic Trajectory}
As mentioned in Sec.~\ref{sec:weak_traj}, the weak-semantic trajectory processing pipeline begins with rule-driven agents performing depth-first exploration across GUI environments. These agents generate extensive interaction sequences by randomly selecting actions from available options at each step, while heuristic pruning eliminates redundant branches to improve efficiency. Although these trajectories lack explicit high-level goals, they provide broad coverage of interface states. To enhance the utility of these unsupervised trajectories, we segment them into coherent sub-sequences based on visual similarity between screenshots.

\begin{figure*}[!h]
  \centering
  \includegraphics[width=0.9\linewidth]{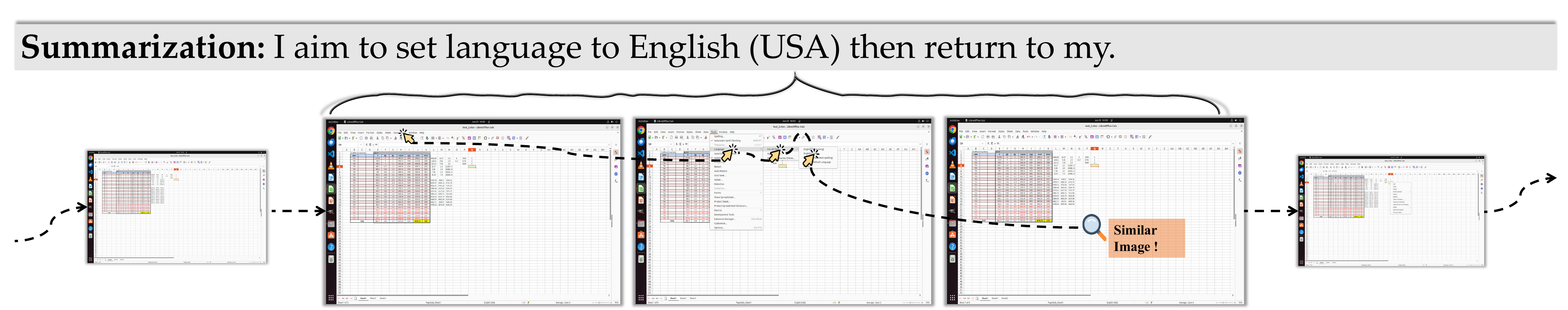}
\caption{An example of a weak semantic trajectory on the Ubuntu platform.}
  \label{fig:ubuntu_weak_trajectory_visualizations}
\end{figure*}

\begin{figure*}[!h]
  \centering
  \includegraphics[width=0.9\linewidth]{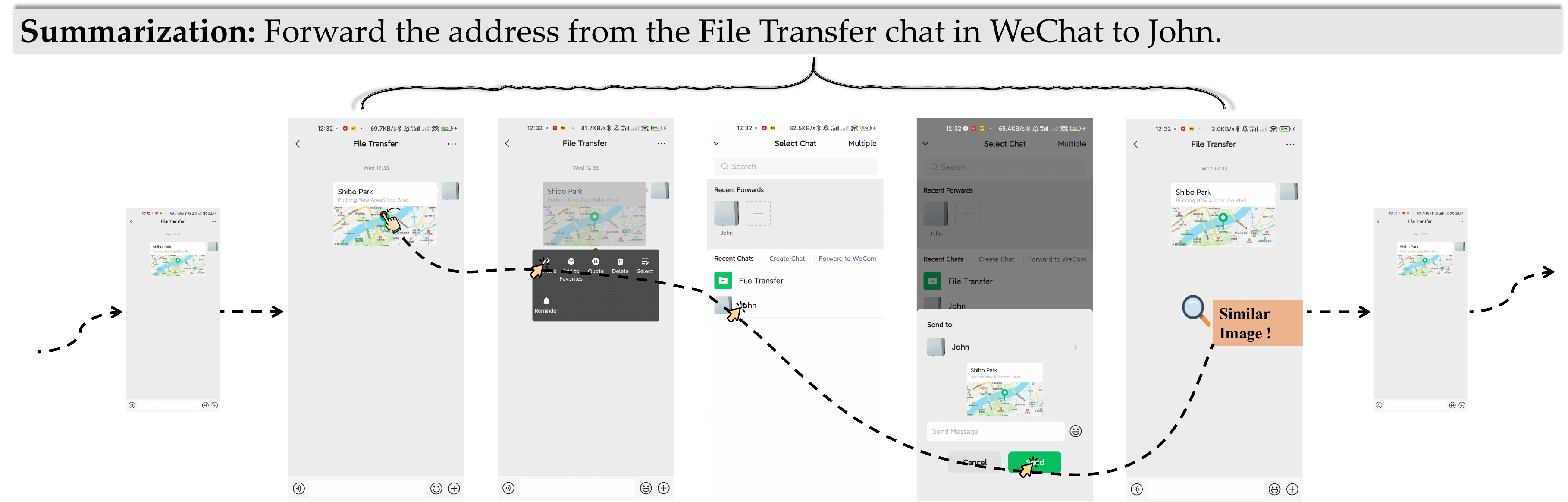}
\caption{An example of a weak semantic trajectory on the Android platform.}
  \label{fig:android_weak_trajectory_visualizations}
\end{figure*}

\begin{figure*}[!h]
  \centering
  \includegraphics[width=0.9\linewidth]{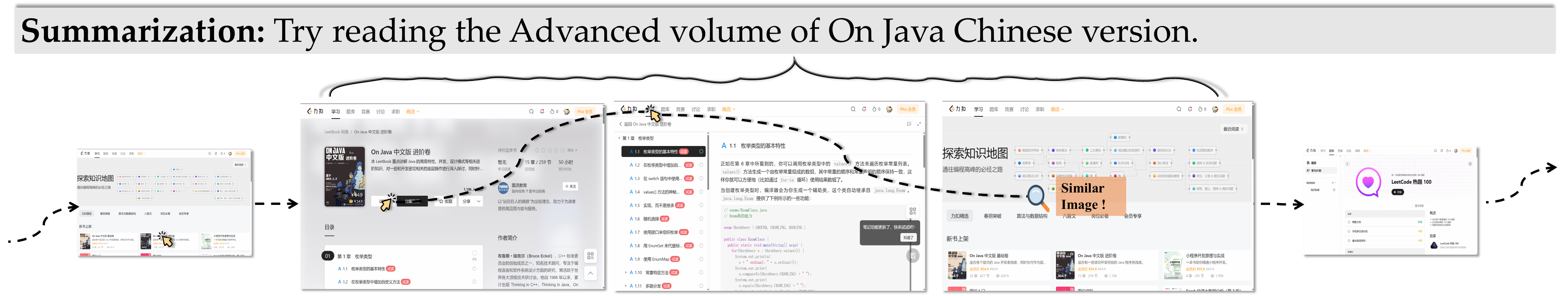}
    \caption{An example of a weak semantic trajectory on the web platform.}
  \label{fig:web_weak_trajectory_visualizations}
\end{figure*}

Fig.~\ref{fig:ubuntu_weak_trajectory_visualizations} demonstrates the exploration pattern in desktop environments, while Fig.~\ref{fig:android_weak_trajectory_visualizations} and~\ref{fig:web_weak_trajectory_visualizations} show the equivalent processes for mobile and web platforms, respectively. In each case, the random-walk agent generates comprehensive navigation patterns through systematic exploration of the interface. These weak-semantic trajectories serve as valuable training signals despite their lack of strong task alignment. They expose models to common GUI interaction patterns and state transitions, complementing the more structured but limited human-curated data.

\subsubsection{Human-Curated Trajectory}
\begin{figure*}[!h]
  \centering
  \includegraphics[width=0.9\linewidth]{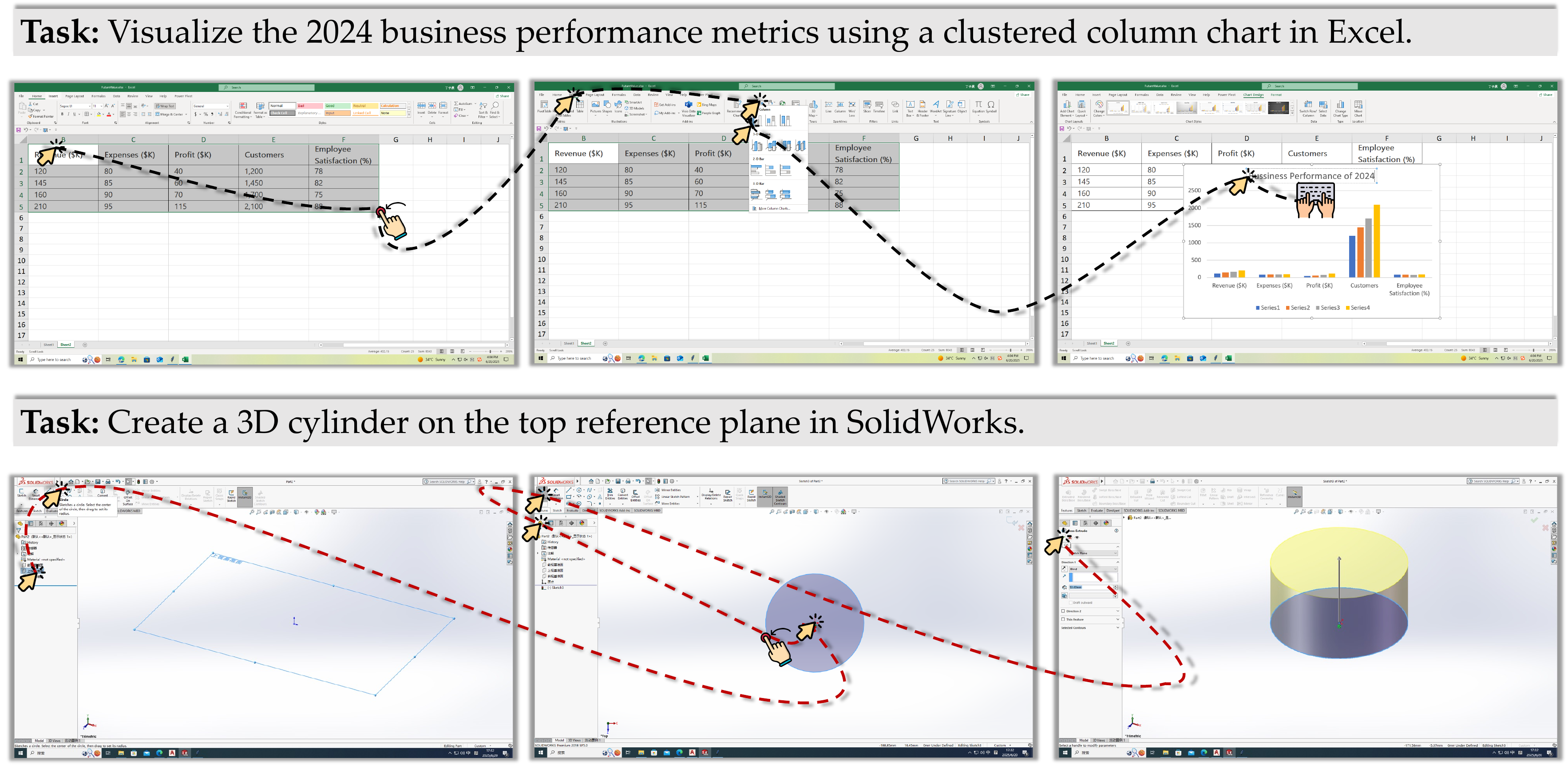}
    \vspace{-0.5em}
\caption{Examples of human-curated trajectories on the Windows platform.}
  \label{fig:windows_trajectory_visualizations}
  \vspace{-0.75em}
\end{figure*}

\begin{figure*}[!h]
  \centering
  \includegraphics[width=0.9\linewidth]{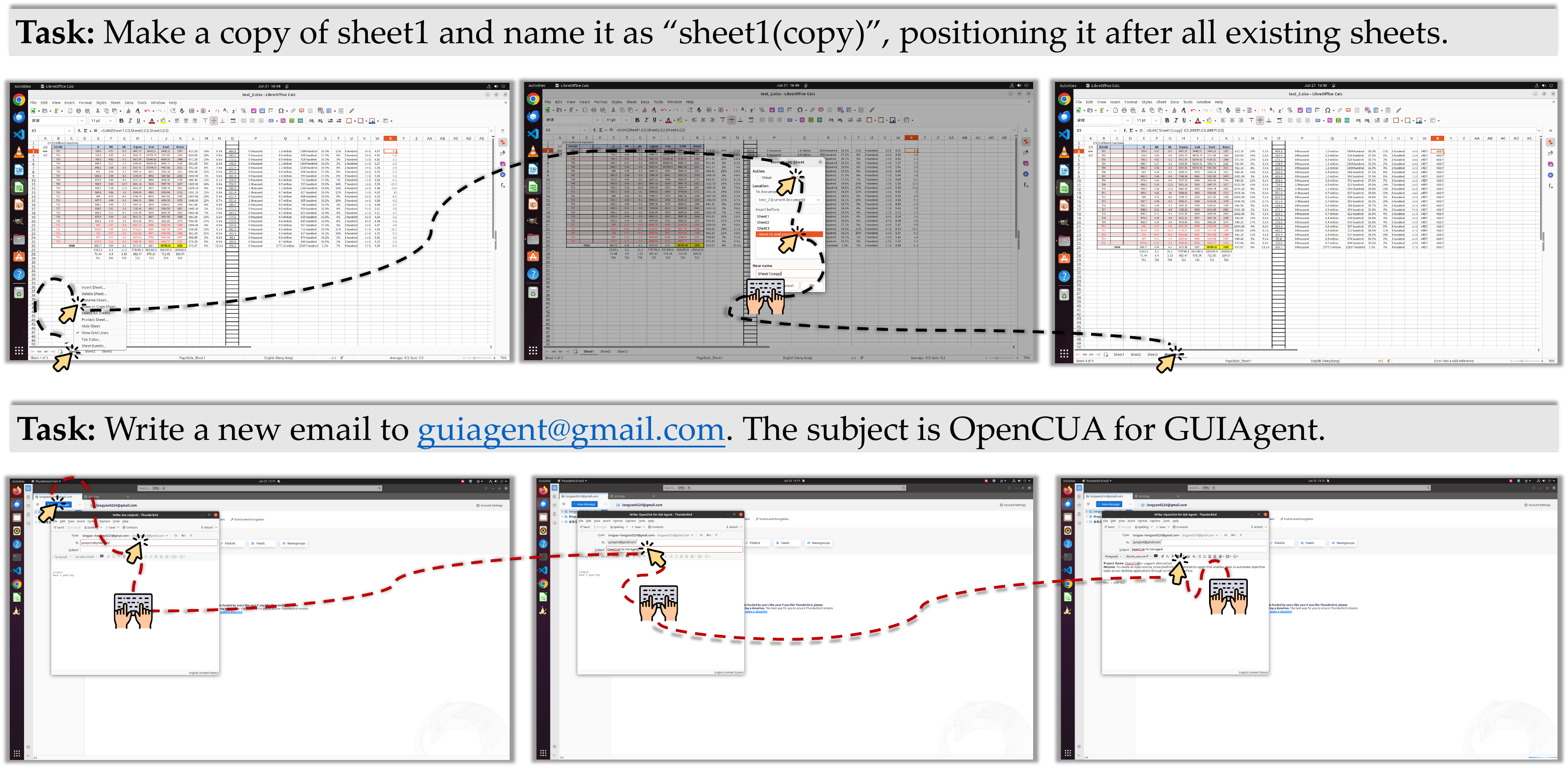}
    \vspace{-0.5em}
\caption{Examples of human-curated trajectories on the Ubuntu platform.}
  \label{fig:ubuntu_trajectory_visualizations}
  \vspace{-0.75em}
\end{figure*}

\begin{figure*}[!h]
  \centering
  \includegraphics[width=0.9\linewidth]{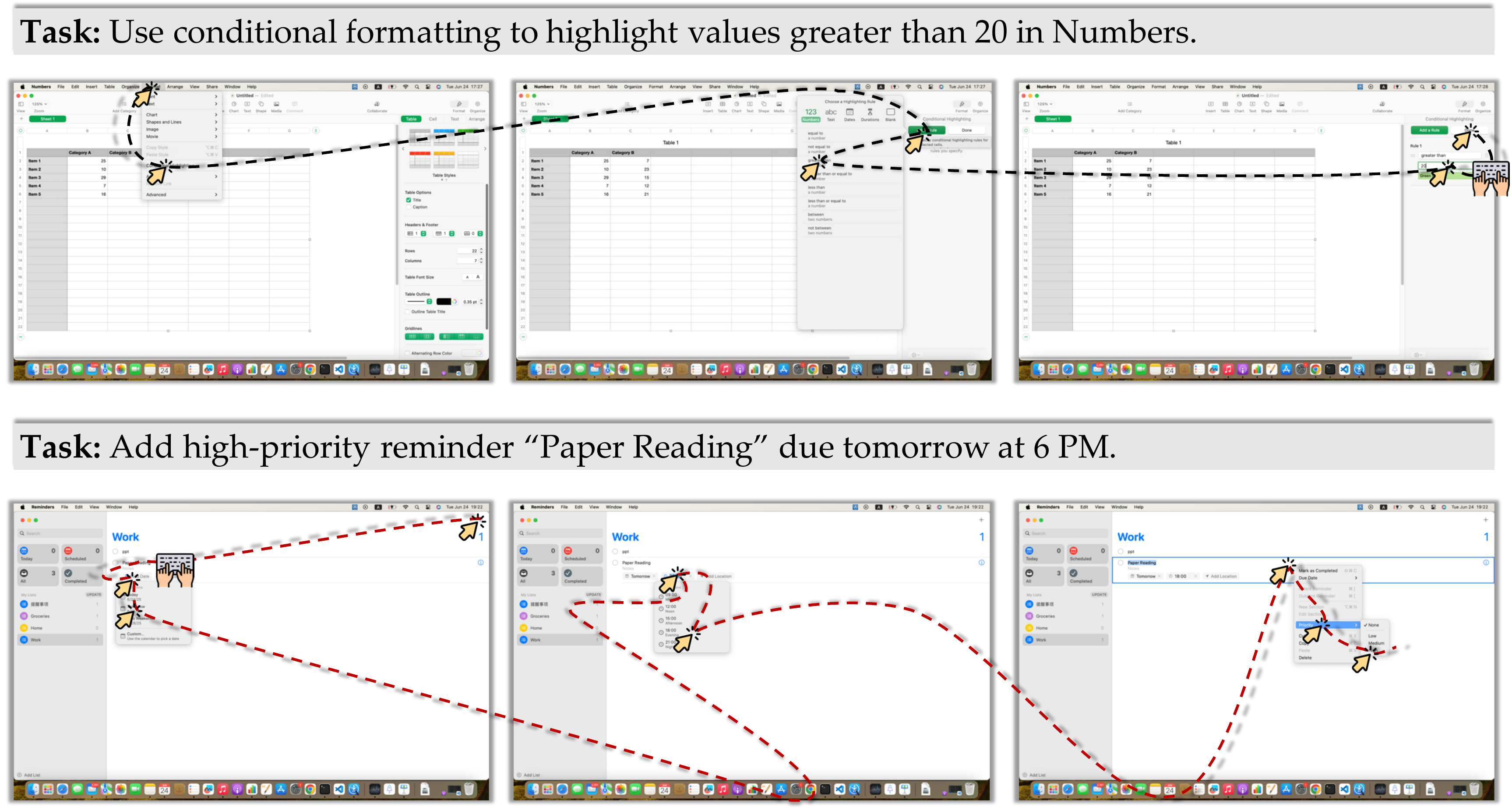}
    \vspace{-0.5em}
\caption{Examples of human-curated trajectories on the macOS platform.}
  \label{fig:macos_trajectory_visualizations}
  \vspace{-0.75em}
\end{figure*}

\begin{figure*}[!h]
  \centering
  \includegraphics[width=0.9\linewidth]{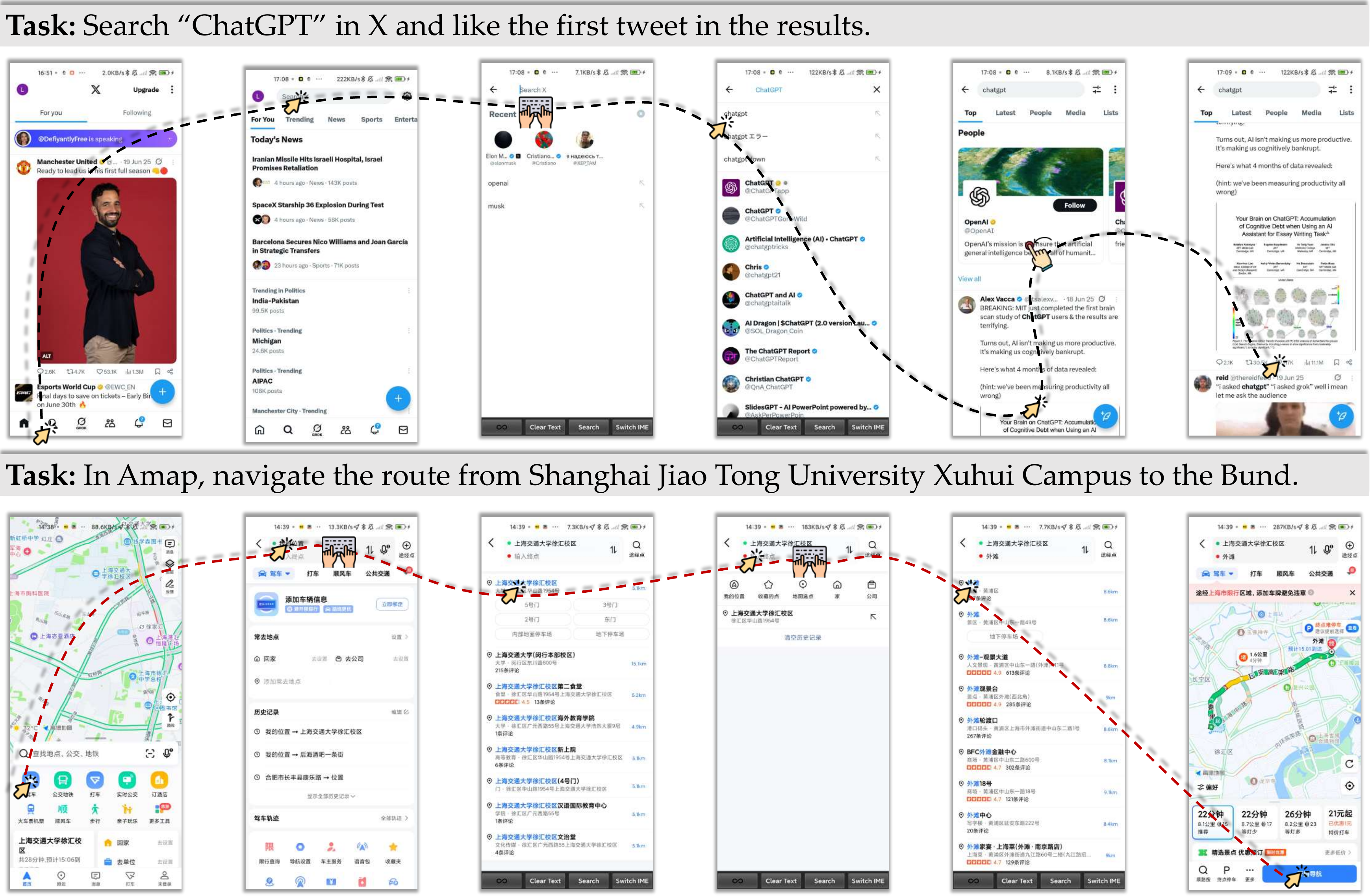}
    \vspace{-0.5em}
\caption{Examples of trajectory data collection on the Android platform.}
  \label{fig:android_trajectory_visualizations}
  \vspace{-0.75em}
\end{figure*}

\begin{figure*}[!h]
  \centering
  \includegraphics[width=0.9\linewidth]{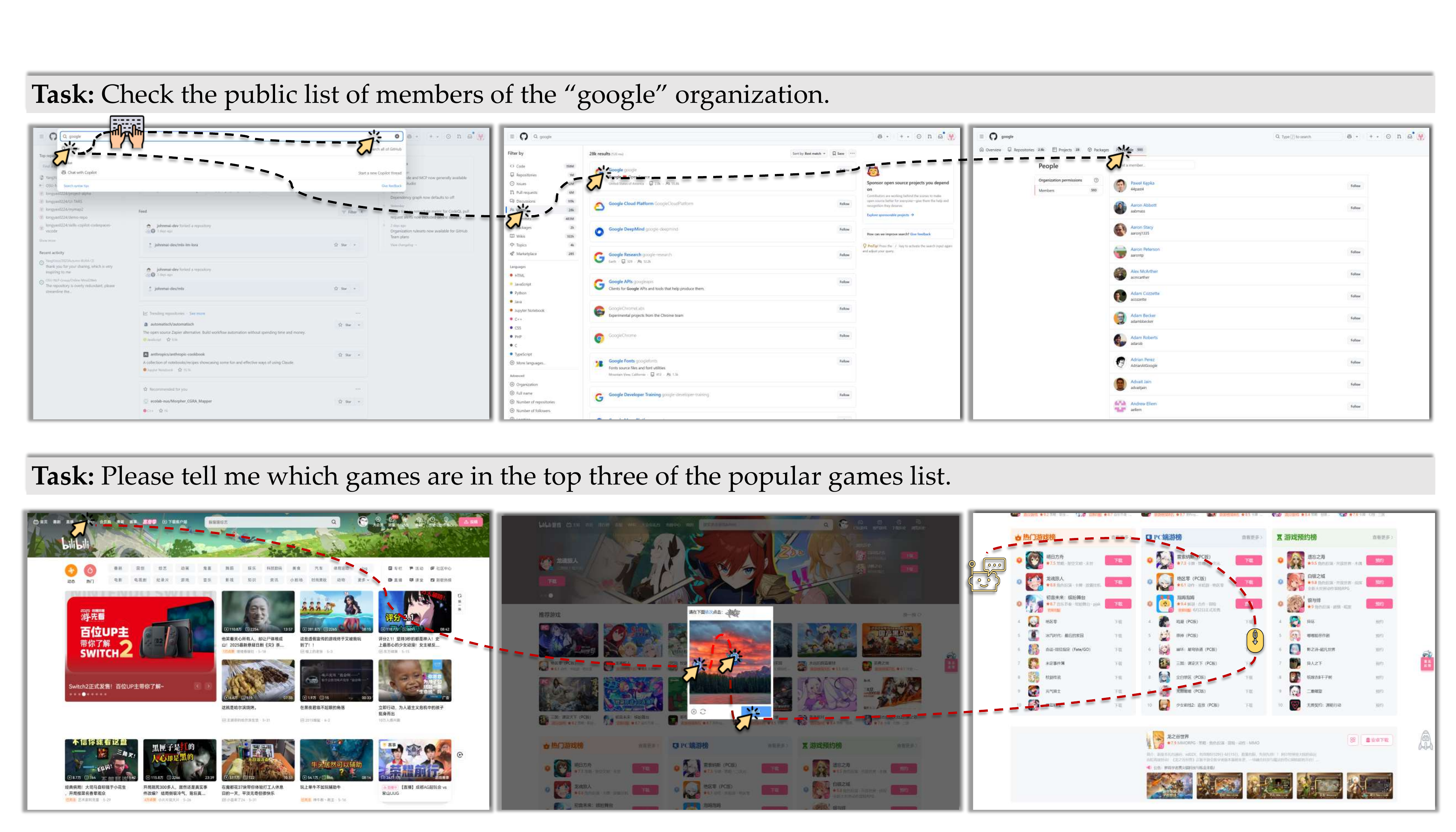}
    \vspace{-0.5em}
\caption{Examples of trajectory data collection on the web platform.}
  \label{fig:web_trajectory_visualizations}
  \vspace{-0.75em}
\end{figure*}

Fig.~\ref{fig:windows_trajectory_visualizations}-\ref{fig:web_trajectory_visualizations} illustrate human-curated trajectories across five platforms: Windows, Ubuntu, macOS, Android and Web. Each trajectory demonstrates precise human-annotated interactions, rendered as \texttt{mouse/gesture} traces over consecutive screenshots, forming high-quality demonstrations for data collection. These trajectories span diverse applications such as Excel, SolidWorks, Gmail, Numbers, Amap, Twitter/X, and GitHub, showcasing real-world complexity in cross-platform environments. The visualizations highlight platform-specific GUI logic (e.g., desktop file operations vs. mobile touch navigation), as well as long-horizon reasoning steps (e.g., multi-page exploration, search-before-edit workflows).

\subsubsection{Trajectory Annotation}
Building upon the annotation schemes detailed in Sec.~\ref{sec:anno_scheme}, we systematically process trajectory data to generate high-quality training corpora. Our trajectory annotation focuses on two key aspects: (1) low-level operational instructions generated for each interaction step, (2) chain-of-thought rationales explaining the decision process. As demonstrated in Table~\ref{tab:vis_trajectory}, these annotations are formally represented using XML tags to distinguish between operational instructions (\texttt{<operation>...</operation>}) and their cognitive justification (\texttt{<think>...</think>}), which support both direct-action execution and reasoned-action paradigms.

\begin{table}[h!]
\centering
\caption{Visualization of trajectory annotation}
\label{tab:vis_trajectory}
\begin{tabular}{|c|m{8cm}|} 
\mytoprule

\multicolumn{2}{|l|}{\textbf{Examples of Two Modes.}} \\
\mymidrule

\raisebox{-0.5\height}{\includegraphics[width=0.35\textwidth]{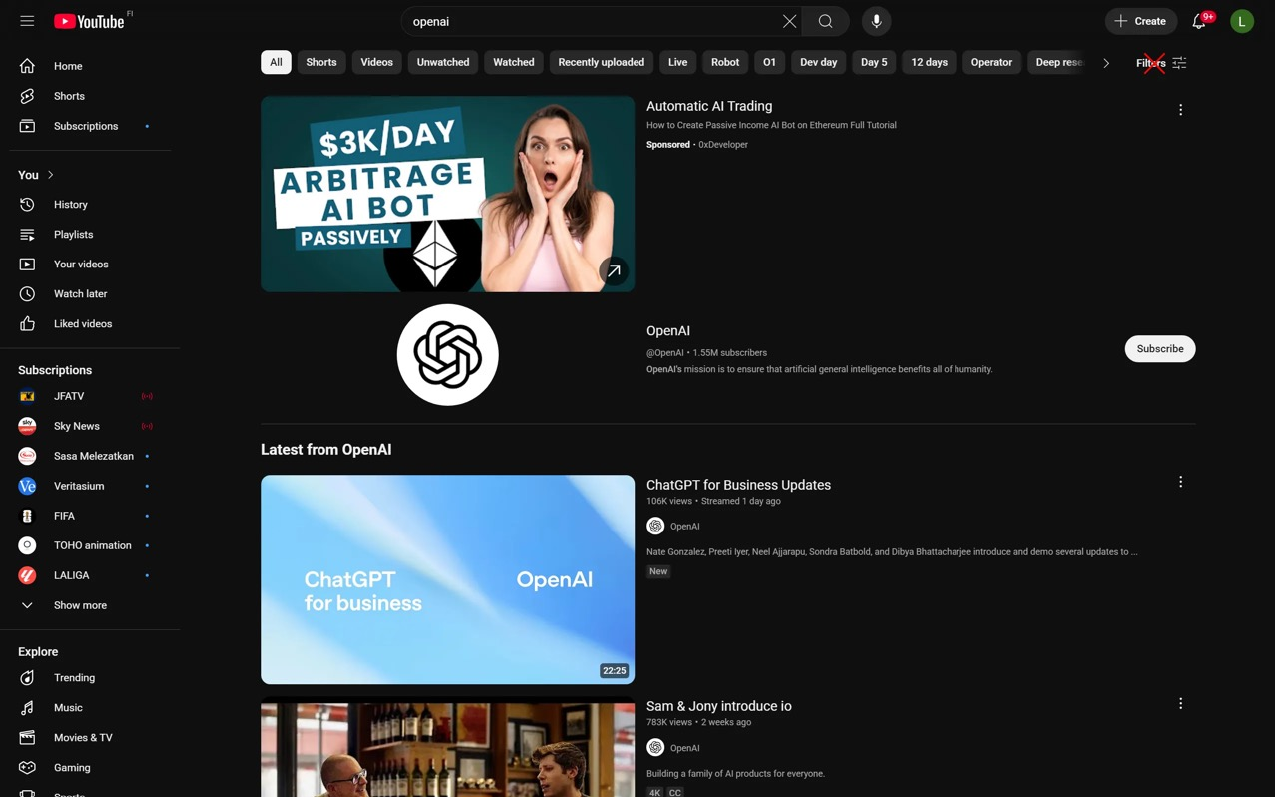}} & 
\underline{[Reasoned-Action Mode]}

\textbf{Q:}
Please generate the next move according to the UI screenshot, task and previous operations.

Task: 

Search openai and like the first video, sorted by view count and video type.

Previous operations:

Step 1: Click on the search box at the top of the YouTube page and type "openai" 

Step 2: Search for OpenAI on YouTube and press Enter to see search results and the OpenAI channel page

\textbf{A:}
\textless think\textgreater 

The YouTube interface shows a search for "openai" with a filters button visible in the top navigation area. Clicking on the filters option would allow sorting videos by criteria such as view count, which is needed to complete the task of finding the most viewed OpenAI videos. This filtering functionality is essential to organize search results in a way that aligns with the requirement to sort by view count and video type before liking the first video. 

\textless /think\textgreater 

\textless operation\textgreater 

Click on the "Filters" button at the top right of the YouTube search results to access advanced filtering options for your search. 

\textless /operation\textgreater 

\textless action\textgreater 

click(x=0.9043, y=0.0788) 

\textless /action\textgreater

\\
\mymidrule
\raisebox{-0.5\height}{\includegraphics[width=0.35\textwidth]{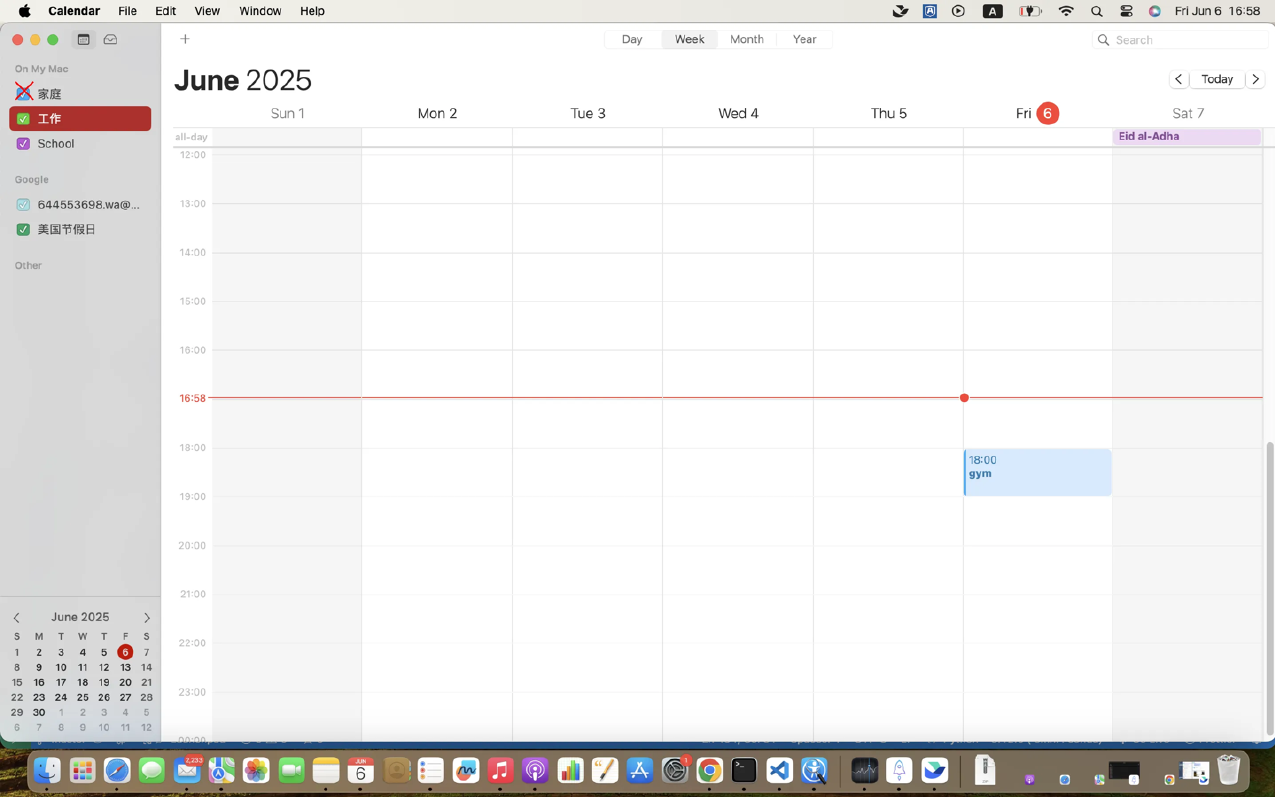}} &
\underline{[Direct-Action Mode]}

\textbf{Q:}
Please generate the next move according to the UI screenshot, task and previous operations. 

Task: 

In the Calendar app, show only events in the 'School' calendar. 

Previous operations: 

None

\textbf{A:}
\textless operation\textgreater

Click on the blue checkbox next to "Family" in the calendar sidebar under "On My Mac" section.

\textless /operation\textgreater  

\textless action\textgreater 

click(x=0.0187, y=0.1128) 

\textless /action\textgreater 
\\

\mybottomrule

\end{tabular}
\end{table}

% \subsection{Case Study}

\subsection{Lessons from Data Acquisition}
% \todo{@all}
\subsubsection{Common Problems}
\begin{figure*}[!h]
  \centering
  \includegraphics[width=0.9\linewidth]{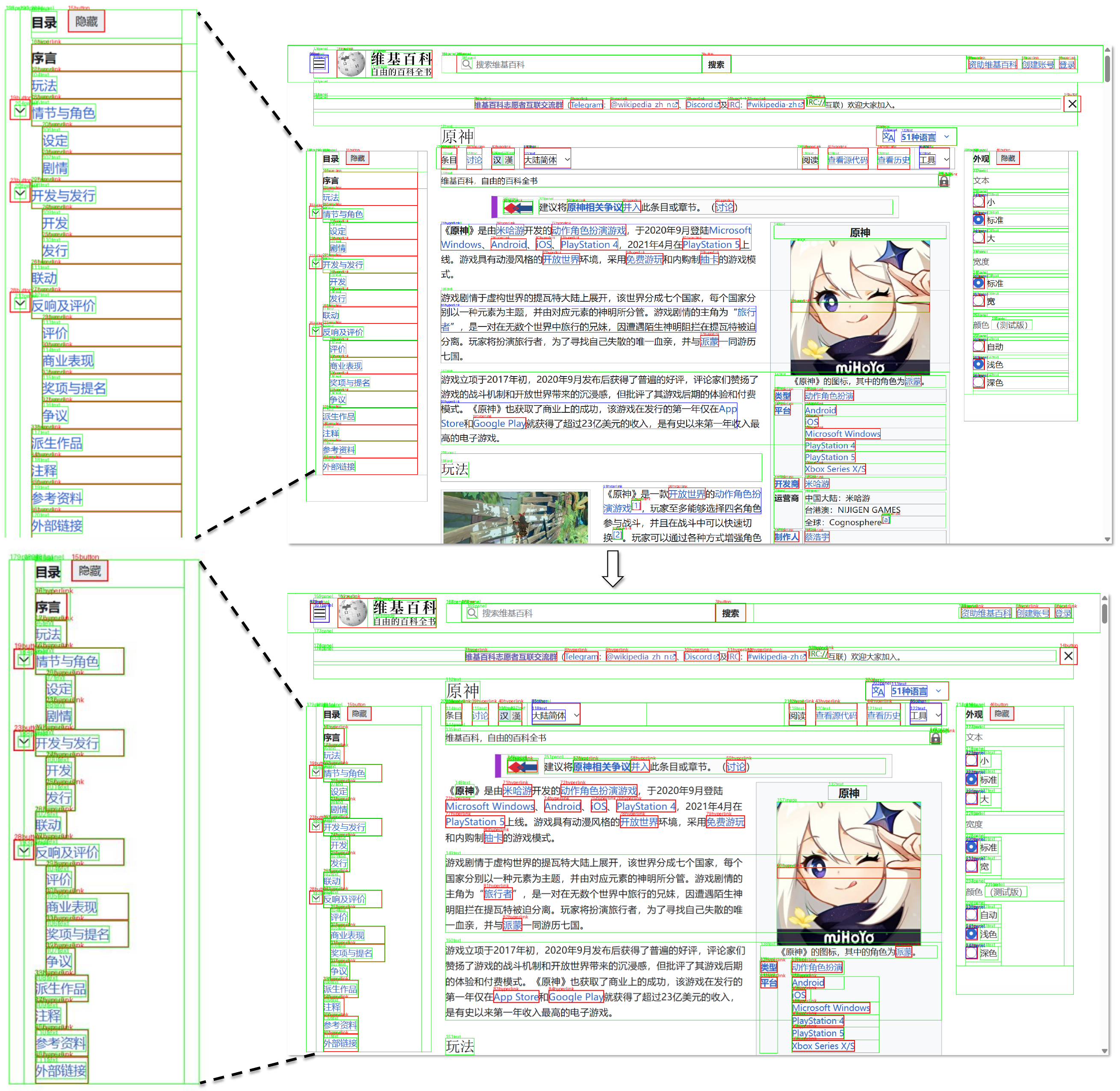}
    \vspace{-0.5em}
\caption{Examples of adaptive bounding box tightening(ABT) algorithm.}
  \label{fig:abt_algorithm}
  \vspace{-0.75em}
\end{figure*}

\textbf{Adaptive Bounding Box Tightening(ABT).} 
The heterogeneity across platforms results in significant discrepancies in UI interaction logic and responsive regions. Therefore, we aim to ensure that the bounding boxes of UI elements are as close as possible to their central regions, in order to minimize inconsistencies in actionable areas across different platforms.
As illustrated at the top of Fig.~\ref{fig:abt_algorithm}, these parsed bounding boxes are aligned with the boundaries of the UI elements; however, their centers, which serve as the ground truth for click prediction, often reside in empty regions rather than in areas with meaningful content.
This misalignment introduces a detrimental click-whitespace bias during model training. To mitigate this issue, we propose the Adaptive Bounding Box Tightening (ABT) algorithm. 
ABT dynamically refines bounding boxes through iterative, proportional contraction of their lateral boundaries. This contraction is guided by the detection of contiguous uniform-color regions, representing target whitespace, within defined tolerance thresholds. The process terminates when no significant contractible uniform regions remain, yielding substantially tighter bounding boxes where centers align with actual content, as shown in the bottom of Fig.~\ref{fig:abt_algorithm}. While ABT's effectiveness is inherently limited by complex backgrounds and gradients, empirical validation confirms its significant value in improving ground truth alignment for interfaces featuring simple solid-color backgrounds. This paradigm remains dominant in modern systems and web design.

\begin{figure*}[!h]
  \centering
  \includegraphics[width=0.9\linewidth]{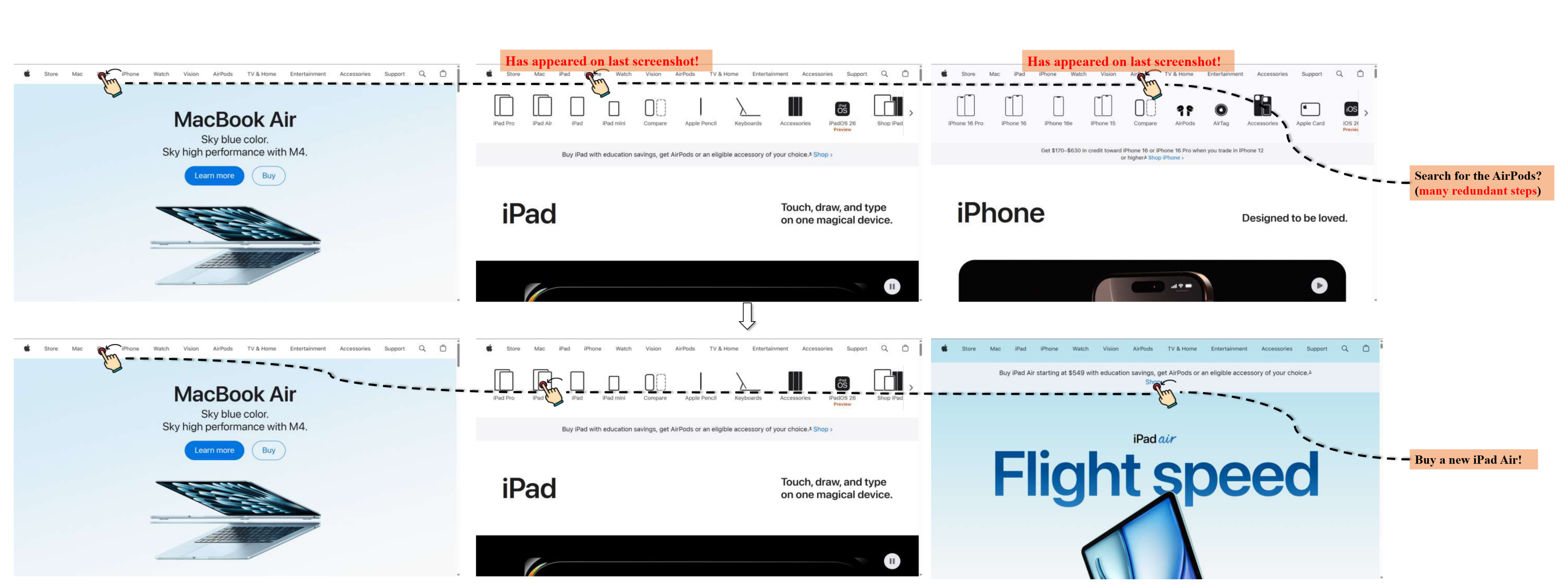}
    \vspace{-0.5em}
\caption{Examples of deep exploration algorithm.}
  \label{fig:deep_exploration_algorithm}
  \vspace{-0.75em}
\end{figure*}

\textbf{Deep Exploration.} 
Modeling GUI platform state transition graphs presents inherent complexity. Random walks, a common approach, suffer from limitations: unpredictable transitions induce pervasive back edges, causing frequent state revisitation or trapping in local loops due to insufficient backtracking mechanisms. To address these issues and enable automated deep exploration for acquiring meaningful weakly semantic trajectories, we propose a single-history-frame element filtering algorithm.
Specifically, we use a queue to maintain all interactive elements appearing in the last screenshot. 
At each exploration step during random walk, some of elements are filtered out when their Intersection over Union (IoU) exceeds a predefined threshold and their textual content exactly matches any element in the queue. 
This guarantees exclusive interaction with elements absent in the preceding state, thereby actively steering exploration toward novel pages. 
This mechanism proves particularly effective for interfaces with persistent components (e.g., navigation bars, sidebars) or dense icon arrays, as evidenced in Fig.~\ref{fig:deep_exploration_algorithm} where it achieves significantly broader page coverage and yields non-redundant, semantically valuable trajectories compared to conventional random walks.

\subsubsection{Windows}
\begin{figure*}[!h]
  \centering
  \includegraphics[width=0.9\linewidth]{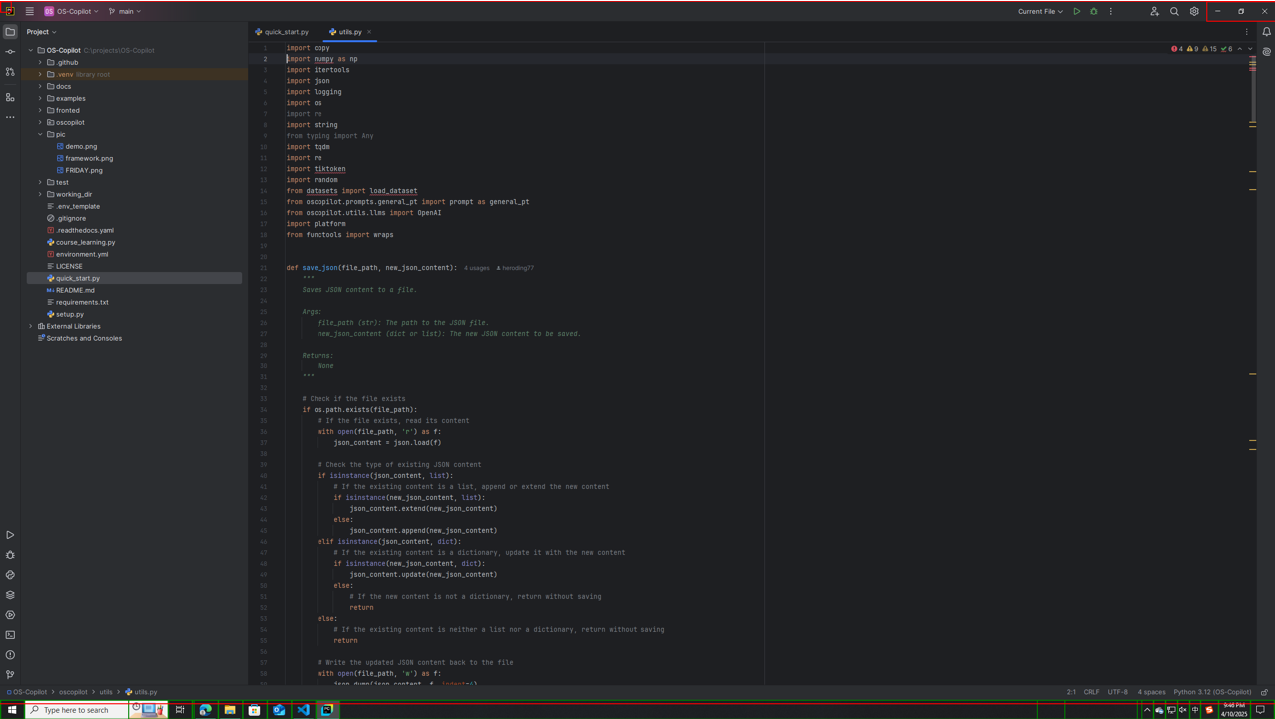}
    \vspace{-0.5em}
\caption{An example of \texttt{Win32} API failing to parse A11y Trees in PyCharm.}
  \label{fig:pycharm_win32}
  \vspace{-0.75em}
\end{figure*}

\begin{figure*}[!h]
  \centering
  \includegraphics[width=0.9\linewidth]{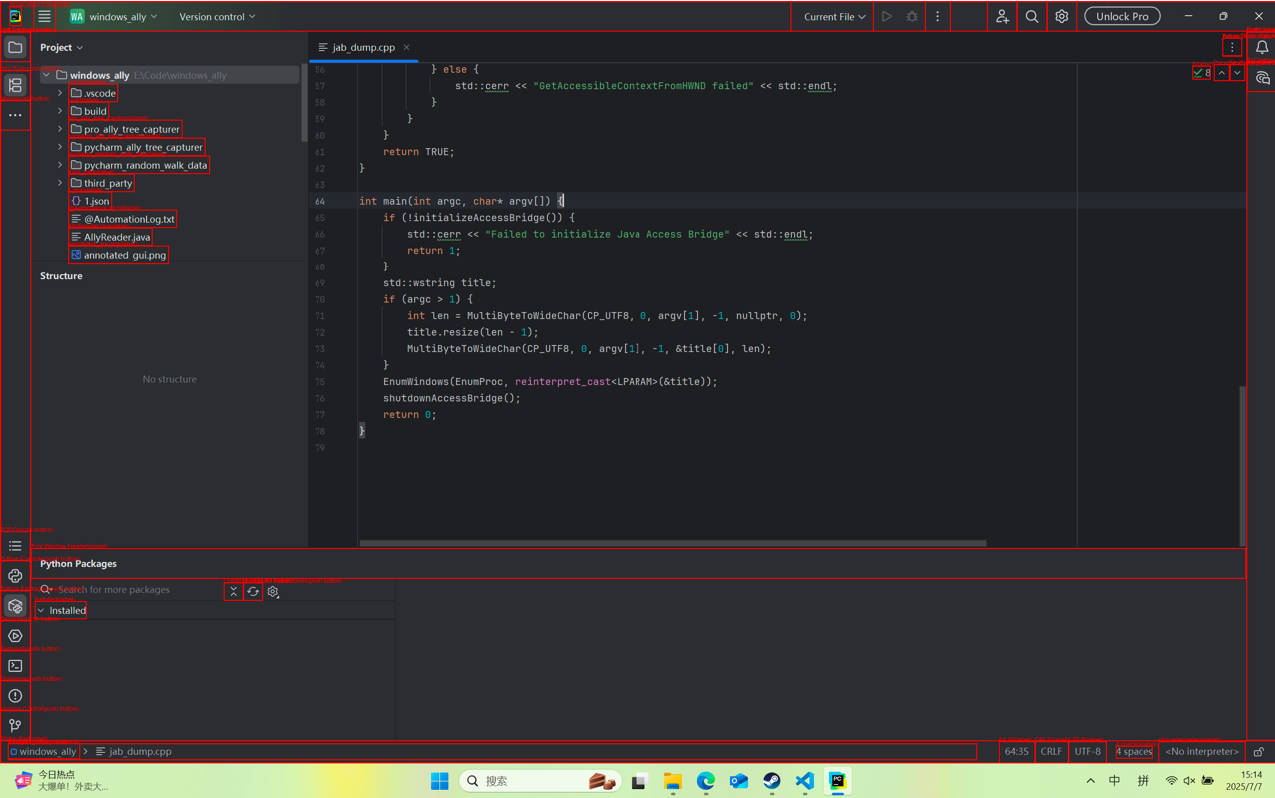}
    \vspace{-0.5em}
\caption{An example of \texttt{JAB} API successfully parsing A11y Trees in PyCharm.}
  \label{fig:pycharm_jab}
  \vspace{-0.75em}
\end{figure*}

\begin{figure*}[!h]
  \centering
  \includegraphics[width=0.9\linewidth]{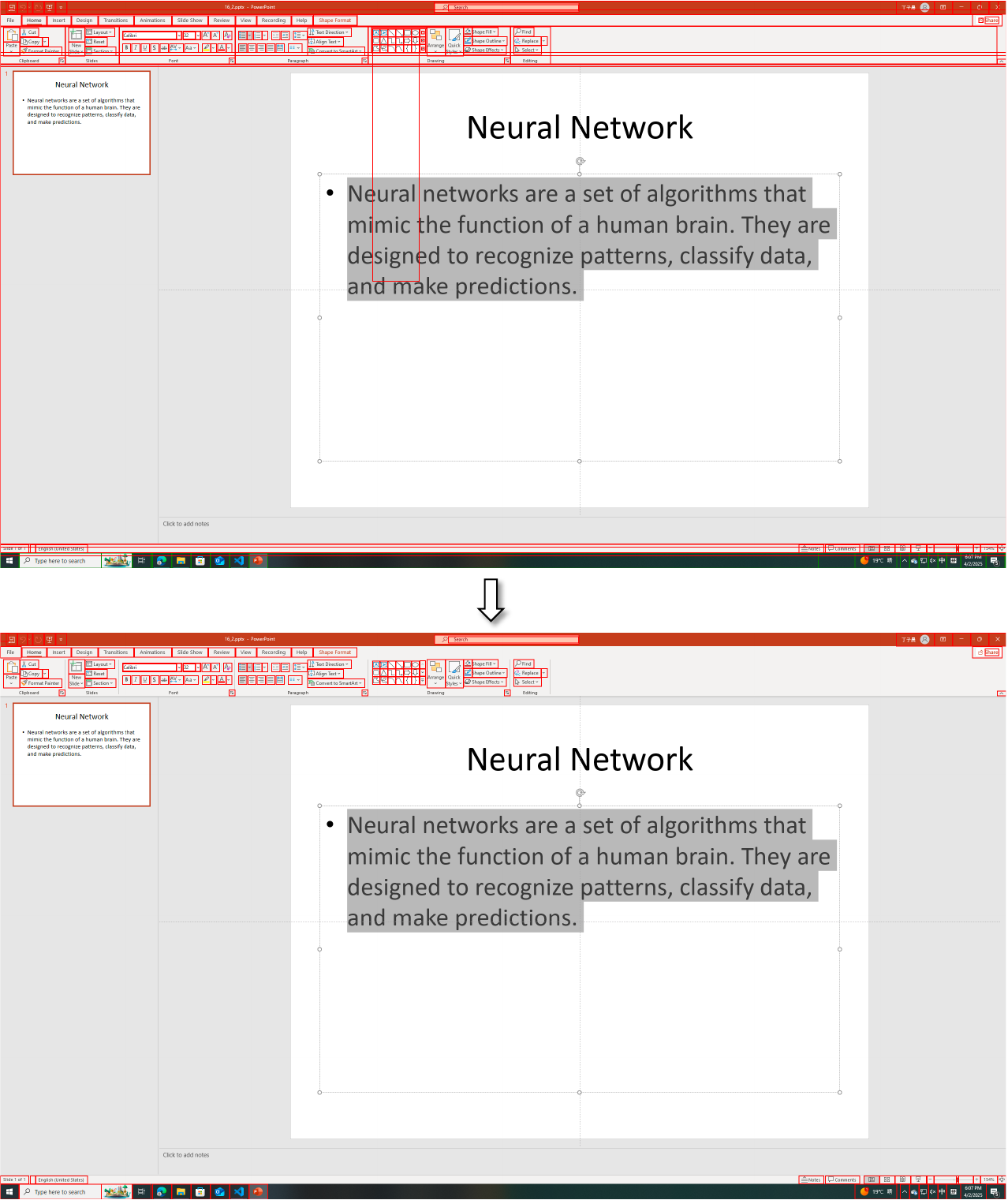}
    \vspace{-0.5em}
\caption{An example of denoising on Windows, transforming a raw A11y Tree~(top) into a clear structure~(down).}
  \label{fig:windows_filter_processing}
  \vspace{-0.75em}
\end{figure*}

\textbf{Cross-Framework UI Parsing Challenges and Denoising Strategies.} 
When processing Java-based software like PyCharm and Android Studio, the standard \texttt{Win32} API exposes significant limitations. As illustrated in Fig.~\ref{fig:pycharm_win32}, the \texttt{Win32} API fails to effectively parse their UI structure, resulting in an incomplete A11y Tree. Consequently, we must switch to using the specialized Java Access Bridge~(\texttt{JAB}\footnote{https://docs.oracle.com/javase/8/docs/technotes/guides/access/jab/index.html}) API. The JAB successfully retrieves the complete A11y Tree (as shown in Fig.~\ref{fig:pycharm_jab}, thus resolving the issue. This requirement to adapt different APIs for various application frameworks significantly increases the complexity of our data collection efforts. Moreover, the raw A11y Trees present challenges: they are typically deeply nested, noisy, and the density of functionally relevant UI elements is low. To address these issues and improve data quality, we apply a set of heuristic filters to prune and refine the tree. ~\ref{fig:windows_filter_processing} visualizes this transformation, showing a comparison of the A11y Tree before and after processing. Our filtering strategies exclude elements from background applications and select elements by their screen-to-area ratio, roles~(e.g., \texttt{button}, \texttt{text}, \texttt{hyperlink}).

\textbf{Data Deduplication and Geometric Refinement.} Data acquisition in Windows faces several significant data quality challenges. First, minimal UI changes following user interactions lead to high redundancy of UI elements due to nearly identical screen captures. Second, lack of layer information in the A11y Trees results in erroneous inclusion of occluded elements~(e.g., \texttt{dropdown}).To overcome these challenges, we implemented a multi-stage refinement pipeline. We first mitigate redundancy with a similarity algorithm that filters images based on the Euclidean distance of their feature vectors. A post-processing filter then identifies occluded elements by detecting solid-color regions within their bounding boxes.

\textbf{Prioritized Random Walk for Automated UI Exploration.} The random walk algorithm is central to our automated data acquisition on the Windows platform. To minimize redundant interactions and enhance element diversity, we have augmented the standard Random Walk with principles from Depth-First Search (DFS). As mentioned in the above common problems, our modified algorithm prioritizes interaction with newly appeared UI elements while concurrently reducing the selection priority of elements that have already been interacted with. If no new elements are detected, or if their count falls below a predefined threshold, the algorithm defaults to interacting with any remaining, previously unvisited elements within the current view's A11y Trees. Furthermore, we account for scenarios where interactions navigate away from the primary application, such as launching a web browser to view a user manual. 
In such cases, our algorithm allows for limited interaction within the external application~(e.g., the browser) before automatically shifting focus back to continue navigating the initial application.

\subsubsection{Ubuntu}
This section details the challenges encountered and solutions developed for autonomous agent interaction with the Ubuntu environment. The primary challenges originate from the inherent structure of the accessibility tree (A11y tree), which serves as the main interface for observing and interacting with the application. Our solutions focus on refining the accessibility tree data and optimizing the agent's interaction strategy to ensure reliable and efficient operation. The successful resolution of these issues is paramount, as a clean, accurate, and efficiently navigable UI representation is the foundation for any effective automated UI-based task.

\begin{figure*}[!h]
  \centering
  \includegraphics[width=0.9\linewidth]{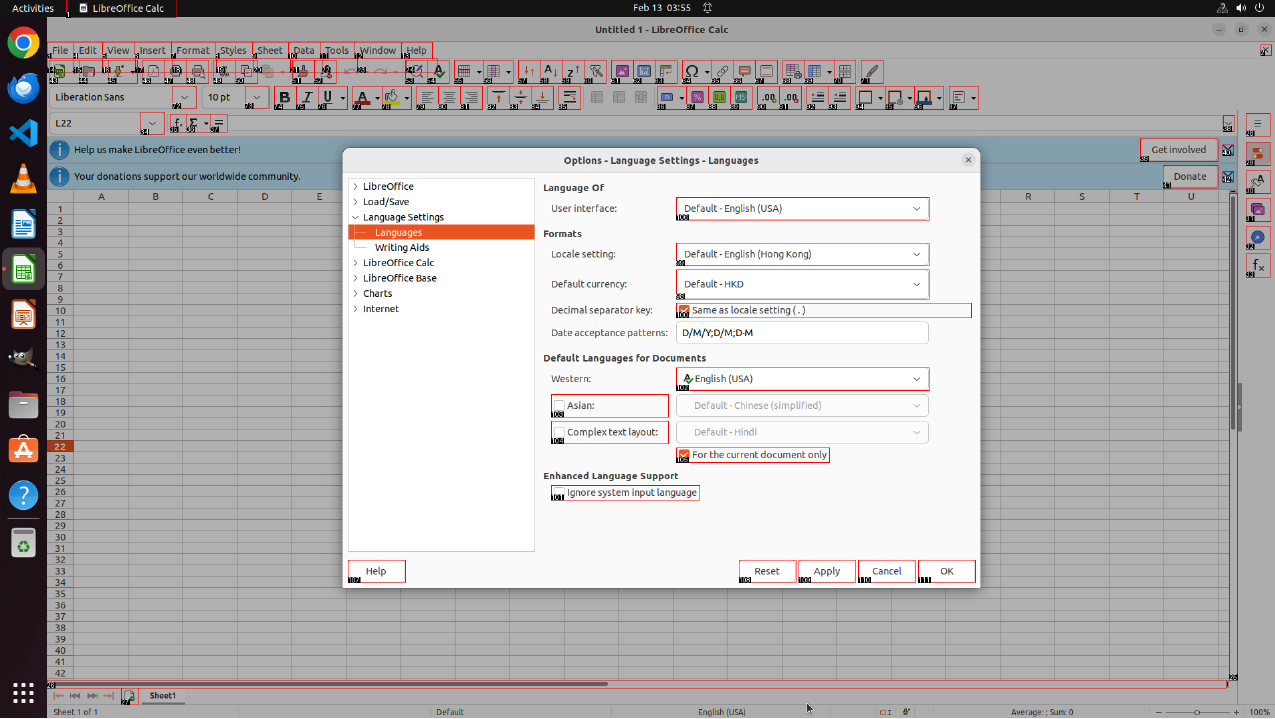}
    \vspace{-0.5em}
\caption{Examples of visual occlusion and invalidity of elements.}
  \label{fig:ubuntu_unvalid}
  \vspace{-0.75em}
\end{figure*}

\begin{figure*}[!h]
  \centering
  \includegraphics[width=0.9\linewidth]{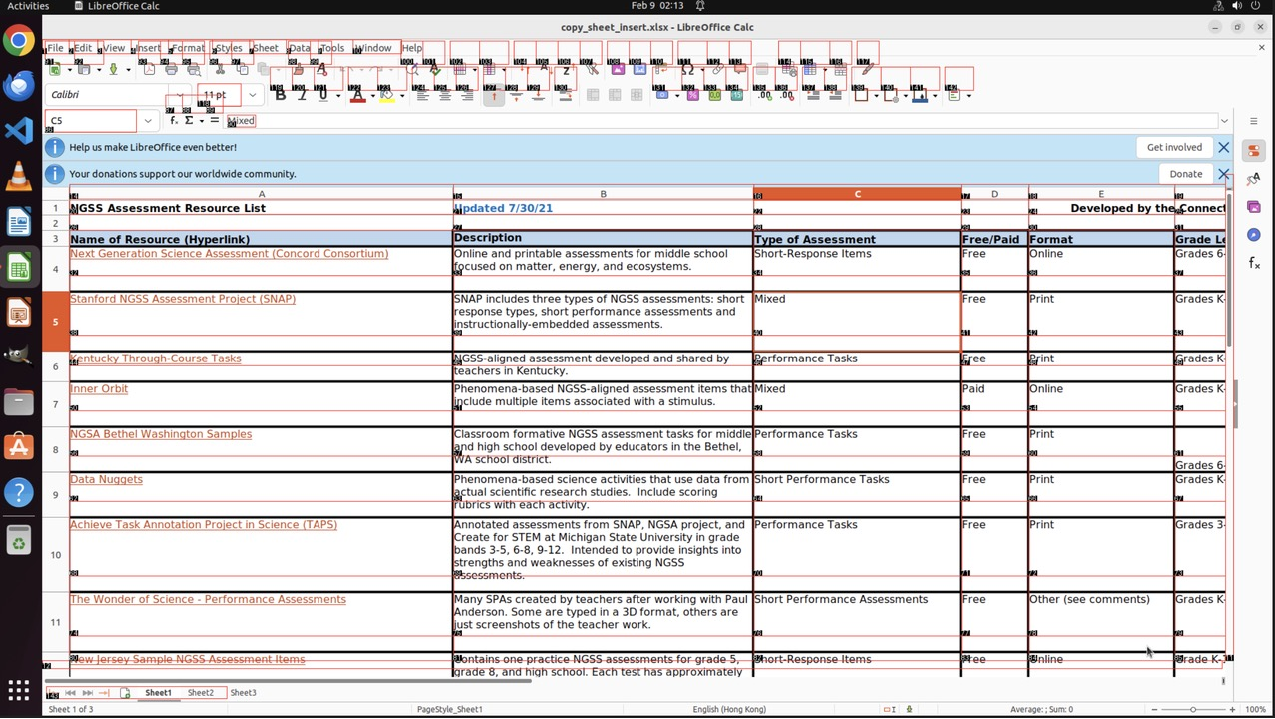}
    \vspace{-0.5em}
\caption{Examples of coordinate offset.}
  \label{fig:ubuntu_offset}
  \vspace{-0.75em}
\end{figure*}

\textbf{Denoising in the Accessibility Tree.} 
The raw data provided by the accessibility tree on Ubuntu is often noisy, containing redundant information and occasional inaccuracies that can mislead an autonomous agent. We identified and implemented solutions for three primary issues. First, the A11tree's hierarchical structure often includes redundant parent elements that do not correspond to distinct interactive components. This is particularly prevalent in applications built with Web, such as Chrome. To address this, we apply a two-stage filtering process. 
We begin by pruning elements whose roles are typically non-interactive or structural based on \texttt{type}, such as `heading', `paragraph', and `section'.
This denoising is critical because it exposes the true, underlying interactive elements, preventing the agent from attempting to interact with large, non-interactive container widgets.
Subsequently, we analyze the geometric relationships of the remaining elements' bounding boxes. If an element's bounding box is significantly occluded by a smaller one (i.e., the smaller box's area occupies a large percentage of the larger box's area), we infer a container-child relationship and discard the larger, containing element. 
Second, the standard accessibility tree does not inherently account for the visual occlusion and invalidity of elements. An element may be present in the tree but be completely obscured by another element on the screen as shown in Fig.~\ref{fig:ubuntu_unvalid}. We tackle this with LLMs. 
Third, we observed that for certain applications, the accessibility tree reports incorrect coordinates for all UI elements immediately after the application is launched, as shown in Fig.~\ref{fig:ubuntu_offset}. The entire tree appears to have a coordinate offset. Through empirical testing, we discovered a practical solution: initiating a short sequence of random interactions within the application window causes the accessibility tree's coordinate system to recalibrate, restoring correct positional data. Ensuring coordinate accuracy is fundamental; without it, any attempt by the agent to click or type at a specific location would fail, rendering automation impossible.

\textbf{GUI Exploration Optimization via an Improved Random Walk.}
A pure random walk over all available UI elements is highly inefficient. To improve the agent's ability to explore an application's state space, we developed a more intelligent interaction strategy. This strategy is based on filtering the action space and prioritizing the exploration of novel UI states.
To reduce the number of futile actions, the agent's action space is constrained to only include elements that are designated as interactive. We maintain a whitelist of interactive \texttt{type}, including `button', `box', `menu', `entry', `link', `bar', and `item'. Conversely, elements with non-interactive roles are excluded from the potential action set. These non-interactive roles include `heading', `static', `document', `label', `cell', `text', `icon', `paragraph', and `section'.
To prevent the agent from becoming trapped in interaction loops within a static UI state, we implemented a state-aware exploration logic. After the agent acts, we only visit newly appeared UI elements. 
These novel elements are given interaction priority, as they are most likely to lead to a new application state. If the action does not yield any new elements, the agent then selects an action randomly from the set of previously known elements that it has not yet interacted with in the current state. 
This process continues until all interactive elements have been exhausted. This exploration strategy is vital for efficiency, as it directs the agent towards discovering new functionalities and application states, thereby maximizing the coverage of the application's features in a limited time and avoiding redundant, non-productive interactions.

\subsubsection{MacOS}
\begin{figure*}[!h]
  \centering
  \includegraphics[width=0.9\linewidth]{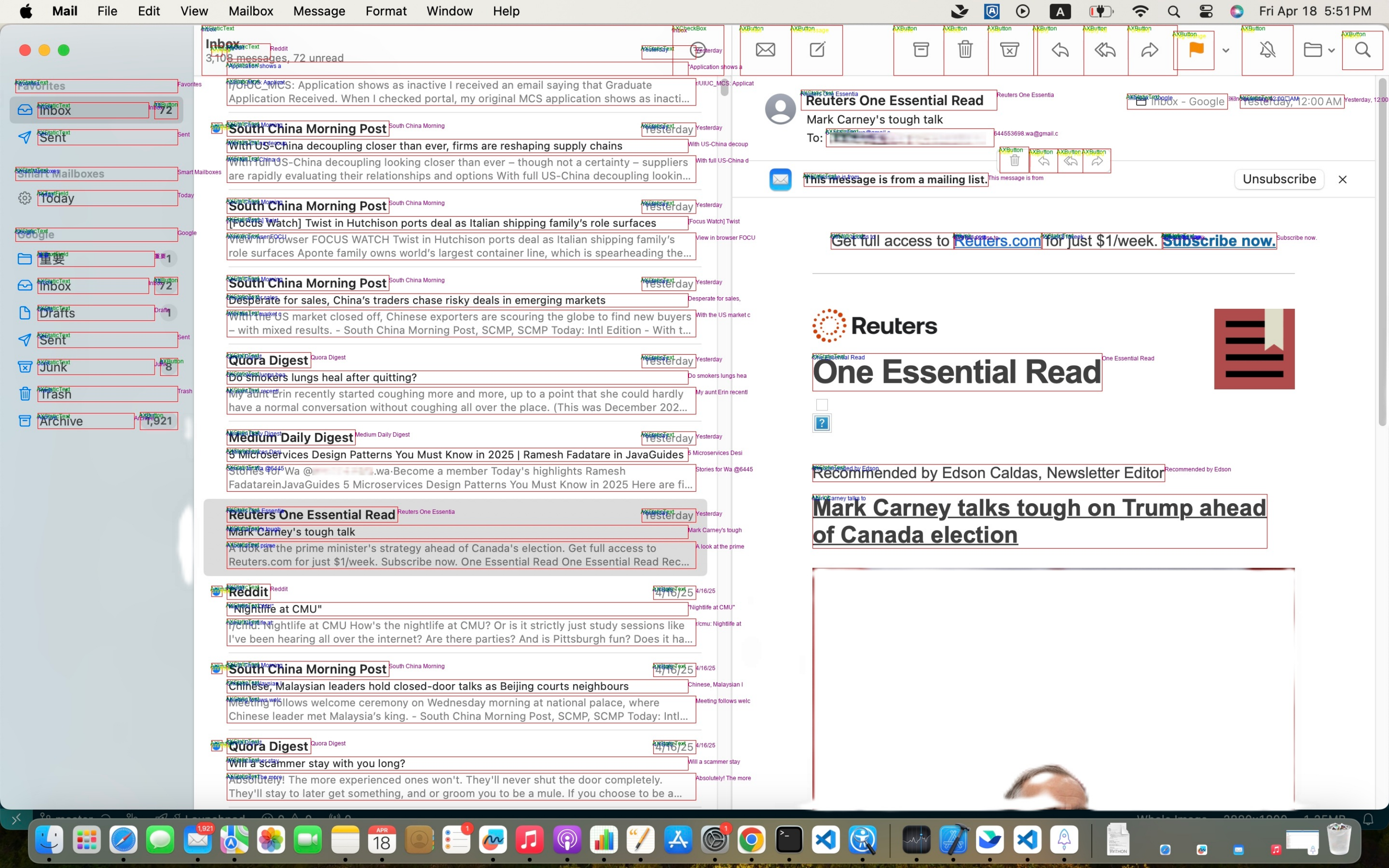}
    \vspace{-0.5em}
  \caption{Refined \texttt{AXTree} overlay on the Mail application: all interactive elements are tightly bound after heuristic pruning.}
  \label{fig:mac_ally_vis}
  \vspace{-0.75em}
\end{figure*}

\textbf{Robust A11y Tree Extraction and Denoising.} % 
The macOS pipeline first locates the active top-level window, then exhaustively traverses its accessibility hierarchy. Every bounding box is mapped from logical coordinates to device pixels by multiplying by the screen-scale factor. After flattening the tree, only nodes whose roles are interactive (e.g.\ \texttt{AXButton}, \texttt{AXPopUpButton}, \texttt{AXTextField}) are retained. Moreover, we would discard boxes with a width or height of 2px or less and remove nodes whose \texttt{text}, \texttt{description}, and \texttt{value} are all empty or punctuation. A role-aware merging process replaces overlapping AXStaticText siblings and their interactive parent with a minimal bounding box. The resulting set contains clean, tightly localised interactive elements.
(see Fig.~\ref{fig:mac_ally_vis}).

\begin{figure*}[!h]
  \centering
  \includegraphics[width=0.9\linewidth]{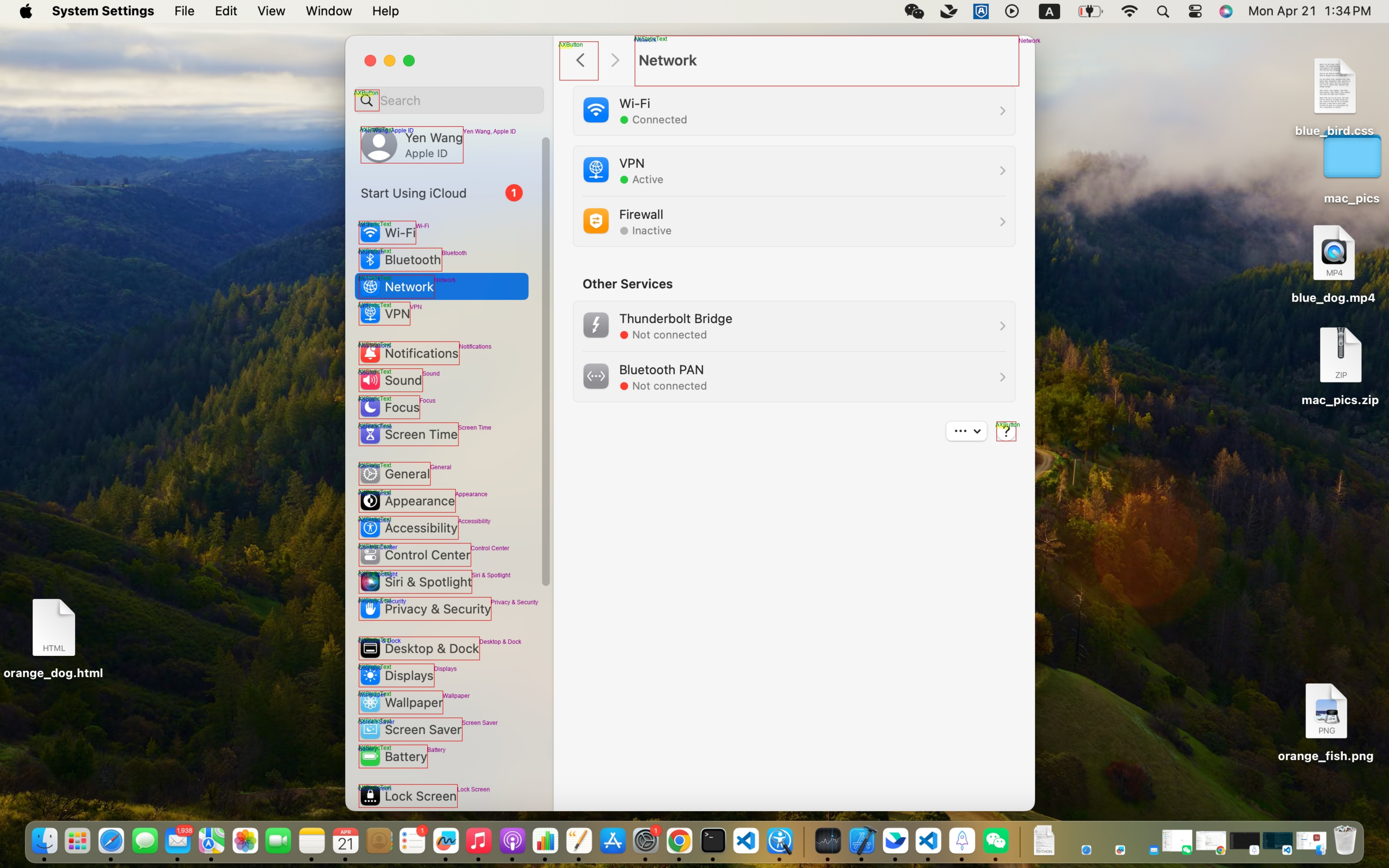}
    \vspace{-0.5em}
  \caption{The failure case in System Settings: the \texttt{AXTree} omits right-pane controls, illustrating the necessity of Omniparser fusion.}
  \label{fig:mac_ally_badcase}
  \vspace{-0.75em}
\end{figure*}

\textbf{Hybrid A11y Tree \& Omniparser combination for System Panels.} %
Several built-in utilities, most notably \emph{System Settings}, draw controls in private layers that have no corresponding accessibility tree~\cite{yu2025omniparser}, as shown in Fig.~\ref{fig:mac_ally_badcase}. To recover these missing widgets, each screenshot is processed by Omniparser, yielding a set of vision-detected bounding boxes. An element would be retained when its IoU with any Omniparser box exceeds $0.15$ or when it is selected during exploration. This combination renders previously invisible elements in the A11y tree, thereby yielding a more comprehensive understanding of macOS applications.

\subsubsection{Mobile}
\begin{figure*}[ht]
  \centering
  \begin{subfigure}[t]{0.18\textwidth}
    \centering
    \includegraphics[height=6cm]{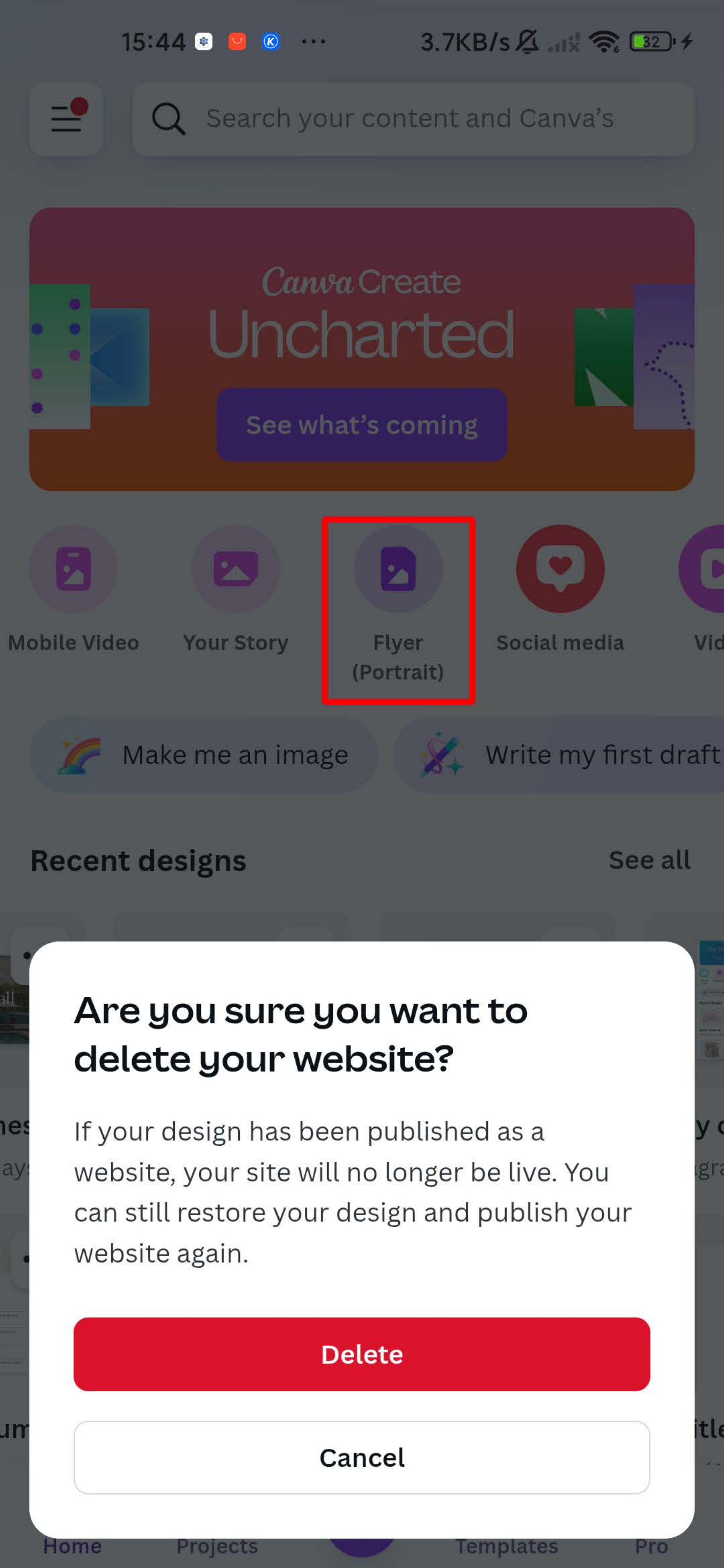}
    \caption{Mobile element occlusion}
    \label{fig:mobile_occlusion}
  \end{subfigure}
  \hspace{0.01\textwidth}
  \begin{subfigure}[t]{0.18\textwidth}
    \centering
    \includegraphics[height=6cm]{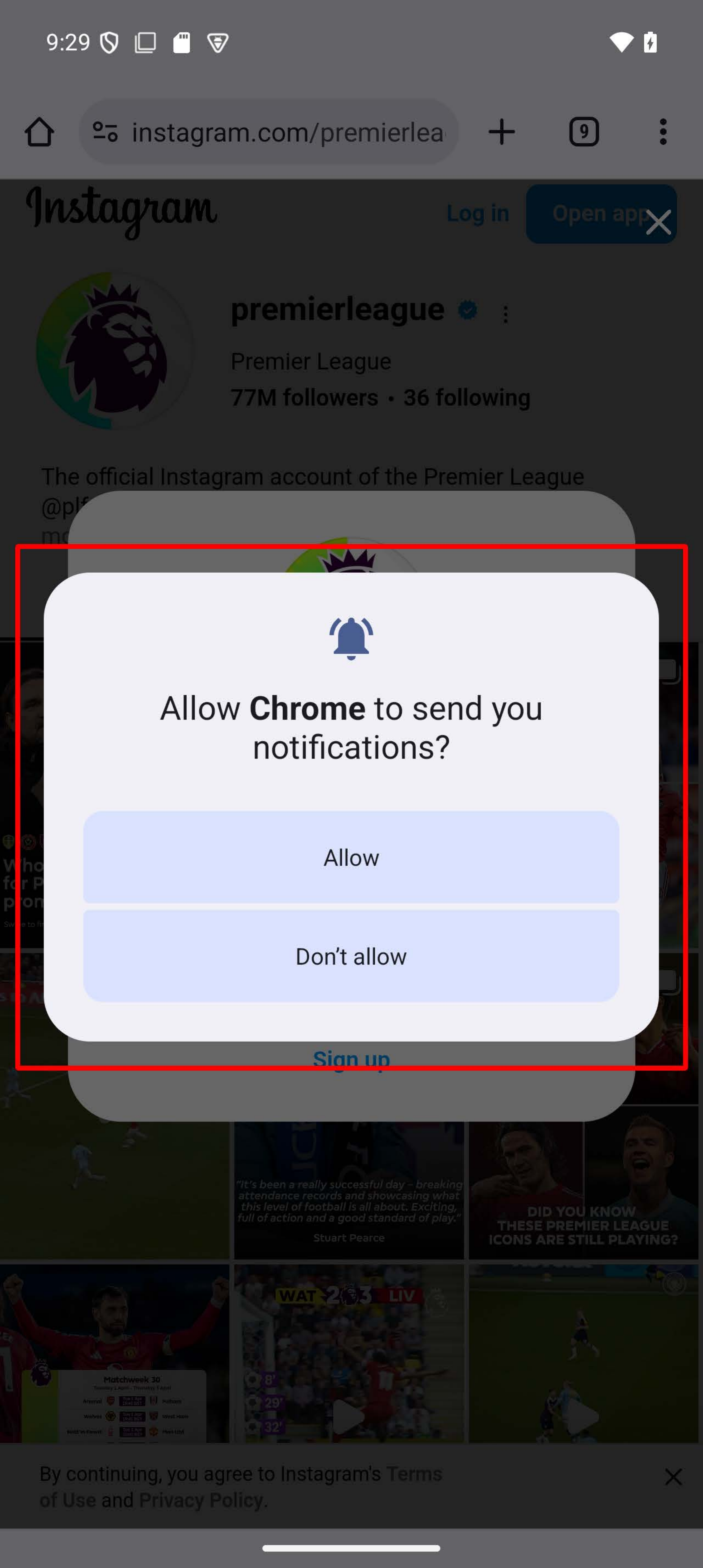}
    \caption{Mobile element attribute loss}
    \label{fig:mobile_missing}
  \end{subfigure}
  \hspace{0.01\textwidth}
  \begin{subfigure}[t]{0.18\textwidth}
    \centering
    \includegraphics[height=6cm]{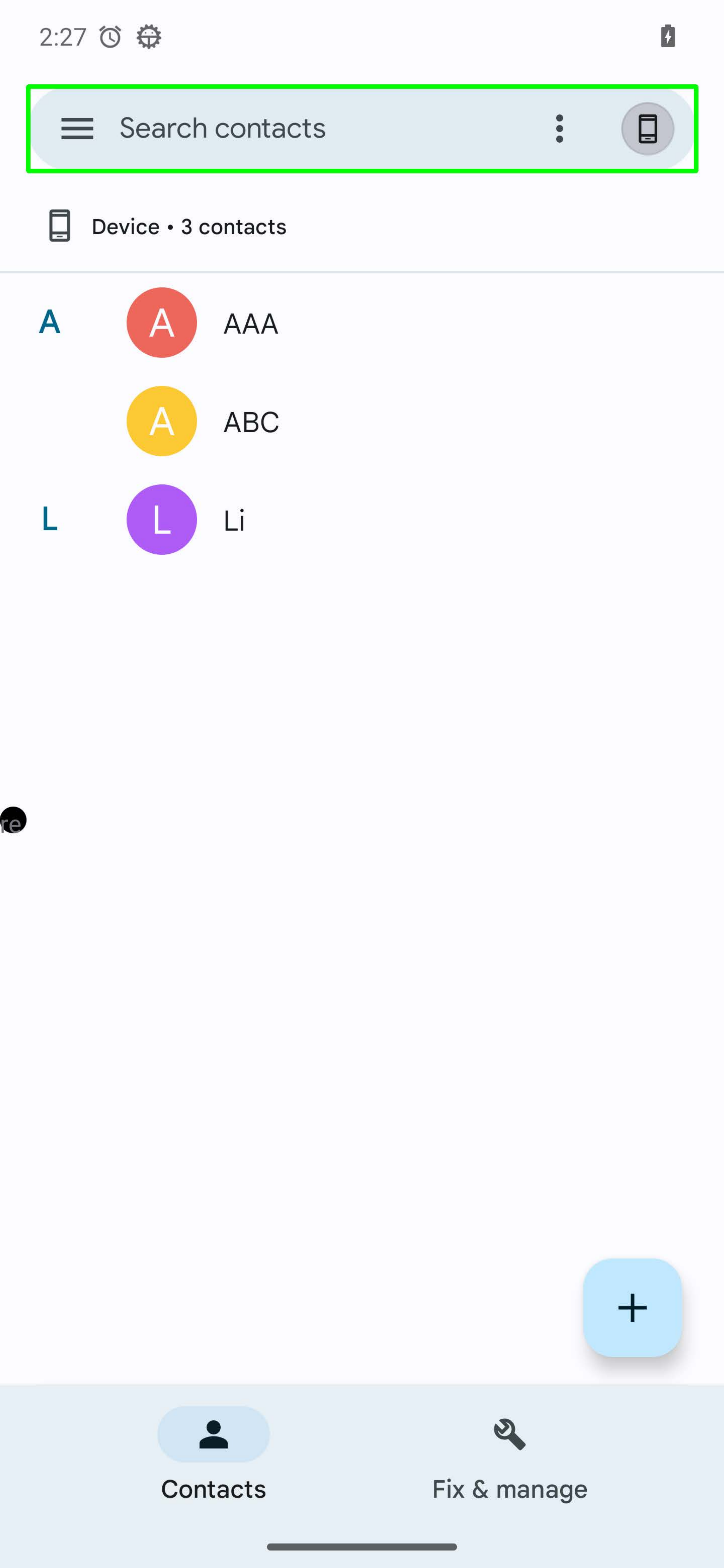}
    \caption{Semantic ambiguity}
    \label{fig:mobile_semantic_ambiguity}
  \end{subfigure}
  \hspace{0.01\textwidth}
  \begin{subfigure}[t]{0.18\textwidth}
    \centering
    \includegraphics[height=6cm]{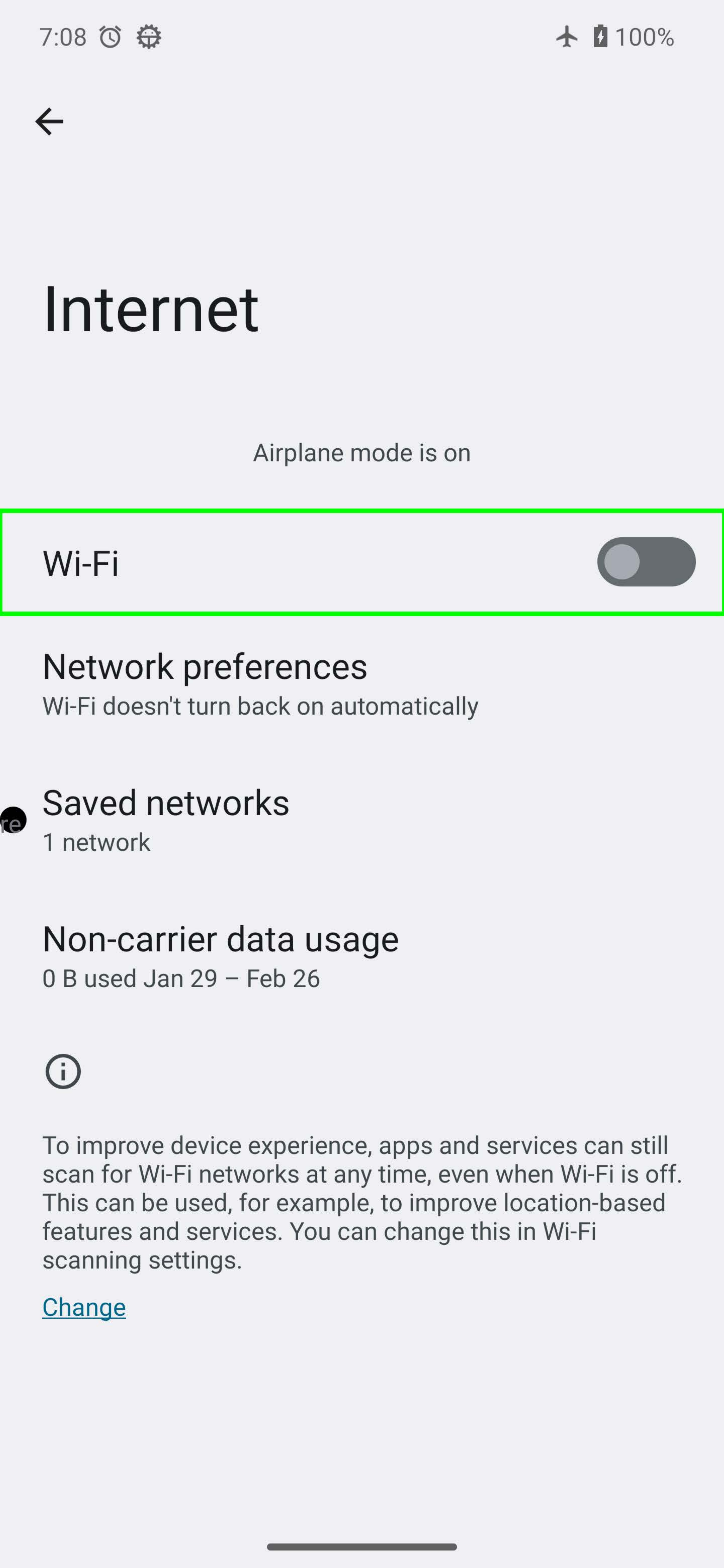}
    \caption{Effective bounding box for setting WiFi in Android}
    \label{fig:mobile_ambiguity_android}
  \end{subfigure}
  \hspace{0.01\textwidth}
  \begin{subfigure}[t]{0.18\textwidth}
    \centering
    \includegraphics[height=6cm]{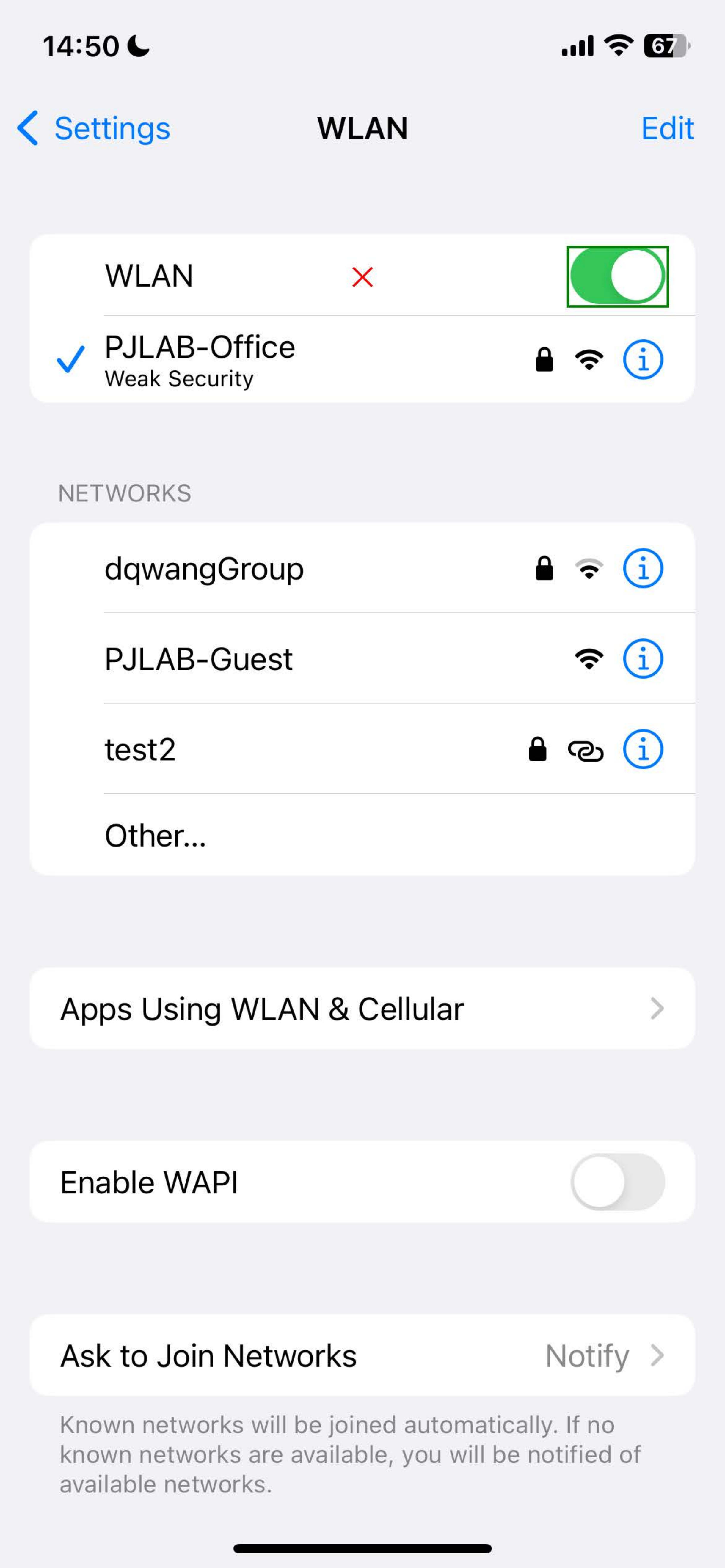}
    \caption{Effective bounding box for setting WiFi in iOS}
    \label{fig:mobile_ambiguity_ios}
  \end{subfigure}
  \vspace{-0.5em}
  \caption{Examples of potential challenges in mobile data acquisition:(a) The problem of occluded elements being indistinguishable during XML extraction.(b) The potential inaccuracy of extracted bounding boxes due to loss of element attributes.(c) The problem of semantic ambiguity caused by insufficiently detailed XML extraction.(d, e) Differences in the functionality of similar regions across different systems or apps.}
  \label{fig:mobile_all_examples}
  \vspace{-0.75em}
\end{figure*}

\textbf{Occlusion and Invisibility Correction.} 
Mobile interfaces frequently employ high-level components such as dialogs, side drawers, and floating menus. These are rendered at the topmost Z-order, so underlying nodes remain in the XML yet can no longer be clicked, producing ``ghost” targets(see Fig.\ref{fig:mobile_occlusion}). To improve visibility and hierarchy accuracy at the source, we replace the traditional \verb|adb shell uiautomator dump| with \verb|uiautomator2.dump_hierarchy()|. The latter prunes recognisably occluded nodes while generating the XML and, for pages that \texttt{adb} fails to parse, still returns a complete hierarchy—significantly increasing data coverage. Coupled with the random-walk heuristic that “prioritises newly appeared elements,” this greatly reduces mis-clicks caused by occlusion. In addition, UIAutomator2 markedly lowers the probability of XML retrieval failures, accelerating exploration efficiency.

\textbf{Attribute Completion and Correction.} 
Many commercial apps do not fully propagate accessibility traits in their custom views; a typical pattern is a parent node with \texttt{clickable=true} while all its children are \texttt{clickable=false}, leading to the issue shown in Fig.\ref{fig:mobile_missing}. Genuine clickable regions are thus ignored. We employ an “inherit-then-suppress” strategy: when a parent is clickable and every descendant is marked non-clickable, the clickable flag is inherited downward; if any descendant is already declared clickable, inheritance stops to avoid creating false hotspots. Experiments show that this method restores the vast majority of missing attributes while maintaining a low false-positive rate.

\textbf{Semantic and Functional Ambiguities.} 
Semantic ambiguity arises when an XML bounding box is too large and covers multiple sub-controls (for example, the playback button, author area and more-options button), making a single node unable to convey precise meaning. In Fig.\ref{fig:mobile_semantic_ambiguity}, the green box shows one clickable bounding-box region in the XML, but taps in different parts of that region may produce different results, creating semantic ambiguity. To address this, we prioritise leaf nodes and tighten their bounding boxes; we only retain a parent node when its centre lies outside every child’s bounds, thus preserving the overall intent of the composite control.
Functional ambiguity occurs when the same layout triggers different actions in different software or operating systems. In Fig.\ref{fig:mobile_ambiguity_android} and Fig.\ref{fig:mobile_ambiguity_ios}, for example, both the text and the icon of a switch are tappable in stock Android settings, whereas in iOS only the icon responds to taps and the rest of the region is inert. To reduce such mispredictions, whenever we detect an “icon + text” sibling pattern we give the icon a higher click priority. This approach produces more consistent cross-device behaviour during training and testing.
By systematically handling overlay occlusion, attribute omissions and both semantic and functional ambiguities, we significantly improve the reliability of mobile-side data collection and increase the success rate of downstream automation tasks.

\begin{figure*}[!h]
  \centering
  \includegraphics[width=0.8\linewidth]{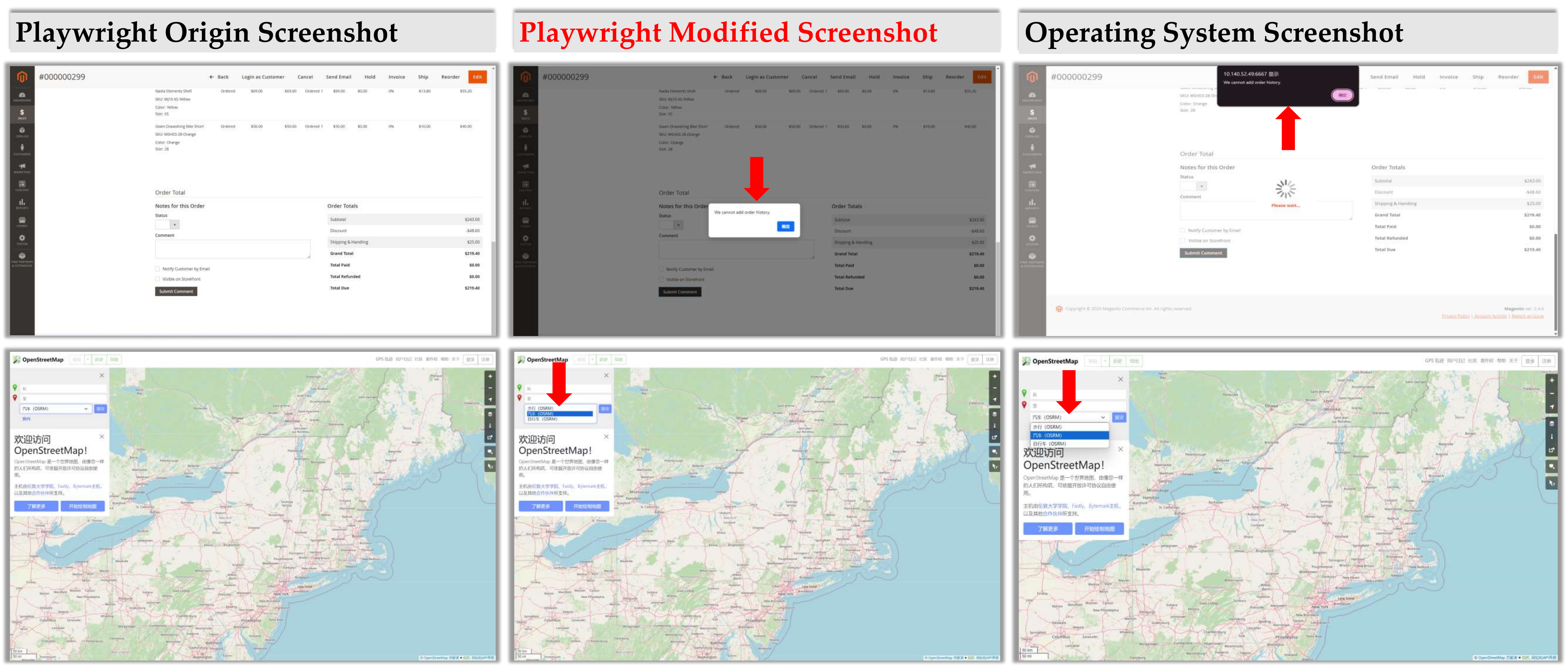}
    \vspace{-0.5em}
\caption{Examples of native browser UI limitations in automation tools.}
  \label{fig:web_playwright_modified}
  \vspace{-0.75em}
\end{figure*}

\begin{figure*}[!tbh]
  \centering
  % Fig.(a)
  \begin{minipage}{\textwidth}
    \centering
    \includegraphics[width=0.5\linewidth, keepaspectratio]{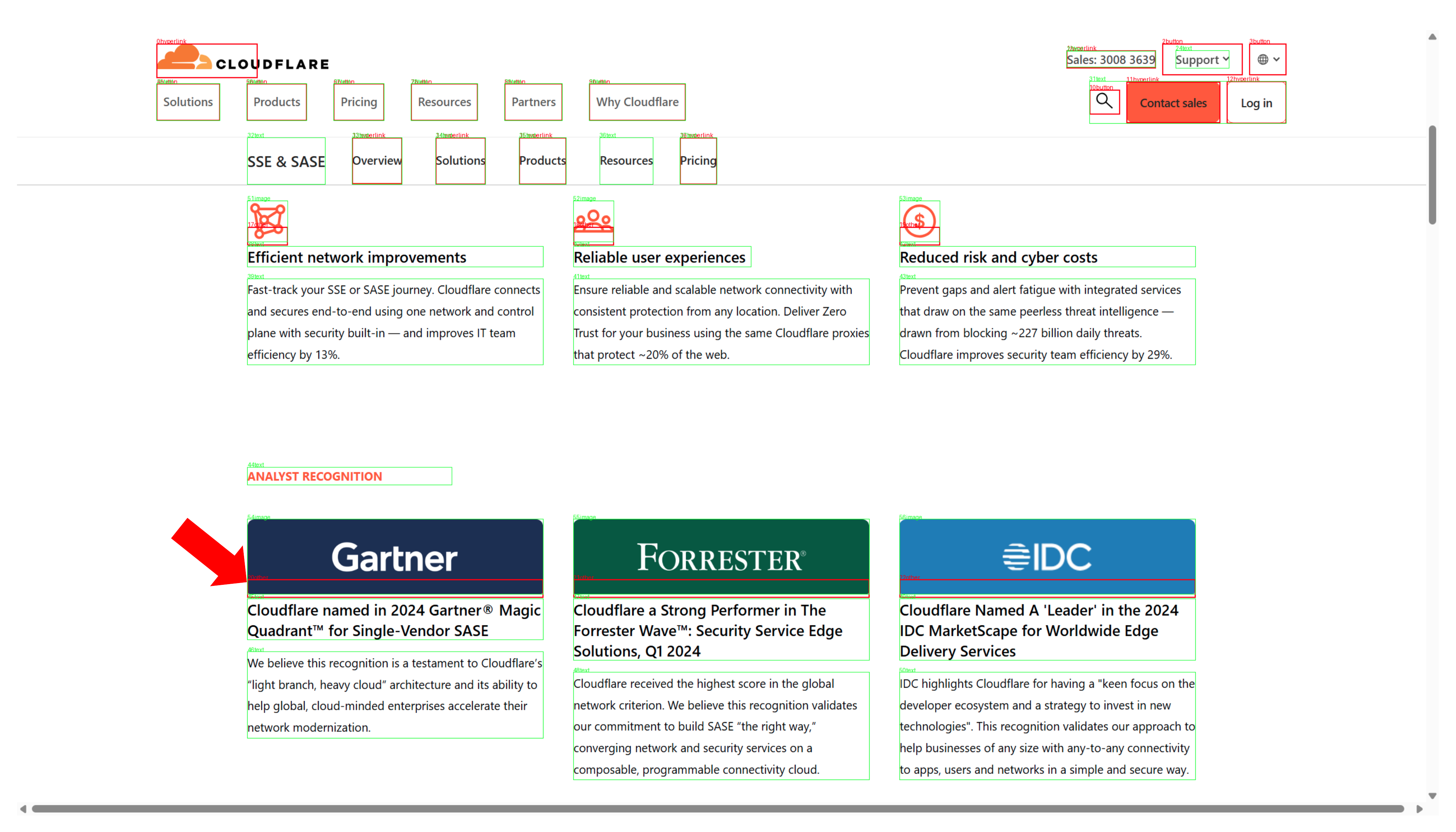}
    \vspace{0.5em}
    \subcaption[a]{Failure Case 1: The red box pointed to by the red arrow is originally an unclickable image element, but it is set as \texttt{role=button} in the HTML.}
    \label{fig:web_data_acquisition_visualization_a}
  \end{minipage}
  
  \vspace{0.5em} 
  
  % Fig.(b)
  \begin{minipage}{\textwidth}
    \centering
    \includegraphics[width=0.5\linewidth, keepaspectratio]{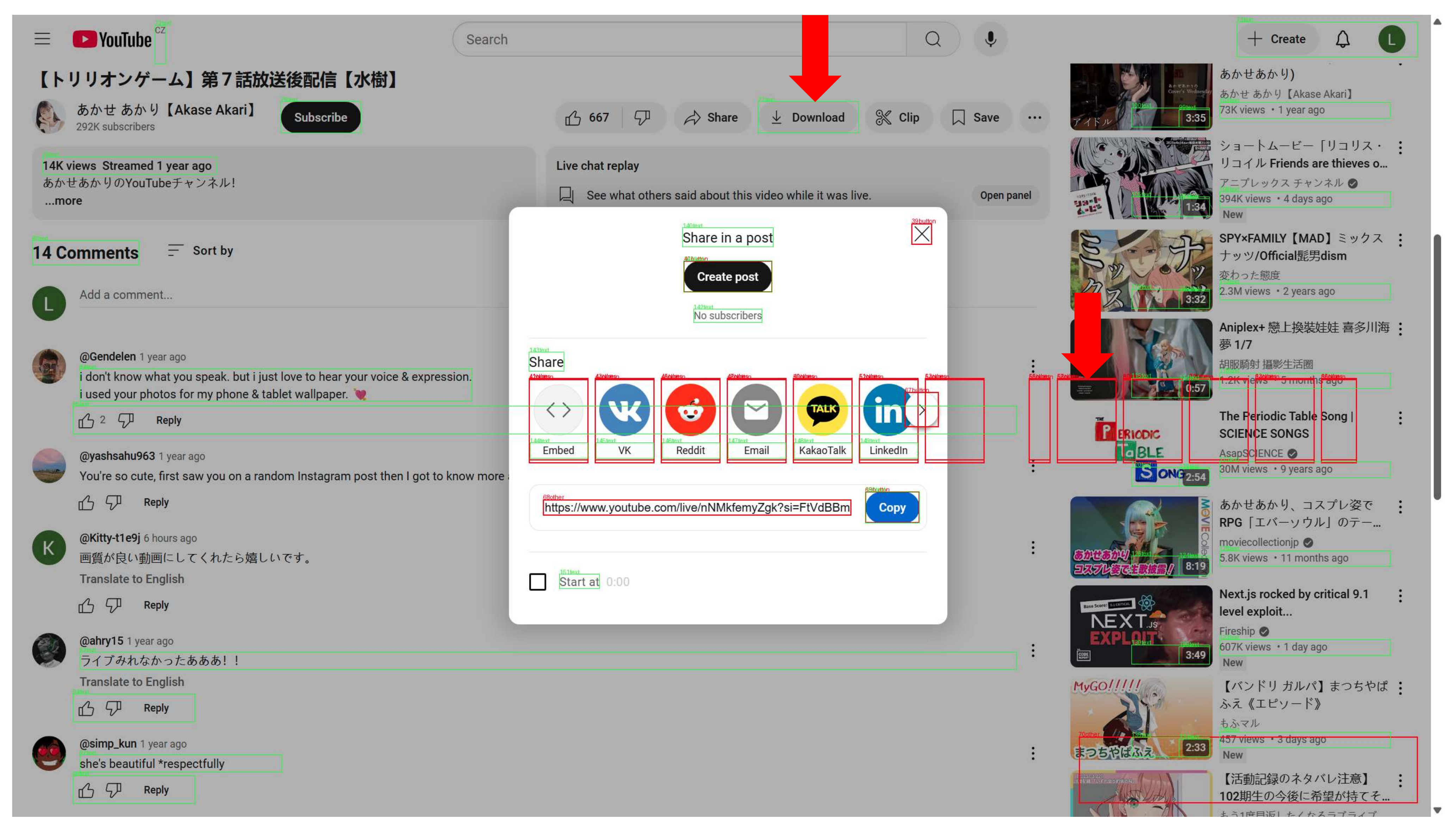}
    \vspace{0.5em}
    \subcaption[b]{Failure Case 2: As indicated by the red arrow, some non-top-level elements and invisible list elements are not filtered out by the rules.}
    \label{fig:web_data_acquisition_visualization_b}
  \end{minipage}
  
  \vspace{0.5em} 
  
  % % Fig.(c)
  \begin{minipage}{\textwidth}
    \centering
    \includegraphics[width=0.5\linewidth, keepaspectratio]{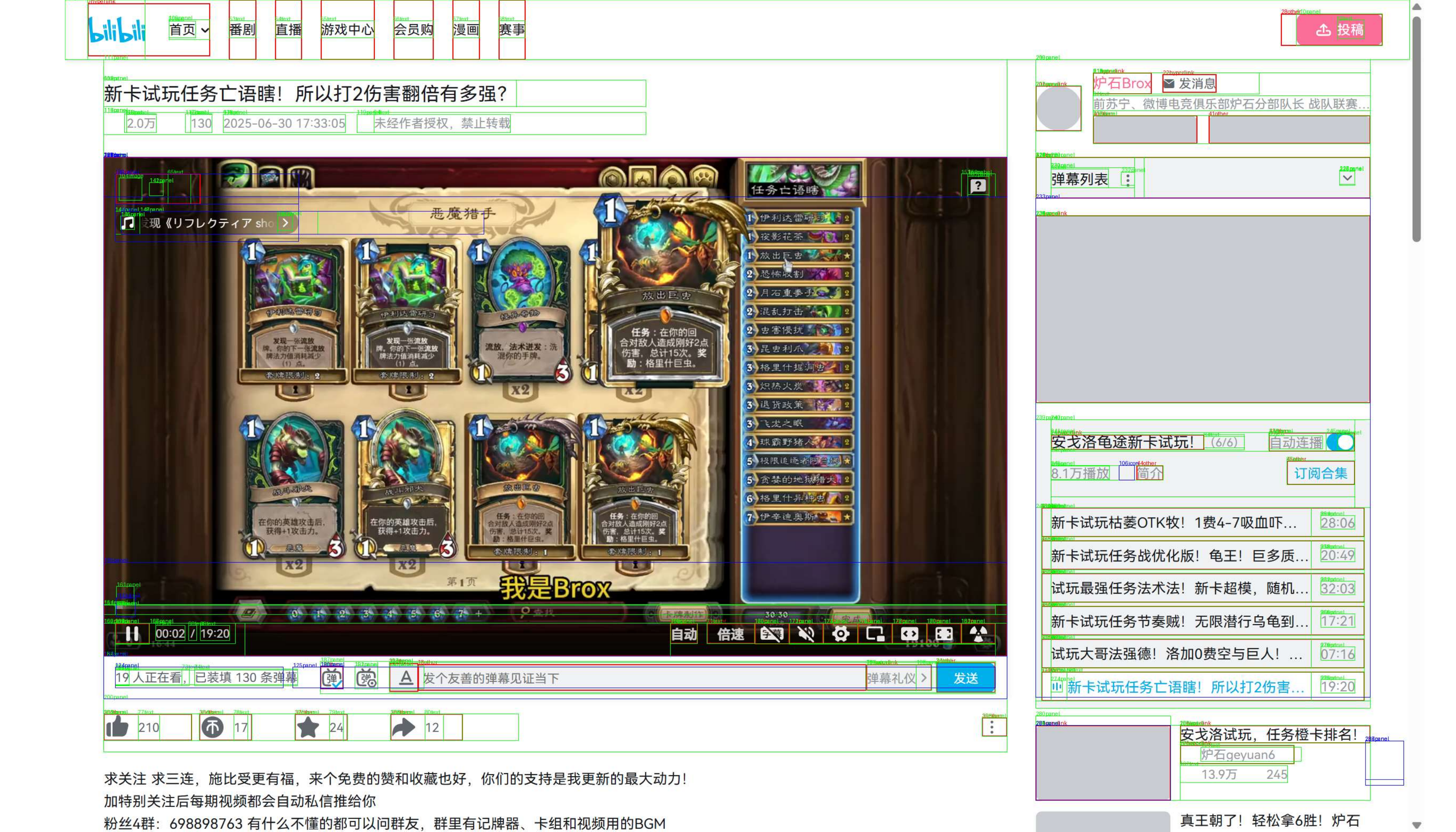}
    \vspace{0.5em}
    \subcaption[c]{Success Case 1: Reduce web page dynamics.}
    \label{fig:web_data_acquisition_visualization_c}
  \end{minipage}
  
  \vspace{0.5em} 
  
  % Fig.(d)
  \begin{minipage}{\textwidth}
    \centering
    \includegraphics[width=0.5\linewidth, keepaspectratio]{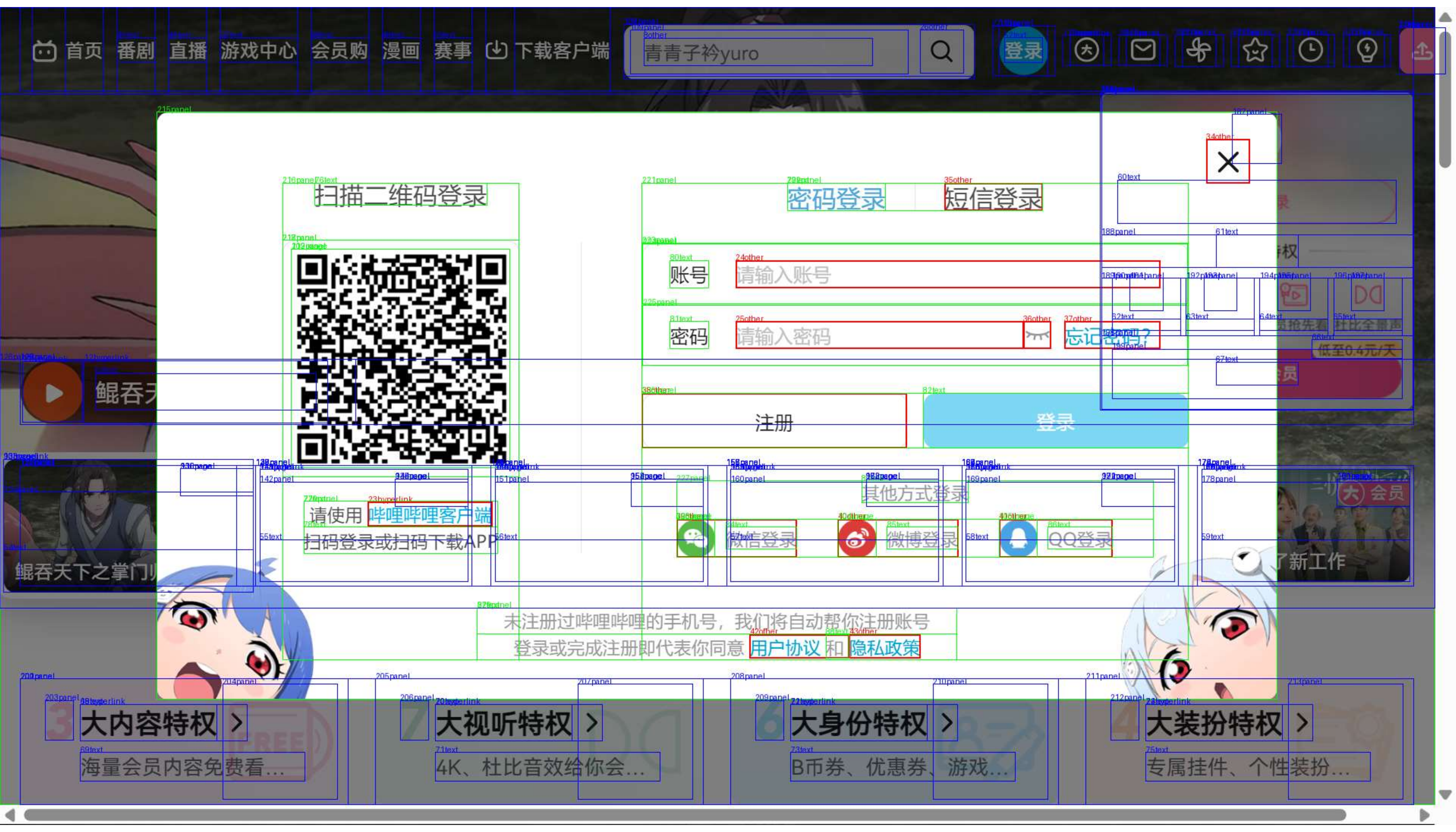}
    \vspace{0.5em}
    \subcaption[d]{Success Case 2: Correctly handle element hierarchy relationships.}
    \label{fig:web_data_acquisition_visualization_d}
  \end{minipage}
  
  \vspace{-0.5em}
  \caption{Examples of visualizations in web data acquisition. (a) shows website developer uses element identity attributes incorrectly, (b) illustrates complexity or particularity of the web leads to problems with hierarchy and visibility analysis, (c) demonstrates we alleviate the dynamic problem of web pages when playing videos, and (d) presents an example of correctly analyzing each element in a page. The red box represents clickable elements, the green box represents non-clickable elements, and the blue box represents illegal elements that have been filtered out.}
  \label{fig:web_data_acquisition_visualization}
  \vspace{-0.75em}
\end{figure*}

\subsubsection{Web}
\label{appendix:web_lessons}

\textbf{Addressing Limitations in Automation Tools.} Automation tools like Automation tools like Selenium and Playwright suffer from a critical limitation where their \texttt{page.screenshot()} function fails to capture native browser UI components rendered outside the DOM. This omission disrupts essential visual feedback for sequential decision-making in web agents. We categorize these problematic elements into two classes: predictable UI triggered by deliberate actions (e.g., context menus, tab navigation, forward/back buttons), and unpredictable UI emerging during tasks (browser dialogs and native \texttt{select} dropdowns). The inherent invisibility of predictable UI components prevents agents from developing interaction intentions for these features; while our methodological constraint limiting interactions to left-clicks effectively eliminates potential negative impacts from this omission. However, to compensate for the unavoidably reducing behavioral diversity in captured data and ensure comprehensive functional coverage, we conducted extensive web data collection in native desktop environments, enriching our training corpus with full-spectrum browser interaction examples. The unpredictable UI category proves more severe, as evidenced in Fig.~\ref{fig:web_playwright_modified} (Playwright Origin vs. OS Screenshot), where missing elements prevent task completion and impact evaluation integrity. Our \textbf{behavioral simulation solution} addresses this: for \texttt{select} elements, JavaScript modifies the \texttt{size} attribute to visually expand options within the DOM, with event listeners reverting the state; for dialogs, an interceptor captures properties, dismisses the native instance, and injects a visually identical DOM-based replica with non-functional buttons. The efficacy of this approach is demonstrated in Fig.~\ref{fig:web_playwright_modified} (Modified Screenshot), which illustrates the successful visual simulation of both UI components. While other potential related issues may exist beyond our current observations, they have not manifested in our evaluation scenarios and thus remain outside the scope of our present investigation. 

\textbf{Metadata Advantages and Parsing Challenges.} Web page content, structured through HTML and DOM trees, inherently provides rich metadata advantages over alternative platforms. JavaScript enables precise element positioning and hierarchical analysis that significantly exceeds capabilities in other contexts, enhancing metadata extraction efficiency as illustrated in Fig.~\ref{fig:web_data_acquisition_visualization_d}. However, the heterogeneity of the web ecosystem—diverse frontend frameworks, inconsistent development practices, and variable standards—prevents comprehensive coverage by data collection algorithms. Two representative challenges emerge: First, as shown in Fig.~\ref{fig:web_data_acquisition_visualization_a}, developers misapply attributes such as \texttt{role=button} to non-interactive images, introducing semantic inconsistencies that cause parsing anomalies. Second, current algorithms exhibit deficiencies in hierarchical analysis and visibility detection, resulting in inadequate filtration of underlying or invisible elements as demonstrated in Fig.~\ref{fig:web_data_acquisition_visualization_b}. Considering the substantial volume of extractable elements in web environments, we propose that maximizing the recall of valid interactive elements should be the primary objective across platforms. This position advocates for aggressive filtering strategies rather than conservative approaches that might inadvertently retain invalid elements. While this methodology may occasionally exclude some valid elements, the benefits of reducing noise in the dataset significantly outweigh the potential costs of missing a limited number of interactive elements.

It is particularly noteworthy that the technical limitations have not been explicitly addressed in the extant literature on WebAgent papers, despite their profound implications for agent functionality and evaluation methodology. We therefore advocate for increased attention to these considerations in future WebAgent research. Additionally, our analysis reveals that web environments lacking browser UI elements significantly constrain an agent's exploration capabilities in the absence of compensatory action mechanisms (\eg, returning to a previous page—a trivial operation when using a browser's back button—may require complex navigation sequences or prove entirely infeasible within the constrained visual context available). Fortunately, the refined WebArena-Lite benchmark evaluation has been specifically designed to eliminate such problematic scenarios, thereby ensuring methodological integrity and evaluation reliability. Nevertheless, based on our findings, we strongly recommend that future research prioritize the execution of web-based tasks within native desktop environments, which may necessitate the development of new benchmarks and the migration of existing benchmarks.

\textbf{Temporal Synchronization in Dynamic Page States.} 
The web platform exhibits substantial dynamism, frequently causing temporal discrepancies between page states during element parsing and screenshot capture. The non-instantaneous nature of parsing further compounds this issue by permitting mid-process element state changes. A characteristic scenario involves the auto-hiding behavior of video player control bars, illustrated in Fig.~\ref{fig:web_data_acquisition_visualization_c}. Current mitigation strategies employ dual measures: Initially awaiting complete page stabilization, followed by proactively triggering state persistence for specific elements—such as maintaining video control visibility through cursor hovering. Nevertheless, managing dynamic content remains a core challenge in web data acquisition.

\textbf{Leveraging Multi-Source Textual Semantics.} 
Web elements contain rich semantic description layers extending far beyond basic \texttt{textContent} compared to other platforms. Functional icons often convey operational semantics through \texttt{alt} and \texttt{title} attributes, while accessibility-compliant sites provide enhanced descriptions via properties like \texttt{aria-label}. Systematically aggregating these multi-source textual features establishes strong semantic associations, furnishing comprehensive contextual grounding for model annotations and effectively suppressing annotation hallucinations.

\subsection{Evaluation}
\subsubsection{Clarifications on Web Online Benchmark}
\label{appendix:web_online_evaluation}

Current web platform evaluation benchmarks can be categorized into two main types based on the website environment. The first type utilizes real websites for online evaluation, primarily derived from the offline evaluation work Mind2Web~\citep{deng2023mind2web}. Examples include Mind2Web-Live~\citep{pan2024webcanvas}, Online-Mind2Web~\citep{xue2025illusion}, and WebVoyager~\citep{he2024webvoyager}, with UI-TARS~\citep{qin2025uitars} employing WebVoyager and Online-Mind2Web for web domain evaluation. The second type conducts evaluations on locally deployed websites, pioneered by WebArena~\citep{zhou2023webarena}, which leverages open-source website code and databases~\citep{sun2024code} to provide highly simulated and interactive local Docker deployment environments for five functionally diverse websites, including GitLab, map services, forums, online shopping, and content management platforms (CMS). WebArena has constructed over 800 web tasks, inspiring derivative evaluation frameworks such as VisualAgentBench~(WebArena-Lite)~\citep{liu2024visualagentbench}
and VisualWebArena~\citep{koh2024visualwebarena}. 
Furthermore, the evaluation protocols can be classified into two categories: rule-based evaluation exemplified by WebArena~\citep{zhou2023webarena} and VLM-as-a-Judge evaluation, such as Online-Mind2Web~\citep{xue2025illusion}.

\textbf{Rationale for Selecting Local Website Environments.}
We deliberately abandoned evaluation benchmarks based on real websites for several compelling reasons. The primary concern is the temporal instability of online environments—tasks that are currently feasible may become impossible due to website updates, domain changes, or site closures. Despite efforts by frameworks like Mind2Web-Live to maintain and update tasks periodically, such updates inevitably compromise evaluation fairness. Additionally, as noted in \citep{xu2024aguvis}, automated tools frequently encounter anti-automation barriers such as reCAPTCHA verification. Moreover, since most target websites are hosted in the United States, researchers in non-US regions (particularly China) face persistent connectivity issues and access restrictions even with VPN services—different VPN providers often yield inconsistent access results. These factors significantly undermine fair model comparison and hinder the extraction of valuable insights from evaluation results.

\textbf{WebArena-Lite-v2.}
Consequently, we focused on the WebArena series, whose locally deployed website environments offer substantial stability and internal accessibility, enabling flexible task construction and evaluation design. Considering that WebArena often includes three or more iterations of the same task template, resulting in repetitive and time-consuming evaluations, we selected the WebArena-Lite subset, which provides 165 high-quality refined tasks. However, our empirical evaluation and manual inspection revealed persistent issues. Therefore, we further refined the benchmark to create \textbf{WebArena-Lite-v2}, comprising 154 tasks optimized for both headed browser environments and headless automation tool environments. Recent developments, such as OpenAI's Operator, demonstrate a transition from headless environments provided by automation tools toward headed desktop browser environments for web agent evaluation. As detailed in ~\ref{appendix:web_lessons}, both environments present distinct advantages and limitations. To facilitate comprehensive ablation studies on these different operational modes, WebArena-Lite-v2 ensures that all tasks can be solved through at least one viable path using desktop action spaces (without specialized web actions like go\_forward, go\_backward, open\_url, or tab\_switch) in both headed and headless environments. Furthermore, all tasks are designed to provide sufficient visual information guidance, eliminating the necessity for DOM information and thus making the benchmark suitable for pure vision-based evaluation (while remaining compatible with SoM or DOM-enhanced assessment). Finally, we implemented comprehensive yet flexible evaluation criteria--comprehensive in accommodating multiple possible solutions through the \texttt{\textbar OR\textbar} operator where satisfying any one solution is sufficient and flexible in employing LLM-based \texttt{fuzzy\_match} for semantic similarity assessment in tasks involving question answering or content completion.

% \textbf{From WebArena-Lite to WebArena-Lite-v2.}
\textbf{Discussions between WebArena-Lite and WebArena-Lite-v2.}
Our refinements encompass both environmental and task improvements. For the evaluation environment, we implemented two significant enhancements. First, we addressed the OpenStreetMap website's limitations, where the official Docker environment lacked local database storage for node information, rendering tasks like ``What is the phone number of Western Pennsylvania Hospital'' impossible to complete. We resolved this by importing Pennsylvania state PBF data, enabling the completion of such tasks. Second, we developed consistent solutions for headless automation environments to overcome the observation challenges with select option dropdowns and dialog windows, as illustrated in ~\ref{appendix:web_lessons} with Fig.~\ref{fig:web_playwright_modified}. Regarding task refinement, we eliminated 11 tasks requiring multi-tab interactions, resulting in a curated set of 154 tasks. We conducted a comprehensive revision of instructions and evaluation functions for all remaining tasks. The instruction refinements encompassed semantic clarification, typographical correction, and minimal reconstruction of impracticable directives (e.g., the instruction ``Re-post the image of the costume contest in this page to the funny subreddit and note ``from /f/pics''' proved infeasible since headless environments lack image URL extraction capabilities). Our evaluation function enhancements incorporated supplementary valid solutions (e.g., for the query ``What is the zip code of Chatham University?'', we augmented the answer from exclusively ``15232'' to ``15232 \texttt{\textbar OR\textbar} 15208'' after identifying multiple Chatham University locations through OpenStreetMap queries) and accommodated semantically equivalent solution expressions (e.g., for ``Show me products under \$100 in `Men Shoes' category'', we recognized both \texttt{\_\_SHOPPING\_\_/clothing-shoes-jewelry/men/shoes.html?price=0-100} and \texttt{\_\_SHOPPING\_\_/clothing-shoes-jewelry.html?cat=145\&price=0-100} as valid pathways to identical content pages). This methodological approach ensures comprehensive answer validation. Additionally, acknowledging language models' inherent variability in textual response generation, we systematically replaced all \texttt{exact\_match} evaluation criteria within the \texttt{string\_match} classification with more nuanced \texttt{must\_include}, \texttt{must\_exclude}, and \texttt{fuzzy\_match} parameters, thereby significantly enhancing evaluation robustness and interpretative flexibility. However, WebArenaLite-v2 still employs static evaluation methodologies for certain tasks (such as when identifying user's most recent order, where the Ground Truth is predetermined as a specific order number or webpage). Although executing evaluations within a local environment has mitigated the impact of this limitation, a critical future direction involves developing evaluation protocols that are both dynamic and precise. This advancement necessitates addressing the challenge of extracting Ground Truth information from web pages that may not have been accessed by the agent during its navigation trajectory. This capability is essential for comprehensive evaluation of agent performance across diverse web interaction scenarios.

\subsection{Prompt Engineering}
\label{sec:prompt_template}
To facilitate reproducibility and offer practical guidance for future research, we include all prompt templates utilized throughout our work in this section. These prompts cover a wide range of use cases, including data filtering, annotation, and the prompts used in our \modelname. Specifically, we detail the instructions employed for GUI understanding, grounding supervision, and trajectory annotation, as well as those used to elicit reasoning traces and alternative actions. Each prompt is carefully crafted to align with the capabilities of large vision-language models such as GPT-4o and Claude-3.7, ensuring high-quality outputs for downstream training. By releasing these prompts, we aim to enhance transparency and support the development of more robust and interpretable computer use agents.

\subsubsection{Prompts for Our Agent}
To ensure generalizable and controllable agent behavior, we design a structured system prompt template for \modelname that explicitly encodes the available action space. This template serves as the foundational context for all three inference paradigms—Grounding Mode, Direct-Action Mode, and Reasoned-Action Mode—guiding the model to produce spatially grounded and semantically aligned outputs. The system prompt defines the operational semantics of each action type, including spatial commands such as \texttt{click(x, y)}, \texttt{dragTo(x, y)}, and \texttt{write(text)}, as well as higher-level control tokens like \texttt{terminate} and \texttt{wait}.

We envision the system prompt as a modular and extensible interface. In future iterations, we aim to decouple the action space definition from the core prompt logic, allowing for a plug-and-play architecture that can dynamically adapt to the interaction paradigms of diverse computing platforms. This modularity would enable seamless integration of device-specific actions, such as \texttt{swipe} for mobile interfaces or \texttt{hotkey} for desktop environments, while preserving consistency in agent behavior. Our design lays the foundation for building a unified prompting framework that can scale to arbitrary GUI-based control systems.
\begin{tcolorbox}[title=System Prompt Template For Action Grounding Mode]
\scriptsize
\begin{Verbatim}[breaklines=true, breaksymbol={}]
You are an autonomous GUI agent capable of operating on desktops, mobile devices, and web browsers. Your primary function is to analyze screen captures and perform appropriate UI actions to complete assigned tasks.

## Action Space
def click(
    x: float | None = None,
    y: float | None = None,
    clicks: int = 1,
    button: str = "left",
) -> None:
    """Clicks on the screen at the specified coordinates. The `x` and `y` parameter specify where the mouse event occurs. If not provided, the current mouse position is used. The `clicks` parameter specifies how many times to click, and the `button` parameter specifies which mouse button to use ('left', 'right', or 'middle')."""
    pass


def doubleClick(
    x: float | None = None,
    y: float | None = None,
    button: str = "left",
) -> None:
    """Performs a double click. This is a wrapper function for click(x, y, 2, 'left')."""
    pass


def rightClick(x: float | None = None, y: float | None = None) -> None:
    """Performs a right mouse button click. This is a wrapper function for click(x, y, 1, 'right')."""
    pass


def moveTo(x: float, y: float) -> None:
    """Move the mouse to the specified coordinates."""
    pass


def dragTo(
    x: float | None = None, y: float | None = None, button: str = "left"
) -> None:
    """Performs a drag-to action with optional `x` and `y` coordinates and button."""
    pass


def swipe(
    from_coord: tuple[float, float] | None = None,
    to_coord: tuple[float, float] | None = None,
    direction: str = "up",
    amount: float = 0.5,
) -> None:
    """Performs a swipe action on the screen. The `from_coord` and `to_coord` specify the starting and ending coordinates of the swipe. If `to_coord` is not provided, the `direction` and `amount` parameters are used to determine the swipe direction and distance. The `direction` can be 'up', 'down', 'left', or 'right', and the `amount` specifies how far to swipe relative to the screen size (0 to 1)."""
    pass


def long_press(x: float, y: float, duration: int = 1) -> None:
    """Long press on the screen at the specified coordinates. The `duration` specifies how long to hold the press in seconds."""
    pass


## Input Specification
- Screenshot of the current screen + task description

## Output Format
<action>
[A set of executable action command]
</action>

## Note
- Avoid action(s) that would lead to invalid states.
- The generated action(s) must exist within the defined action space.
- The generated action(s) should be enclosed within <action></action> tags.
\end{Verbatim}
% \end{tcblisting}
\end{tcolorbox}

\begin{tcolorbox}[title=System Prompt Template For Direct Action Mode]
\scriptsize
\begin{Verbatim}[breaklines=true, breaksymbol={}]
You are an autonomous GUI agent operating on the **{PLATFORM}** platform(s). Your primary function is to analyze screen captures and perform appropriate UI actions to complete assigned tasks.

## Action Space
def click(
    x: float | None = None,
    y: float | None = None,
    clicks: int = 1,
    button: str = "left",
) -> None:
    """Clicks on the screen at the specified coordinates. The `x` and `y` parameter specify where the mouse event occurs. If not provided, the current mouse position is used. The `clicks` parameter specifies how many times to click, and the `button` parameter specifies which mouse button to use ('left', 'right', or 'middle')."""
    pass


def doubleClick(
    x: float | None = None,
    y: float | None = None,
    button: str = "left",
) -> None:
    """Performs a double click. This is a wrapper function for click(x, y, 2, 'left')."""
    pass


def rightClick(x: float | None = None, y: float | None = None) -> None:
    """Performs a right mouse button click. This is a wrapper function for click(x, y, 1, 'right')."""
    pass


def scroll(clicks: int, x: float | None = None, y: float | None = None) -> None:
    """Performs a scroll of the mouse scroll wheel at the specified coordinates. The `clicks` specifies how many clicks to scroll. The direction of the scroll (vertical or horizontal) depends on the underlying operating system. Normally, positive values scroll up, and negative values scroll down."""
    pass


def moveTo(x: float, y: float) -> None:
    """Move the mouse to the specified coordinates."""
    pass


def dragTo(
    x: float | None = None, y: float | None = None, button: str = "left"
) -> None:
    """Performs a drag-to action with optional `x` and `y` coordinates and button."""
    pass


def press(keys: str | list[str], presses: int = 1) -> None:
    """Performs a keyboard key press down, followed by a release. The function supports pressing a single key or a list of keys, multiple presses, and customizable intervals between presses."""
    pass


def hotkey(*args: str) -> None:
    """Performs key down presses on the arguments passed in order, then performs key releases in reverse order. This is used to simulate keyboard shortcuts (e.g., 'Ctrl-Shift-C')."""
    pass


def keyDown(key: str) -> None:
    """Performs a keyboard key press without the release. This will put that key in a held down state."""
    pass


def keyUp(key: str) -> None:
    """Performs a keyboard key release (without the press down beforehand)."""
    pass


def write(message: str) -> None:
    """Write the specified text."""
    pass


def call_user() -> None:
    """Call the user."""
    pass


def wait(seconds: int = 3) -> None:
    """Wait for the change to happen."""
    pass


def response(answer: str) -> None:
    """Answer a question or provide a response to an user query."""
    pass


def terminate(status: str = "success", info: str | None = None) -> None:
    """Terminate the current task with a status. The `status` specifies the termination status ('success', 'failure'), and the `info` can provide additional information about the termination."""
    pass


## Input Specification
- Screenshot of the current screen + task description + your past interaction history with UI to finish assigned tasks.

## Output Format
<operation>
[Next intended operation description]
</operation>
<action>
[A set of executable action commands]
</action>

## Note
- Avoid action(s) that would lead to invalid states.
- The generated action(s) must exist within the defined action space.
- The generated operation and action(s) should be enclosed within <operation></operation> and <action></action> tags, respectively.
\end{Verbatim}
\end{tcolorbox}

\begin{tcolorbox}[title=System Prompt Template For Reasoned-Action Mode]
\scriptsize
\begin{Verbatim}[breaklines=true, breaksymbol={}, breakanywhere=true]
You are an autonomous GUI agent operating on the **{PLATFORM}** platform. Your primary function is to analyze screen captures and perform appropriate UI actions to complete assigned tasks.

## Action Space
def click(
    x: float | None = None,
    y: float | None = None,
    clicks: int = 1,
    button: str = "left",
) -> None:
    """Clicks on the screen at the specified coordinates. The `x` and `y` parameter specify where the mouse event occurs. If not provided, the current mouse position is used. The `clicks` parameter specifies how many times to click, and the `button` parameter specifies which mouse button to use ('left', 'right', or 'middle')."""
    pass


def doubleClick(
    x: float | None = None,
    y: float | None = None,
    button: str = "left",
) -> None:
    """Performs a double click. This is a wrapper function for click(x, y, 2, 'left')."""
    pass


def rightClick(x: float | None = None, y: float | None = None) -> None:
    """Performs a right mouse button click. This is a wrapper function for click(x, y, 1, 'right')."""
    pass


def scroll(clicks: int, x: float | None = None, y: float | None = None) -> None:
    """Performs a scroll of the mouse scroll wheel at the specified coordinates. The `clicks` specifies how many clicks to scroll. The direction of the scroll (vertical or horizontal) depends on the underlying operating system. Normally, positive values scroll up, and negative values scroll down."""
    pass


def moveTo(x: float, y: float) -> None:
    """Move the mouse to the specified coordinates."""
    pass


def dragTo(
    x: float | None = None, y: float | None = None, button: str = "left"
) -> None:
    """Performs a drag-to action with optional `x` and `y` coordinates and button."""
    pass


def press(keys: str | list[str], presses: int = 1) -> None:
    """Performs a keyboard key press down, followed by a release. The function supports pressing a single key or a list of keys, multiple presses, and customizable intervals between presses."""
    pass


def hotkey(*args: str) -> None:
    """Performs key down presses on the arguments passed in order, then performs key releases in reverse order. This is used to simulate keyboard shortcuts (e.g., 'Ctrl-Shift-C')."""
    pass


def keyDown(key: str) -> None:
    """Performs a keyboard key press without the release. This will put that key in a held down state."""
    pass


def keyUp(key: str) -> None:
    """Performs a keyboard key release (without the press down beforehand)."""
    pass


def write(message: str) -> None:
    """Write the specified text."""
    pass


def call_user() -> None:
    """Call the user."""
    pass


def wait(seconds: int = 3) -> None:
    """Wait for the change to happen."""
    pass


def response(answer: str) -> None:
    """Answer a question or provide a response to an user query."""
    pass


def terminate(status: str = "success", info: str | None = None) -> None:
    """Terminate the current task with a status. The `status` specifies the termination status ('success', 'failure'), and the `info` can provide additional information about the termination."""
    pass


## Input Specification
- Screenshot of the current screen + task description + your past interaction history with UI to finish assigned tasks.

## Output Format
```
<think>
[Your reasoning process here]
</think>
<operation>
[Next intended operation description]
</operation>
<action>
[A set of executable action command]
</action>
```

## Note
- Avoid actions that would lead to invalid states.
- The generated action(s) must exist within the defined action space.
- The reasoning process, operation and action(s) in your response should be enclosed within <think></think>, <operation></operation> and <action></action> tags, respectively
\end{Verbatim}
\end{tcolorbox}

\begin{tcolorbox}[title=User Prompt Template For Direct-Action Mode and Reasoned-Action Mode]
\scriptsize
\begin{Verbatim}[breaklines=true, breaksymbol={}]
Please generate the next move according to the UI screenshot, the task and previous operations.

Task:
{instruction}

Previous operations:
{history}
...
\end{Verbatim}
\end{tcolorbox}

\subsubsection{Prompts for Annotations}
\label{sec:prompt_template_anno}
To support reproducibility and transparency, we release all annotation-related prompts used in our data processing pipeline. These prompts cover a wide range of tasks, including trajectory filtering, GUI understanding, grounding supervision and chain-of-thought generation for goal-directed demonstrations. Each prompt is carefully designed to elicit accurate and semantically consistent annotations from large vision-language models such as GPT-4o and Claude-3.7.

Empirically, our prompts have demonstrated strong effectiveness in producing high-quality labels, which in turn significantly benefit the training of general-purpose computer use agents. By sharing these templates, we aim to standardize annotation practices in this emerging domain and foster broader progress in building scalable and open computer use systems. We hope this contributes to lowering the barrier for future research and accelerating the development of robust, multimodal GUI agents.
\begin{tcolorbox}[title={Prompt For Element Appearance{,} Layout and Functionality}]
\scriptsize
\begin{Verbatim}[breaklines=true, breaksymbol={}]
You are a GUI analysis agent, and you are currently working with a {os_name} device. You will be provided with the following resources:
1. The first image is a original screenshot from an {application}.
2. The second image is marked to highlight the selected element. 
3. The A11Tree attributes of the selected element: {element_a11tree}.
                           
Your task is to generate detailed descriptions of this marked element from appearance and position. Each description must uniquely identify the element and adhere to the following structure:

{
  "appearance": "A detailed visual description of the element, including its shape, color, size, text content (if any), and any distinguishing features.",
  "position": "A clear description of the element's location on the screen, including its relative position to nearby elements (e.g., 'below the search bar', 'to the right of the logo'), its order in a sequence (e.g., 'third button in the top navigation bar'), and its general area (e.g., 'top-left corner of the window'). Avoid using direct coordinates or the red indicator.",
}

## Guidelines for Generating Descriptions:
1. **Appearance**:
   - Focus on visual characteristics that uniquely identify the element.
   - Include details such as color, shape, size, text content (if applicable), icons, borders, shadows, or patterns.
   - If the element contains text, describe the font style, size, and content briefly.
   - Please avoid using {marker} as part of your description. Because we draw {marker} for reference and they does not exist in the original screenshot.

2. **Position**:
   - Describe the element's location relative to other prominent elements in the UI that uniquely identify the element.
   - Specify its general area (e.g., 'top-right corner', 'center of the screen') and its order in a group (e.g., 'second icon in the toolbar').
   - Please avoid using {marker} as part of your description. Because we draw {marker} for reference and they does not exist in the original screenshot.
   - Avoid vague terms like 'near' or 'close to'. Instead, use precise language such as 'directly below', 'aligned with', or 'to the left of'.


## Example Output:
{
  "appearance": "A circular icon with a white background and a magnifying glass symbol in black, surrounded by a thin gray border.",
  "position": "Located in the top-right corner of the application window, directly to the right of the profile avatar icon.",
}

## Important Notes:
- Do not copy or paraphrase the content of the A11Tree attributes directly.
- Please avoid using {marker} as part of your description. Because we draw {marker} for reference and they does not exist in the original screenshot.
- Ensure each description is detailed enough to uniquely identify the element without ambiguity.

RETURN THE DICTIONARY IN STRICT JSON FORMAT:
\end{Verbatim}
\end{tcolorbox}

\begin{tcolorbox}[title=Prompt For Screen Transition Captioning and User Intention Prediction]
\scriptsize
\begin{Verbatim}[breaklines=true, breaksymbol={}]
You are a GUI agent currently operating on a {os_name} device. You will be provided with:
1. The first image is a screenshot from an {application}, which are marked with {marker} to highlight the selected element.
2. The second image is the results of the operation {action} executed on the selected element.
3. The third image is a sub-image, which is cropped from the screenshot around the selected element and is marked with {marker}.
4. The A11Tree attributes of the selected element: {element_a11tree}.
                           
Your task is to analyze these two consecutive screenshots and complete the following tasks:
1. **State Transition Explanation**: Describe the state change caused by the operation. This should include a detailed description of the first screenshot, the action performed on the element, the differences observed in the second screenshot compared to the first, and an explanation of the most likely user action that occurred between the two frames.
2. **User Intention Inference**: Based on the action performed and the differences between the two screenshots, infer the user's intent. Explain what the user likely aimed to achieve and how the action led to the observed changes in the GUI.

Your response should be formatted as follows:
{
"state-transition": "...",
"user-intention": "...",
}

## Example Output:
{
"state-transition": "In the first screenshot, the main dashboard of the Bluecoins app is displayed with a calendar showing February 2025, and the date '3' is highlighted. After tapping on the '3', the second screenshot navigates the app to a detailed calendar view for February 2025, showing tabs like 'CATEGORIES,' 'ACCOUNTS,' 'TRANSACTIONS,' and 'REMINDERS,' with no transactions listed.",
"user-intention": "The user likely wanted to view detailed transactions and account categories for the selected date.",
}

## Important Notes:
- Avoid directly copying the A11Tree attributes of the element when writing instructions.
- Ensure the instructions are clear, unambiguous, and concise, preferably described in a single sentence.
- Do not reference the distinctive red indicator when describing UI elements.

RETURN THE DICTIONARY IN STRICT JSON FORMAT:
\end{Verbatim}
\end{tcolorbox}

\begin{tcolorbox}[title=Prompt For Interface Captioning]
\scriptsize
\begin{Verbatim}[breaklines=true, breaksymbol={}]
You are a GUI analysis agent currently working with a {os_name} device. You will receive a full screenshot of an {application}. Your objective is to produce comprehensive descriptions of the screenshot's contents and functionality. These descriptions should thoroughly explain each visible element by covering its visual attributes, spatial arrangement, and purpose within the interface.

## Key Requirements for Descriptions:
- Contextual Details: Explain the interface's overall structure and the spatial relationships between elements.
- Visual Characteristics: colors, shapes, icons, text labels, and other distinguishing visual properties.
- User Interaction: Specify how users can interact with each element and the expected results of those interactions.
- Functional Purpose: Clarify the screenshot's role within the broader application workflow.

## Important Notes:
- Synthesize the attribute information to create natural, user-friendly descriptions.
- Maintain conciseness while ensuring the descriptions are sufficiently detailed to convey the GUI's structure and operation.

PLEASE GENERATE CAPTION:
\end{Verbatim}
\end{tcolorbox}

\begin{tcolorbox}[title=Prompt For LLM-assist Filter]
\scriptsize
\begin{Verbatim}[breaklines=true, breaksymbol={}]
You are a GUI analysis agent tasked with evaluating a user interface on a {os_name} device. You will be provided with the following resources:
1. The first image is a full screenshot of an {application}, where the area of interest is highlighted with {marker}.
2. The second image is a sub-image, which is cropped from the screenshot around the selected element and is marked with {marker}.
                
Your objective is to determine whether the marked area resides in the topmost layout and can be directly clicked. Your response must be returned in JSON format, adhering to the structure below:
```json
{"answer": "No"}
```
The value of `"answer"` can only be one of the following:
- `"Yes"`: Indicates that the marked area is in the topmost view and contains a clickable or valid element that is the focus element of the current interface.
- `"No"`: Indicates that the marked area is obstructed, intercepted, non-interactive, or otherwise non-clickable due to errors, loading issues, or the absence of a valid interactive element, or the marked area is not the focus element of the current interface.
                
Here are some conditions that make an area non-clickable:
- The marked area resides in the background and is not the focus element of the current interface.
- The image displays an error or fails to load content properly.
- The marked area corresponds to an empty or blank region with no visible or interactive elements.
- The marked area contains anomalies such as overlapping elements, misplaced components, or other irregularities that hinder proper interaction.
- The marked area located in background and not the focus element of the current interface.

RETURN THE DICTIONARY IN STRICT JSON FORMAT:
\end{Verbatim}
\end{tcolorbox}

\begin{tcolorbox}[title=Prompt For High-Level Objective in Weak-Semantic Trajectories]
\scriptsize
\begin{Verbatim}[breaklines=true, breaksymbol={}]
You are an expert in designing and analyzing GUI navigation tasks. specializing in evaluating a user’s interaction trajectory within an {application} on a {os_name} device to deduce their overarching navigation goal.

You will be given the following information:
1. **Initial State Image**: A visual representation of the starting point of the interaction shown in the first image.  
2. **Final State Image**: A visual representation of the endpoint of the interaction shown in the second image.  
3. **Interaction Trajectory**: A detailed log of each step taken by the user, including the intent behind each action:  
{history}  

Your task is to craft a concise summary (1-2 sentences) that describes the navigation journey by focusing on the goal and outcome. 
1. **Identifies the user’s core objective**:  
   - Emphasize the transition from the initial state to the final state (implicitly or explicitly).  
   - Focus on the user's overall intent as inferred from the interaction history and the final state, avoiding overly detailed descriptions of operational steps (e.g., describe the task as "updating preferences" rather than "toggle the switch").  
2. **Highlights the functionality of the final state**:  
   - Briefly describe the primary function of the final state, focusing on what the user can accomplish or access as a result of completing the navigation task.  

For example:
- The phone is displaying Amap's app info page. My goal is to access the "My Guide" section on Amap's homepage from here. 
- To view Amap's notification permission, I want to move from Amap's homepage to system settings page for Amap.
- Starting from Amap's battery usage settings, I need to reach the "Offline Maps" section in the app's main interface.
- With the aim of saving posts in Instagram, please advance from the home screen to "Saved Posts" tab from Instagram's homepage.
- The screenshot shows the Chrome app info page. I want to go from here to the "History" section in Chrome's main menu.  

Now, based on the provided input, assuming you are the user, please generate an instruction of the operational navigation goal by using the first-person present tense or imperative sentence:
\end{Verbatim}
\end{tcolorbox}

\begin{tcolorbox}[title=Prompt For Low-Level Instructions in All Trajectories]
\scriptsize
\begin{Verbatim}[breaklines=true, breaksymbol={}]
You are a GUI agent currently operating on a {os_name} device. Your task is to generate a concise and clear operational instruction for interacting with the selected UI element. These instructions should be relevant to the operation and include operated details such as UI appearance, text content, position, order, file names, or other relevant content visible in the screenshots. Instructions can involve the appearance, position, or functional description of the selected element, but it must ensure that the generated instruction uniquely corresponds to the selected element.

You will be provided with:
1. The first image is a original screenshot from an {application}, which are manually marked to highlight the selected element.
2. The second image is the results of the operations ```{action}``` executed on the selected element. If the action is 'terminate', then the second image does not exist.
3. The third image is a sub-image cropped from the original screenshot, focusing on the selected element, which is highlighted with a red bounding box and arrow for better visibility.
4. The A11Tree attributes of the selected element: {element_a11tree}.
                        
REMEBER:
- Do NOT directly copying the A11Tree attributes of the selected elements as instructions.
- Do NOT reference the distinctive red indicator when describing UI elements.

Directly generate the operational instruction which can uniquely correspond to the selected element and contain all operations. Avoid "highlighted", "red box", "red circle" and "red point" in your output:
\end{Verbatim}
\end{tcolorbox}

\begin{tcolorbox}[title=Prompt For Rationales in All Trajectories]
\scriptsize
\begin{Verbatim}[breaklines=true, breaksymbol={}]
You are a GUI agent operating on a {os_name} device. Your task is to analyze the potential reason behind operations.

You will be provided with:
1. The first image is a original screenshot from an {application}, which are marked to highlight the selected element.
2. The second image is the results of the operations ```{action}``` executed on the selected element. If the action is 'terminate', then the second image does not exist.
3. The third image is a sub-image cropped from the original screenshot, focusing on the selected element for better visibility.
4. The A11Tree attributes of the selected element: {element_a11tree}.
5. The task objective is `{task_objective}` and history trace is `{history}`.
                  
Guidelines:
- Examine the selected UI element and relevant contextual features that support task completion, considering both the objective and interaction history. {marker} higlighted in image is manually added to assist in identifying elements and **should not** been mentioned.
- Provide your reasoning in three sentences, ensuring alignment with the goal and labeled action, but do not cite the actual action or bounding box as justification, as these reflect hindsight rather than predictive insight.
- Restrict your analysis to details from the first image only, and avoid referencing image order.

For example:
The screenshot shows a file dialog with active selection on format dropdown. Changing the format completes the file configuration sub-task. Next, click 'Save' to confirm the selection.  


Focus only on the thoughts leading up to the event, not what happens after. Do not refer to visual cues like highlights, red boxes, or circles in your description and think aloud as you work on this task:
\end{Verbatim}
\end{tcolorbox}

\begin{tcolorbox}[title=Prompt For Instruction Boost]
\scriptsize
\begin{Verbatim}[breaklines=true, breaksymbol={}]
You are a helpful assistant to refine the given user instructions. The refined instructions should be clear, polite, and structured as a direct request or question, often including:
- A specific action or configuration change.
- Optional context or reasoning (e.g., "I want to ensure my browsing is private").
- A conversational yet concise tone

**Some Examples for reference:**
- "Configure the system to show seconds in the taskbar clock."
- "Can you configure VS Code to automatically check for updates on startup?"
- "Could you assist me in cleaning up my computer by removing any tracking data that Chrome might have stored?"
- "I want to hear something soft and beautiful music when Windows starts up. Can you set that MP3 file I like as my startup sound?"
- "I don't want to see all these news on the home page of Microsoft Edge. Remove them in Page settings."

**Output Format:**
You should provide various styles and the output should be structured as follows:
```
Can you ...;
I want to ...; 
I don't want to ...;
...;
```

**Input instruction**: {task_objective}
Rewrite the provided input instructions, ensuring they are actionable, polite, and include necessary details. Use ";" to separate different output:
\end{Verbatim}
\end{tcolorbox}

\end{document}